\documentclass[10pt,twocolumn,letterpaper]{article}

\usepackage{3dv}
\usepackage{times}
\usepackage{epsfig}
\usepackage{graphicx}
\usepackage{amsmath}
\usepackage{amssymb}

\usepackage{subfigure} 
\usepackage{amsfonts} 
\usepackage{placeins}
\usepackage{algorithm}
\usepackage{algorithmic}
\usepackage{array}
\usepackage{overpic}
\usepackage{wrapfig}
\usepackage{bm}
\usepackage{comment}
\usepackage{pgfplots}
\usepackage{pgf}
\usepackage{tikz}
\usepackage{soul}
\usepackage{tabstackengine}

\newcommand{\X}{\mathcal{X}}

\newcommand{\new}[1]{#1}
\definecolor{pose}{rgb}{0.92900,0.69400,0.12500}
\definecolor{our}{rgb}{0.00000,0.44700,0.74100}
\definecolor{style}{rgb}{0.85000,0.32500,0.09800}
\newcommand{\pose}[1]{\textcolor{pose}{#1}}
\newcommand{\our}[1]{\textcolor{our}{#1}}
\newcommand{\style}[1]{\textcolor{style}{#1}}
\newtheorem{property}{Property}

\newcommand{\veryshortarrow}[1][3pt]{\mathrel{%
   \hbox{\rule[\dimexpr\fontdimen22\textfont2-.2pt\relax]{#1}{.4pt}}%
   \mkern-4mu\hbox{\usefont{U}{lasy}{m}{n}\symbol{41}}}}

\usepackage[pagebackref=true,breaklinks=true,letterpaper=true,colorlinks,bookmarks=false]{hyperref}

\threedvfinalcopy 

\def\threedvPaperID{37} 
\def\httilde{\mbox{\tt\raisebox{-.5ex}{\symbol{126}}}}

\ifthreedvfinal\fi
\begin{document}

\title{Instant recovery of shape from spectrum via latent space connections}

\author{Riccardo Marin\\
University of Verona\\
{\tt\small riccardo.marin\_01@univr.it}
\and
Arianna Rampini\\
Sapienza University of Rome\\
{\tt\small rampini@di.uniroma1.it}
\and
Umberto Castellani\\
University of Verona\\
{\tt\small umberto.castellani@univr.it}
\and
Emanuele Rodol\`a\\
Sapienza University of Rome\\
{\tt\small rodola@di.uniroma1.it}
\and
Maks Ovsjanikov\\
Ecole Polytechnique, IP Paris\\
{\tt\small maks@lix.polytechnique.fr}
\and
Simone Melzi\\
Ecole Polytechnique, IP Paris \\
Sapienza University of Rome
\\
{\tt\small melzi@di.uniroma1.it}
}

\maketitle

\begin{abstract}
We introduce the first learning-based method for recovering shapes from Laplacian spectra. 
\new{Our model consists of a cycle-consistent module that maps between learned latent vectors of an auto-encoder and sequences of eigenvalues. This module provides an efficient and effective linkage between Laplacian spectrum and geometry.}
Our data-driven approach replaces the need for ad-hoc regularizers required by prior methods, while providing more accurate results at a fraction of the computational cost. Our learning model applies without modifications across different dimensions (2D and 3D shapes alike), representations (meshes, contours and point clouds), as well as across different shape classes, and admits arbitrary resolution of the input spectrum without affecting complexity. 
\new{The increased flexibility allows us to address notoriously difficult tasks in 3D vision and geometry processing within a unified framework, including shape generation from spectrum, mesh super-resolution, shape exploration, style transfer, spectrum estimation from point clouds, segmentation transfer and point-to-point matching.}
\end{abstract}

\section{Introduction}
\label{sec:introduction}
Constructing compact encodings of geometric shapes lies at the heart of 2D and 3D Computer Vision. While earlier approaches have concentrated on handcrafted representations, with the advent of geometric deep learning \cite{bronstein2017geometric,masci2016geometric}, data-driven \emph{learned} feature encodings have gained prominence. A desirable property in many applications, such as shape exploration and synthesis, is to be able to recover the shape from its (latent) encoding, and various auto-encoder architectures have been designed to solve this problem \cite{achlioptas2018learning,litany2018deformable,mo2019structurenet,gao2019sdm}. Despite significant progress in this area, the structure of the latent vectors is arduous to control. For example, the dimensions of the latent vectors typically lack a canonical ordering, while invariance to various geometric deformations is often only learned by data augmentation or complex constraints on the intermediate features.

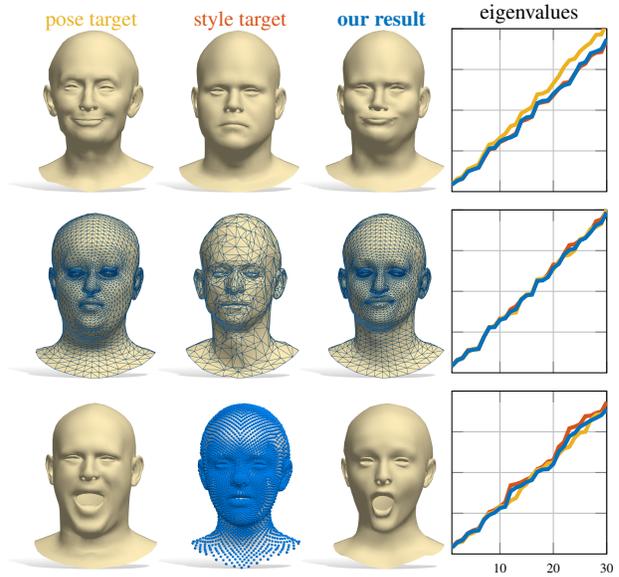
\begin{figure}[!t]
\centering
\vspace{0.3cm}
  \setlength{\tabcolsep}{0pt}
  \begin{tabular}{l r}
  \hspace{-0.2cm}
  
    \begin{minipage}{0.74\linewidth}

     \begin{overpic}[trim=0cm 0cm 0cm 0cm,clip,width=0.94\linewidth]{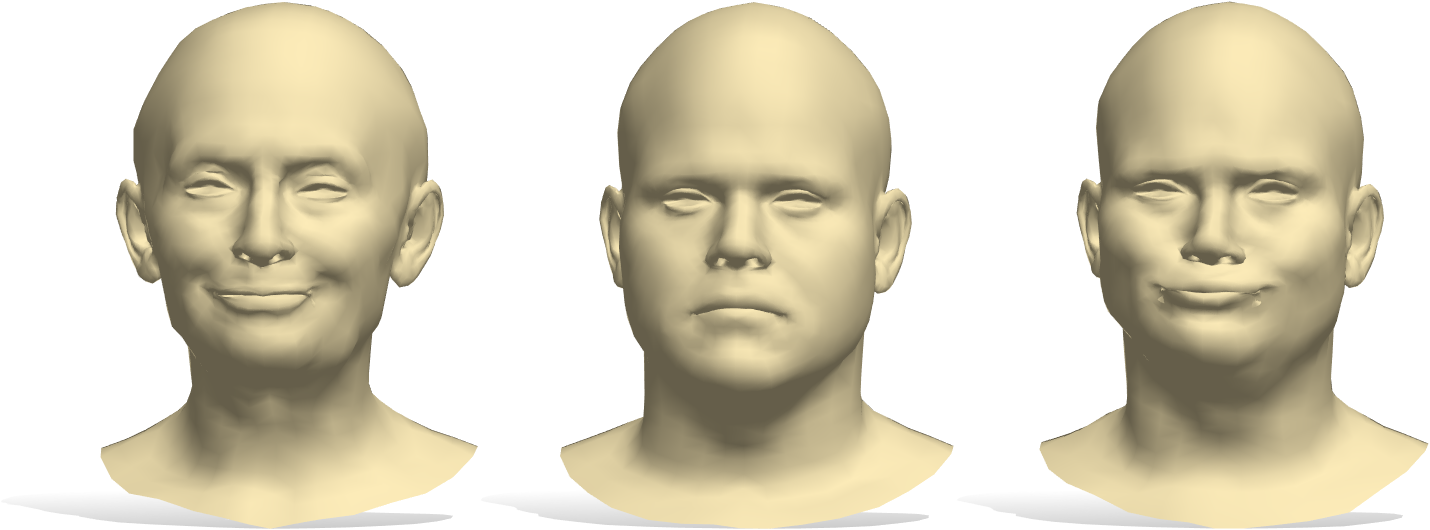}
          \put(8.5,38.2){\footnotesize \pose{pose target}}
        \put(42.5,38.2){\footnotesize \style{style target}}
        \put(75.5,38.2){\footnotesize \textbf{\our{our result}}}
    
   \end{overpic}
 \vspace{0.3cm}
 
 \begin{overpic}[trim=0cm 0cm 0cm 0cm,clip,width=0.94\linewidth]{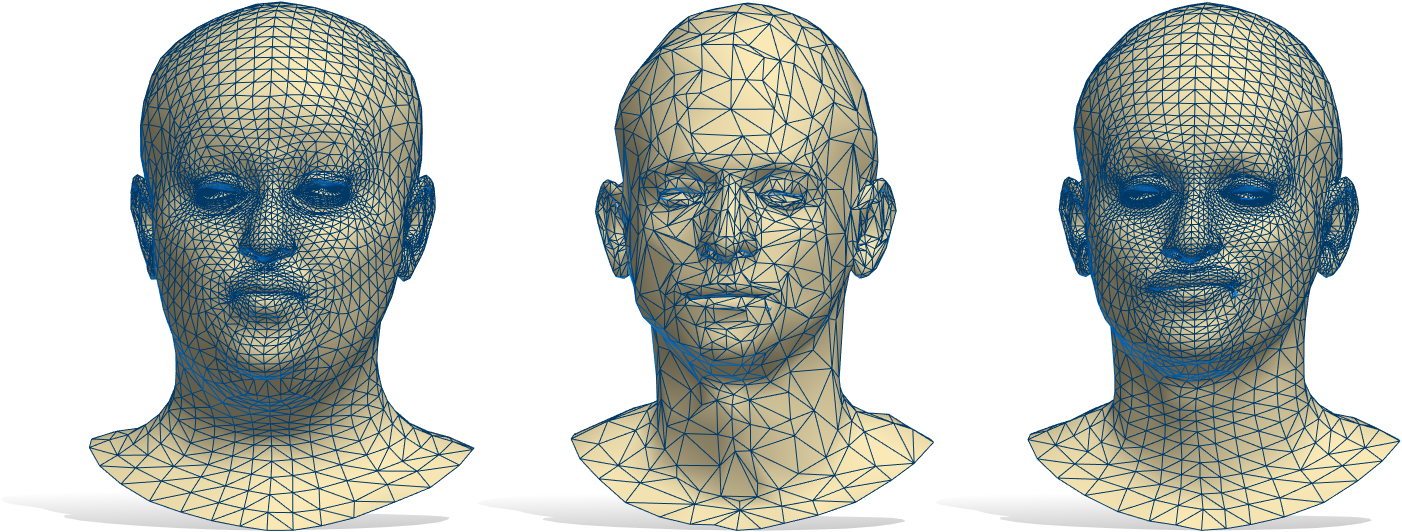}
    
   \end{overpic}
   
 \vspace{0.3cm}
 
 \begin{overpic}[trim=0cm 0cm 0cm 0cm,clip,width=0.94\linewidth]{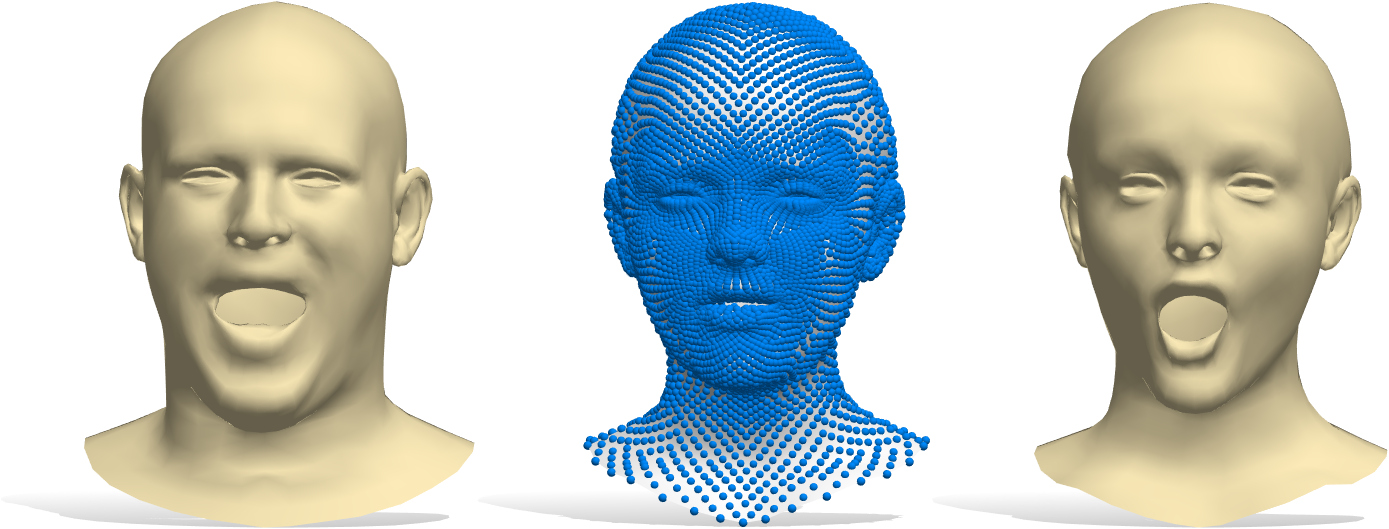}
    
   \end{overpic}
 
    \end{minipage}%
    &
\hspace{-0.6cm}
  
  \begin{minipage}{0.26\linewidth}
 \vspace{-0.3cm}
    
%
%
\definecolor{mycolor1}{rgb}{0.85000,0.32500,0.09800}%
\definecolor{mycolor2}{rgb}{0.92900,0.69400,0.12500}%
\definecolor{mycolor3}{rgb}{0.00000,0.44700,0.74100}%
\begin{tikzpicture}

\begin{axis}[%
width=0.95\linewidth,
height=\linewidth,
at={(0.758in,0.481in)},
scale only axis,
xmin=1,
xmax=30,
ymin=0,
ymax=2000,
xtick={10, 20, 30},
xticklabels={ , , },
ytick={500, 1000, 1500, 2000},
yticklabels={,  ,  },
xmajorgrids,
ymajorgrids,
every x tick label/.append style={font=\color{black}, font=\tiny},
every y tick label/.append style={font=\color{black}, font=\tiny},
title={\footnotesize eigenvalues},
title style={yshift=-0.75em},
axis background/.style={fill=white},
legend style={legend cell align=left,align=left,draw=white!15!black}
]
\addplot [color=mycolor1,solid,line width=1.5pt]
  table[row sep=crcr]{%
1	88.3991523402342\\
2	132.877175166086\\
3	157.533858589753\\
4	254.018259526285\\
5	277.000743109669\\
6	293.626727038495\\
7	410.179936684996\\
8	519.651989988256\\
9	529.364972645713\\
10	608.429572535167\\
11	637.275530022479\\
12	674.444189589679\\
13	712.990219448453\\
14	863.66149700102\\
15	885.297316186071\\
16	917.83830134501\\
17	1082.40772865865\\
18	1095.97085519538\\
19	1110.15434484794\\
20	1179.61286353169\\
21	1261.40147031271\\
22	1334.80606661506\\
23	1343.76075356628\\
24	1447.58295048208\\
25	1533.40967717059\\
26	1570.07209606373\\
27	1667.58581976392\\
28	1705.3784390348\\
29	1714.43273379947\\
30	1861.40993421404\\
};

\addplot [color=mycolor2,solid,line width=1.5pt]
  table[row sep=crcr]{%
1	90.28044258827\\
2	158.539796609419\\
3	183.866060129477\\
4	274.84667400481\\
5	302.754043578377\\
6	334.133191746416\\
7	462.955916634064\\
8	577.232883032865\\
9	596.021404297334\\
10	660.916226156052\\
11	711.145818931198\\
12	801.907965435575\\
13	825.420594542214\\
14	960.842239496039\\
15	998.53662478193\\
16	1039.69825479782\\
17	1192.9759794115\\
18	1204.65136667601\\
19	1235.87360805497\\
20	1328.1283775152\\
21	1414.94621622385\\
22	1529.97946862345\\
23	1611.77078596784\\
24	1631.39228882996\\
25	1722.17786111811\\
26	1781.04807645489\\
27	1828.2137847098\\
28	1925.44319850625\\
29	1929.87223870498\\
30	2074.91974579112\\
};

\addplot [color=mycolor3,solid,line width=1.5pt]
  table[row sep=crcr]{%
1	85.6136855505045\\
2	135.820517641082\\
3	161.696509951749\\
4	246.834876901113\\
5	272.577130983847\\
6	294.473467814782\\
7	418.737870455242\\
8	528.732889503008\\
9	537.381207459221\\
10	616.032076712776\\
11	649.689329671978\\
12	664.542480384075\\
13	696.723807368296\\
14	870.322848659685\\
15	911.560284458103\\
16	923.690307976056\\
17	1085.87836211517\\
18	1114.57678301966\\
19	1119.61999053223\\
20	1201.22189334032\\
21	1263.53521023093\\
22	1305.08511285\\
23	1324.62934981694\\
24	1439.35230583782\\
25	1566.41958179824\\
26	1588.13581509802\\
27	1698.335229754\\
28	1720.59444841301\\
29	1749.11152609645\\
30	1872.0244133914\\
};


\end{axis}
\end{tikzpicture}%

 \vspace{-0.15cm}

%
%
\definecolor{mycolor1}{rgb}{0.85000,0.32500,0.09800}%
\definecolor{mycolor2}{rgb}{0.92900,0.69400,0.12500}%
\definecolor{mycolor3}{rgb}{0.00000,0.44700,0.74100}%
\begin{tikzpicture}

\begin{axis}[%
width=0.95\linewidth,
height=\linewidth,
at={(0.758in,0.481in)},
scale only axis,
xmin=1,
xmax=30,
ymin=0,
ymax=2000,
xtick={10, 20, 30},
xticklabels={ , , },
ytick={500, 1000, 1500, 2000},
yticklabels={,  ,  },
xmajorgrids,
ymajorgrids,
every x tick label/.append style={font=\color{black}, font=\tiny},
every y tick label/.append style={font=\color{black}, font=\tiny},
axis background/.style={fill=white},
legend style={legend cell align=left,align=left,draw=white!15!black}
]
\addplot [color=mycolor1,solid,line width=1.5pt]
  table[row sep=crcr]{%
1	82.5881500244141\\
2	158.514190673828\\
3	184.762008666992\\
4	277.695770263672\\
5	288.848175048828\\
6	290.600280761719\\
7	440.019134521484\\
8	560.083862304688\\
9	571.2138671875\\
10	665.074768066406\\
11	669.392517089844\\
12	779.250854492188\\
13	815.903930664063\\
14	894.261962890625\\
15	952.103332519531\\
16	969.56591796875\\
17	1132.56213378906\\
18	1140.72021484375\\
19	1207.36730957031\\
20	1328.67932128906\\
21	1337.91833496094\\
22	1458.46691894531\\
23	1568.1181640625\\
24	1578.71520996094\\
25	1629.20190429688\\
26	1659.57263183594\\
27	1737.34997558594\\
28	1861.50561523438\\
29	1883.65173339844\\
30	1988.09387207031\\
};

\addplot [color=mycolor2,solid,line width=1.5pt]
  table[row sep=crcr]{%
1	86.3383139092741\\
2	148.332991374687\\
3	177.671099124447\\
4	254.090552396558\\
5	293.799486815285\\
6	303.738530773102\\
7	443.569994848689\\
8	559.342383067259\\
9	564.886110564849\\
10	650.092959222793\\
11	693.461650089226\\
12	729.684925166302\\
13	740.697451141666\\
14	907.76895930657\\
15	963.35182675226\\
16	973.775854726357\\
17	1145.81070000986\\
18	1153.18794238848\\
19	1181.76778456283\\
20	1268.34978972411\\
21	1367.79121844941\\
22	1434.97949931278\\
23	1449.02387246026\\
24	1503.40933523374\\
25	1641.4417205254\\
26	1663.10657024077\\
27	1769.70409141226\\
28	1790.6180472477\\
29	1828.57964306798\\
30	2018.35716394941\\
};

\addplot [color=mycolor3,solid,line width=1.5pt]
  table[row sep=crcr]{%
1	83.5339660049848\\
2	153.005144092871\\
3	179.581207692812\\
4	267.94914017119\\
5	283.602658970878\\
6	294.049444750388\\
7	439.171029460367\\
8	555.983005857916\\
9	560.174805672672\\
10	648.192102751779\\
11	670.259215590373\\
12	752.552440044371\\
13	779.785422113178\\
14	899.445488866064\\
15	951.953360954814\\
16	958.136836593646\\
17	1129.30337789975\\
18	1133.08964602617\\
19	1173.31996743406\\
20	1287.70104054125\\
21	1336.51489939567\\
22	1448.1326919952\\
23	1508.65787712268\\
24	1523.55214419919\\
25	1624.38239349378\\
26	1653.57001678545\\
27	1727.59036522703\\
28	1811.58114919144\\
29	1842.80923847472\\
30	1972.58543047206\\
};


\end{axis}
\end{tikzpicture}%
    
     \vspace{-0.15cm}
    
%
%
\definecolor{mycolor1}{rgb}{0.85000,0.32500,0.09800}%
\definecolor{mycolor2}{rgb}{0.92900,0.69400,0.12500}%
\definecolor{mycolor3}{rgb}{0.00000,0.44700,0.74100}%
\begin{tikzpicture}

\begin{axis}[%
width=0.95\linewidth,
height=\linewidth,
at={(0.758in,0.481in)},
scale only axis,
xmin=1,
xmax=30,
ymin=0,
ymax=2000,
xtick={10, 20, 30},
xticklabels={10, 20, 30},
ytick={500, 1000, 1500, 2000},
yticklabels={,  ,  },
xmajorgrids,
ymajorgrids,
every x tick label/.append style={font=\color{black}, font=\tiny},
every y tick label/.append style={font=\color{black}, font=\tiny},
axis background/.style={fill=white},
legend style={legend cell align=left,align=left,draw=white!15!black}
]
\addplot [color=mycolor1,solid,line width=1.5pt]
  table[row sep=crcr]{%
1	79.5686340332031\\
2	147.751129150391\\
3	176.951705932617\\
4	254.563293457031\\
5	290.340118408203\\
6	309.812347412109\\
7	414.487243652344\\
8	514.449768066406\\
9	539.833435058594\\
10	581.752075195313\\
11	635.413879394531\\
12	842.165649414063\\
13	858.546203613281\\
14	880.501892089844\\
15	915.809692382813\\
16	944.787414550781\\
17	1040.875\\
18	1069.78332519531\\
19	1134.54809570313\\
20	1148.50378417969\\
21	1314.31457519531\\
22	1374.32775878906\\
23	1542.52368164063\\
24	1563.27648925781\\
25	1592.87255859375\\
26	1692.53637695313\\
27	1697.88586425781\\
28	1730.66271972656\\
29	1748.22521972656\\
30	1865.64123535156\\
};

\addplot [color=mycolor2,solid,line width=1.5pt]
  table[row sep=crcr]{%
1	82.6492687544362\\
2	133.579013349477\\
3	151.604342859165\\
4	233.796412254886\\
5	259.660362044112\\
6	297.953549406802\\
7	428.776312998129\\
8	495.719910294317\\
9	517.145781903565\\
10	585.624597503925\\
11	589.688070709374\\
12	654.376283706083\\
13	686.470296790777\\
14	806.946106465201\\
15	857.140910215396\\
16	964.371384211749\\
17	1030.61962876152\\
18	1048.25451512194\\
19	1112.29329572646\\
20	1123.04609724129\\
21	1224.47034842448\\
22	1293.18633760695\\
23	1314.44618927495\\
24	1428.51156762868\\
25	1489.84847032913\\
26	1499.27458977992\\
27	1646.5062240243\\
28	1708.75539150798\\
29	1718.13417372683\\
30	1767.95453182715\\
};

\addplot [color=mycolor3,solid,line width=1.5pt]
  table[row sep=crcr]{%
1	76.982615904312\\
2	141.66938032615\\
3	168.608109833798\\
4	251.378337221368\\
5	280.34891585547\\
6	288.750822064438\\
7	419.643637443515\\
8	478.064257753249\\
9	507.004392787097\\
10	563.174936610813\\
11	578.944492769447\\
12	768.716568354763\\
13	813.799485754868\\
14	838.762468231502\\
15	884.14188860122\\
16	907.327415089648\\
17	1008.08161082509\\
18	1036.39746259518\\
19	1040.52755449064\\
20	1101.99193038128\\
21	1254.97200279225\\
22	1321.08449820412\\
23	1444.98343827297\\
24	1483.88567082626\\
25	1561.9852641987\\
26	1599.62885832924\\
27	1638.7286676973\\
28	1669.35912006669\\
29	1704.08694166583\\
30	1778.26443824192\\
};


\end{axis}
\end{tikzpicture}%
    
    \end{minipage}%
  \end{tabular}
\vspace{2mm}

\caption{\label{fig:teaser}Our spectral reconstruction enables correspondence-free style transfer. Given pose and style ``donors'' (left and middle columns respectively), we synthesize a new shape with the pose of the former and the style of the latter. The generation is driven by a learning-based eigenvalues alignment (rightmost plots). Our approach handles different resolutions (middle row) and representations (bottom row; the surface underlying the point cloud is for visualization purposes only).\vspace{-2mm}}
\end{figure}

At the same time, a classical approach in spectral geometry is to encode a shape using the sequence of eigenvalues (spectrum) of its Laplacian operator. This representation is useful since: (1) it does not require any training, (2) it can be computed on various data representations, such as point clouds or meshes, regardless of sampling density, (3) it enjoys well-known theoretical properties such as a natural ordering of its elements and invariance to isometries, and (4) as shown recently \cite{isosp,hamiltonian}, alignment of eigenvalues often promotes near-isometries, which is useful in multiple tasks such as non-rigid shape retrieval and matching problems.

Unfortunately, although encoding shapes via their Laplacian spectra \new{can be} straightforward \new{(at least for meshes)}, the inverse problem of recovering the shape is very difficult. Indeed, it is well-known that certain pairs of non-isometric shapes can have the same spectrum, or in other words ``one cannot hear the shape of a drum'' \cite{gordon1992one}. At the same time, recent evidence suggests that such cases are pathological and that \emph{in practice} it might be possible to recover a shape from its spectrum \cite{isosp}. Nevertheless, existing approaches \new{\cite{isosp}}, while able to \new{deform a shape into another} with a given spectrum, can produce highly unrealistic shapes with strong artifacts \new{failing in a large number of cases}.


In this paper, we combine the strengths of data-driven auto-encoders with those of spectral methods. Our key idea is to construct a single architecture capable of synthesizing a shape from a learned latent code and from its Laplacian eigenvalues. We show that by explicitly training networks that aim to translate between the learned latent codes and the spectral encoding, we can both recover a shape from its eigenvalues and moreover endow the latent space with certain desirable properties. Remarkably, our shape-from-spectrum solution is extremely efficient since it requires a single pass through a trained network, unlike expensive iterative optimization methods with ad-hoc regularizers \new{\cite{isosp}}. Furthermore, our trainable module acts as a proxy to differentiable eigendecomposition, while encouraging geometric consistency within the network. 
Overall, our key \textbf{contributions} can be summarized as follows:
\begin{itemize}
    \item We propose the first learning-based model \new{to robustly recover} shape from Laplacian spectra \emph{in a single pass};
    \item \new{For the first time, we provide a bidirectional  linkage  between  learned  3D  latent space and spectral geometric properties of 3D shapes;}
    \item Our model is {\em general}, in that it applies with no modifications to different classes even across different geometric representations and dimensions \new{and to data that does not belong to the datasets used at training time};
    \item We showcase our approach in multiple applications (e.g., Fig.~\ref{fig:teaser}), and show significant improvement over the state of the art; see Fig.~\ref{fig:isospec_bad} for an example.
\end{itemize}

\section{Related work}
%
%
%
Spectral quantities and in particular the eigenvalues of the Laplace-Beltrami operator provide an informative summary of the intrinsic geometry. For example, closed-form estimates and analytical bounds for surface area, genus and curvature in terms of the Laplacian eigenvalues have been obtained \cite{chavel1984eigenvalues}.
Given these properties, spectral shape analysis has been exploited in many computer vision  and computer graphics tasks such as shape retrieval \cite{reuter2005laplace}, 
description and matching \cite{sun2009concise,aubry2011wave,bronstein2011shape,ovsjanikov2012functional}, mesh segmentation \cite{reuter2010hierarchical}, sampling~\cite{oztireli2010spectral} and compression \cite{Karni:2000:SCM:344779.344924} among many others.
Typically, the intrinsic properties of the shape are computed from its explicit representation and are used to encode compact geometric features invariant to isometric deformations.
%
%
%

%
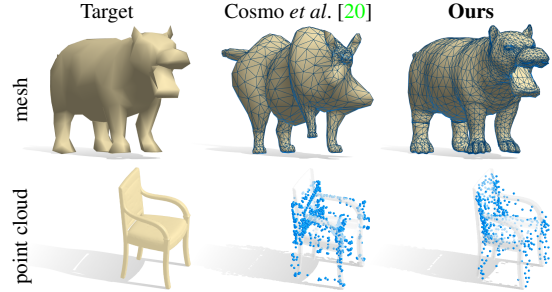
\begin{figure}
\vspace{0.1cm}
\centering

  \begin{overpic}
  [trim=0cm -0.2cm 0cm 0cm,clip,width=0.85\linewidth]{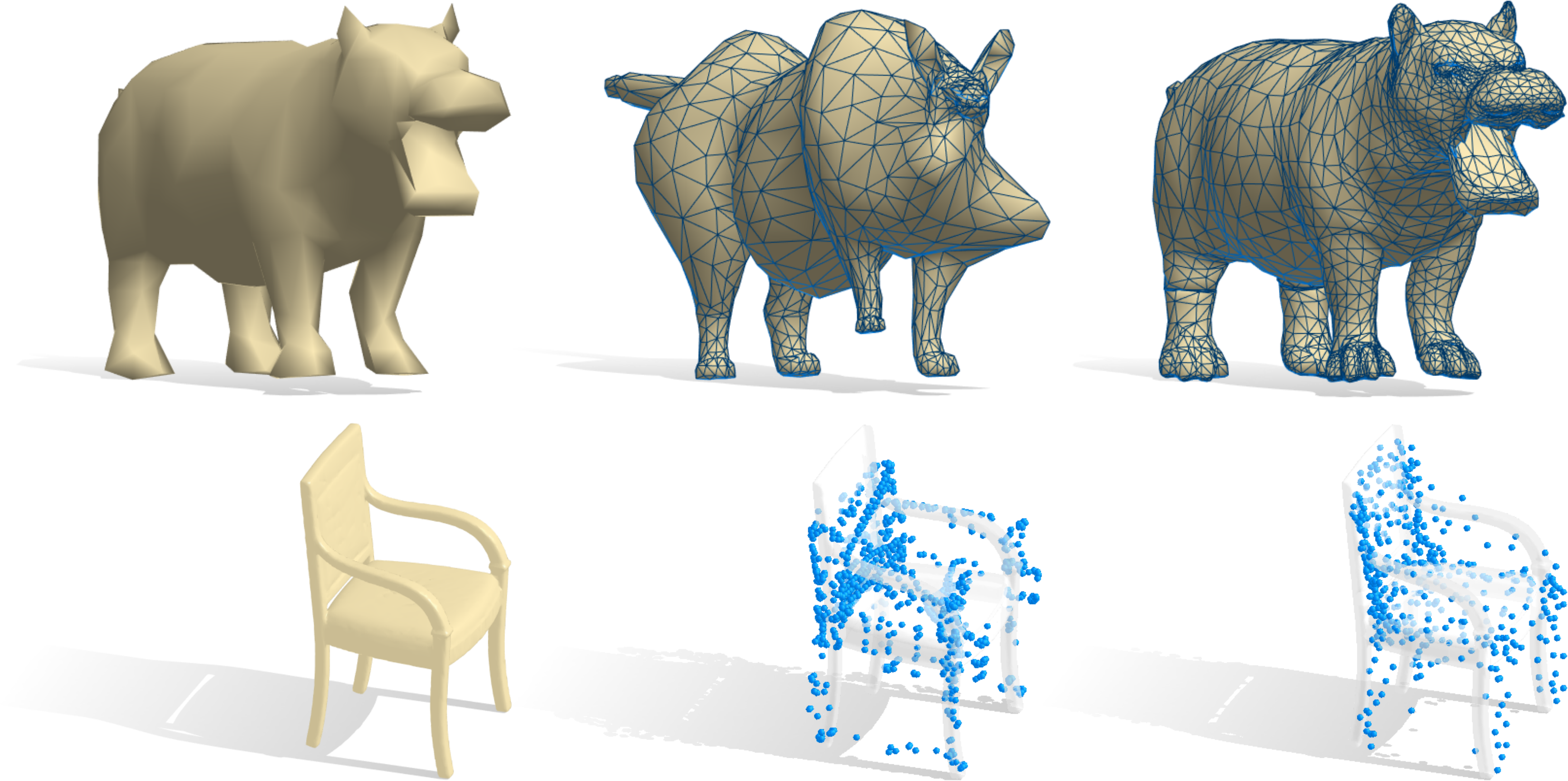}
    \put(12,51.5){\footnotesize Target}
    \put(39,51.5){\footnotesize Cosmo \etal~\cite{isosp}}
    \put(81,51.5){\footnotesize \textbf{Ours}}
    \put(-0.75,32){\rotatebox{90}{\footnotesize mesh}}
    \put(-0.75,1.6){\rotatebox{90}{\footnotesize point cloud}}

\end{overpic}
  \caption{\label{fig:isospec_bad} Comparison in estimating a shape from its Laplacian spectrum  between the state-of-the-art method \cite{isosp} (middle) and ours (right) for a mesh and a point cloud. The shapes recovered by our method are significantly closer to the target.}
\end{figure}
Recently, several works have started to address the inverse problem: namely, recovering an extrinsic embedding from the intrinsic encoding \cite{BosEynKouBro15,isosp}. 
This is closely related to the fundamental theoretical question of ``hearing the shape of the drum'' \cite{kac1966can,gordon1992one}. Although counterexamples have been proposed to show that 
in certain scenarios multiple shapes might have the same spectrum, there is recent work that proposes effective practical solutions to this problem.   
In \cite{BosEynKouBro15} the shape-from-operator method was proposed, aiming at obtaining the extrinsic shape from a Laplacian matrix where the 3D reconstruction was recovered after the estimation of the Riemannian metric in terms of edge lengths. In \cite{corman:hal-01741949} the intrinsic and extrinsic relations of geometric objects have been extensively defined and evaluated from both theoretical and practical aspects. The authors revised the framework of functional shape differences \cite{Rustamov:2013:MEI:2461912.2461959} to account of extrinsic structure extending the reconstruction task to non-isometric shapes and models obtained from physical simulation and animation. Several works have also been proposed to recover shapes purely from Laplacian \emph{eigenvalues} \cite{chu2005inverse,aasen2013shape,panine2016towards} or with mild additional information such as excitation amplitude in the case of musical key design \cite{bharaj2015computational}. 
Most closely related to ours in this area is the recent \emph{isopectralization} approach introduced in \cite{isosp}, that aims directly to estimate the 3D shape from the spectrum. 
This approach works well in the vicinity of a good solution but is both computationally expensive and, as we show below, can quickly produce unrealistic instances\new{, failing in a large number of cases in 3D, as shown in Fig.~\ref{fig:isospec_bad}} for two examples.
%

In this paper we contribute to this line of work, and propose to replace the heuristics used in previous methods such as \cite{isosp} with a purely data-driven approach for the first time. Our key idea is to design a deep neural network, that both constraints the space of solutions based on the set of shapes given at training, and at the same time, allows us to solve the isospectralization problem with a \emph{single forward pass}, thus avoiding expensive and error-prone optimization.
%
%

We note that a related idea has been recently proposed in \cite{huang2019operatornet} via the so-called OperatorNet architecture.  However, that work is based on shape difference operators \cite{Rustamov:2013:MEI:2461912.2461959} and as such requires a fixed source shape and functional maps to each shape in the dataset to properly synthesize a shape. Our approach is based on Laplacian eigenvalues alone and thus is completely correspondence-free.

Our approach also builds upon the recent work on learning generative shape models. A range of techniques have been proposed using the volumetric representations \cite{3dgan}, point cloud auto-encoders \cite{geodisent,achlioptas2018learning}, generative models based on meshes and implicit functions \cite{SURFNET,ATLASNET,litany2018deformable,kostrikov2018surface,chen2019learning}, and part structures \cite{GRASS,mo2019structurenet,gao2019sdm,SAGnet19}, among many others.

Although generative models, and in particular auto-encoders, have shown impressive performance, the structure of the latent space is typically difficult to control or analyze directly. To address this problem, some methods proposed a disentanglement of the latent space \cite{SAGnet19,geodisent} to split it in more semantic regions. Perhaps most closely related to ours in this domain, is the work in \cite{geodisent}, where the shape spectrum is used to promote disentanglement of the latent space intro intrinsic and extrinsic components, that can be controlled separately. Nevertheless, the resulting network does not allow to synthesize shapes from their spectra.

Extending the studies of these approaches, our work provides the first way to connect the learned latent space to the spectral one, thus inheriting the benefits and providing the versatility of moving across the two representations.  
This allows our network to synthesize shapes from their spectra, and also to relate shapes with very different input structure
(\eg, meshes and point clouds) across a vastness of sampling densities, enabling several novel applications.



\section{Background}\label{sec:bg}

We model shapes as connected 2-dimensional Riemannian manifolds $\X$ embedded in $\mathbb{R}^3$, possibly with boundary $\partial\X$, equipped with the standard metric. 
On each shape $\X$ we consider its positive semi-definite Laplace-Beltrami operator $\Delta_\X$, generalizing the classical notion of Laplacian from the Euclidean setting to curved surfaces.

\vspace{1ex}\noindent\textbf{Laplacian spectrum.}
$\Delta_\X$ admits an eigendecomposition
\begin{align}
    \Delta_\X \phi_i(x) = \lambda_i \phi_i(x) \quad~~ & x\in\mathrm{int}(\X)\\
    \langle \nabla \phi_i(x), \hat{n} (x) \rangle = 0 \quad~~ & x\in\partial\X\label{eq:bc}
\end{align}
into eigenvalues $\{\lambda_i\}$ and associated eigenfunctions $\{\phi_i\}$\footnote{Similarly to \cite{isosp} we use homogeneous Neumann boundary conditions; see Eq.~\eqref{eq:bc}, where $\hat{n}(x)$ denotes the outward normal to the boundary.}. 

The Laplacian eigenvalues of $\X$ (its {\em spectrum}) form a discrete set, which is canonically ordered into a non-decreasing sequence
\begin{align}
    \mathrm{Spec}(\X) := \{0=\lambda_0<\lambda_1\le\lambda_2\le\cdots\}\,.
\end{align}

In the special case where $\X$ is an interval in $\mathbb{R}$, the eigenvalues $\lambda_i$ correspond to the (squares of) oscillation frequencies of Fourier basis functions $\phi_i$. This provides us with a connection to classical Fourier analysis, and with a natural notion of hierarchy induced by the ordering of the eigenvalues.
In the light of this analogy, in practice, one is usually interested in a limited bandwidth consisting of the first $k>1$ eigenvalues; typical values in geometry processing applications range from $k=30$ to $100$.

Furthermore, the spectrum is {\em isometry-invariant}, \ie, it does not change with  deformations of the shape that preserve geodesic distances (\eg, changes in pose). 


%
\begin{figure}[t]
  \centering
  \begin{overpic}
  [trim=0cm -4cm 0cm 0cm,clip,width=0.9\linewidth]{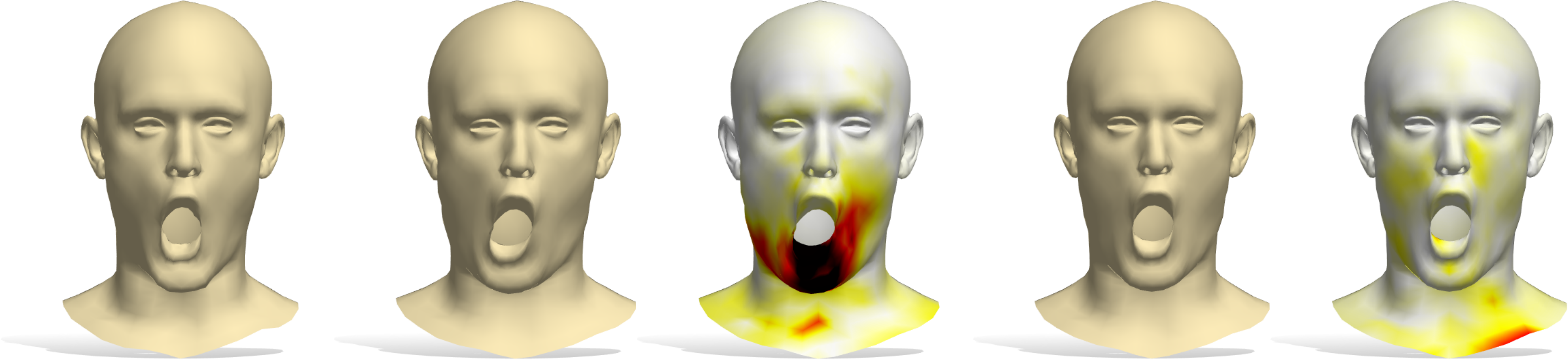}
  \put(5.2,2.7){\footnotesize unknown}
  \put(7.8,0.2){\footnotesize target}
  \put(35,2){\footnotesize linear FEM}
  \put(75,2){\footnotesize cubic FEM}
   \end{overpic}
  
  \caption{\label{fig:FEM}Reconstruction examples of our shape-from-spectrum pipeline. We show the results obtained with two different inputs: the eigenvalues of the Laplacian discretized with linear FEM, and those of the cubic FEM discretization. The heatmap encodes point-wise reconstruction error, growing from white to dark red.}
  
\end{figure}
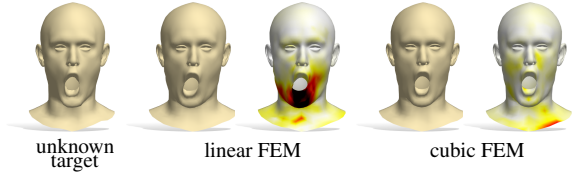

\vspace{1ex}\noindent\textbf{Discretization.}
In the discrete setting, we represent shapes as triangle meshes $X=(V,T)$ with $n$ vertices $V$ and $m$ triangular faces $T$; depending on the application, we will also consider unorganized point clouds. Vertex coordinates in both cases are represented by a matrix $\mathbf{X}\in\mathbb{R}^{n\times 3}$.

The Laplace-Beltrami operator $\Delta_\X$ is discretized as a $n\times n$ matrix via the finite element method (FEM)~\cite{ciarlet2002finite}. In the simplest setting (\ie, linear finite elements), this discretization corresponds to the cotangent Laplacian \cite{pinkall1993computing}; however, in this paper we use {\em cubic} FEM (see \eg \cite[Sec. 4.1]{reuter2010hierarchical} for a clear treatment), since it yields a more accurate discretization as shown in Fig.~\ref{fig:FEM}. Differently from \cite{isosp,hamiltonian}, this comes at virtually no additional cost for our pipeline, as we show in the sequel.
On point clouds, \new{$\Delta_\X$} can be discretized using the approach described in \cite{clarenz2004finite,boscaini2016anisotropic}.


\vspace{-0.0cm}

\section{Method}
Our main contribution is a deep learning model for recovering shapes from Laplacian eigenvalues. Our model operates in an end-to-end fashion: given a spectrum as input, it directly yields a shape with a single forward pass, thus avoiding expensive test-time optimization.

\setlength{\columnsep}{7pt}
\setlength{\intextsep}{1pt}
\begin{wrapfigure}[3]{r}{0.4\linewidth}
\vspace{-0.5cm}
\begin{center}
\begin{overpic}
[trim=0cm 0cm 0cm 0cm,clip,width=1.0\linewidth]{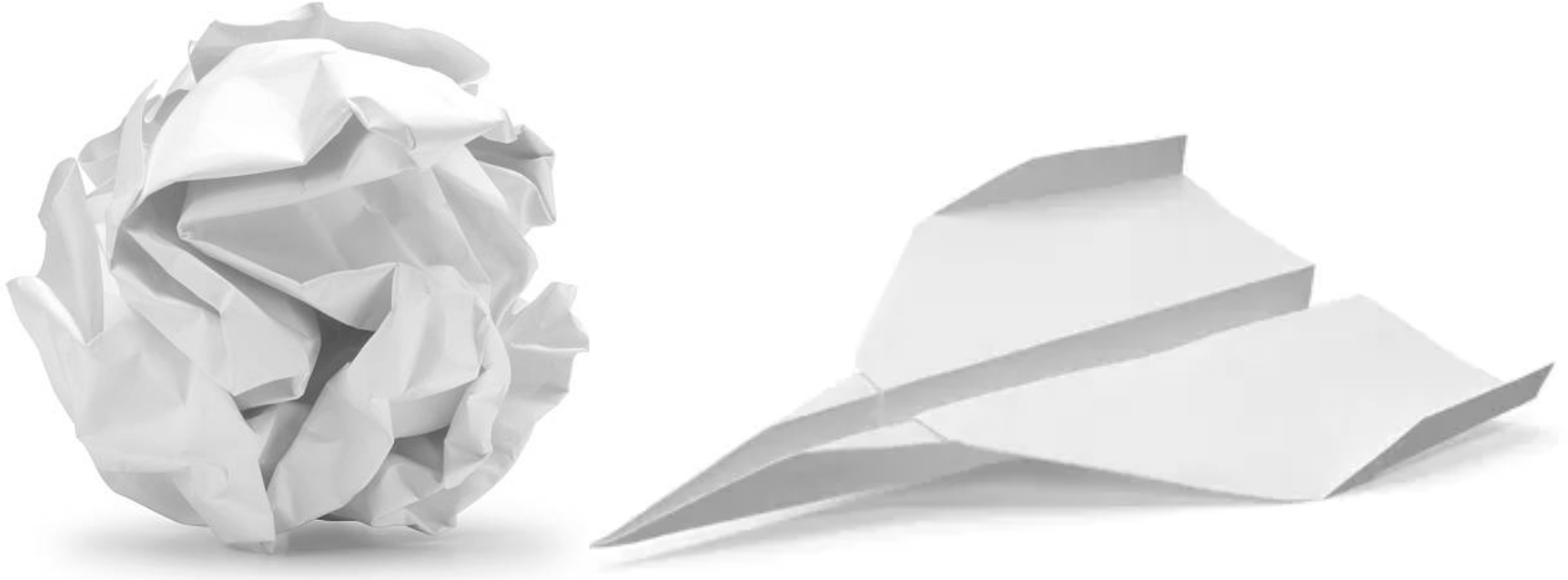}
\end{overpic}
\end{center}
\end{wrapfigure}
\vspace{1ex}\noindent\textbf{Motivation.} Our rationale lies in the observation that shape semantics can be learned from the data, rather than by relying upon the definition of ad-hoc regularizers~\cite{isosp}, often resulting in unrealistic reconstructions.
For example, a sheet of paper can be {\em isometrically} crumpled or folded into a plane (see inset figure). Since both embeddings have the same eigenvalues, the desirable reconstruction must be imposed as a prior.
By taking a data-driven approach, we make our method aware of the ``space of realistic shapes'', yielding both a dramatic improvement in accuracy and \new{efficiency,} and enabling new \new{interactive} applications.

\begin{figure}[t]
  \centering
  \begin{overpic}
  [trim=0cm -0.2cm 0cm 0cm,clip,width=1\linewidth]{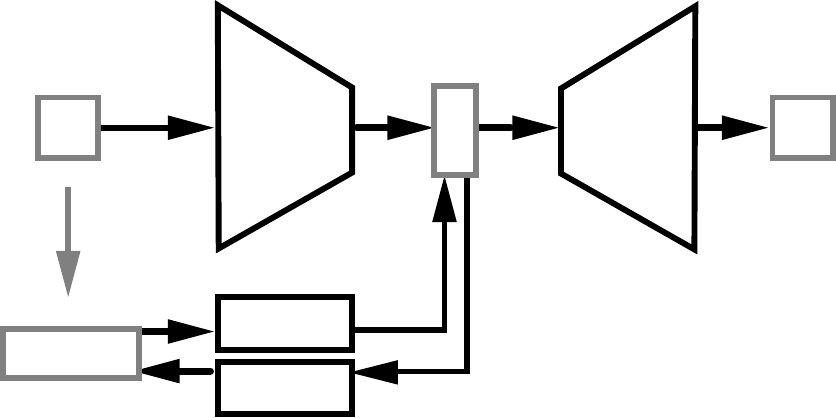}
  \put(6.3,35.5){$\X$}
  \put(94.2,35.5){$\tilde{\X}$}
  \put(0.8,8.9){$\mathrm{Spec}(\X)$}
  \put(32,36){$\mathbf{E}$}
  \put(74,36){$\mathbf{D}$}
  \put(33,12.5){$\pi$}
  \put(33,5.5){$\rho$}
  \put(53,36){$\mathbf{v}$}
  \end{overpic}
  \caption{\label{fig:model}Our network model. The input shape $\X$ and its Laplacian spectrum $\mathrm{Spec}(\X)$ are passed, respectively, through an AE enforcing $\X\approx\tilde{\X}$, and an invertible module $(\pi,\rho)$ mapping the eigenvalue sequence to a latent vector $\mathbf{v}$. The two branches are trained simultaneously, forcing $\mathbf{v}$ to be updated accordingly. The trained model allows to recover the shape purely from its eigenvalues via the composition $D(\pi(\mathrm{Spec}(\X)))\approx\X$.
  \vspace{-2mm}}
\end{figure}
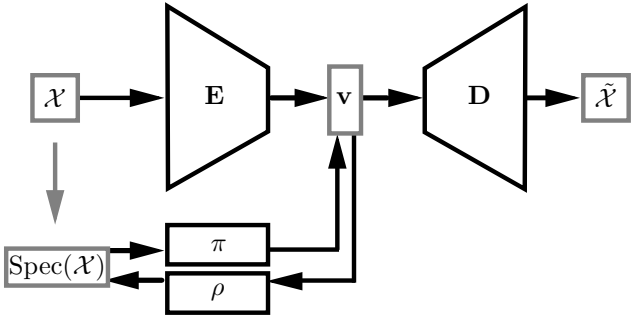

\vspace{1ex}\noindent\textbf{Latent space connections.}
Our key idea is to construct an auto-encoder (AE) neural network architecture, augmented by explicitly modeling the connections between the latent space of the AE and the Laplacian spectrum of the input shape; see Fig.~\ref{fig:model} for an illustration of our learning model.

Loosely speaking, our approach can be seen as implementing a coupling between two latent spaces: a learned one that operates on the shape embedding $\X$, and the one provided by the eigenvalues $\mathrm{Spec}(\X)$. In the former case, the \emph{encoder} $E$ is trainable, whereas the mapping $\X \rightarrow \mathrm{Spec}(\X)$ is provided via the eigen-decomposition and fixed a priori.  
Finally, we introduce the two coupling mappings $\pi, \rho$, trained with a \new{bidirectional} loss, to both enable communication across the latent spaces and to tune the learned space by endowing it with structure contained in $\mathrm{Spec}(\X)$.

We phrase our overall training loss as follows:
\begin{align}
    \ell &= \ell_\X + \alpha\ell_\lambda\,,\quad\mathrm{with}\label{eq:loss}\\
    \ell_\X &= \frac{1}{n}\| D(E(\mathbf{X})) - \mathbf{X} \|_F^2 \label{eq:rloss}\\
    \ell_\lambda &= \frac{1}{k}(\| \pi(\bm{\lambda}) - E(\mathbf{X}) \|_2^2 + \| \rho(E(\mathbf{X})) - \bm{\lambda} \|_2^2)\label{eq:llambda}
\end{align}
where $\bm{\lambda}$ is a vector containing the first $k$ eigenvalues in $\mathrm{Spec}(\X)$, \new{$\mathbf{X}$ is the matrix of point coordinates, $E$ is the encoder, $D$ is the decoder (Fig.~\ref{fig:model})}, $\|\cdot\|_F$ denotes the Frobenius norm, and $\alpha=10^{-4}$ controls the relative strengths of the reconstruction loss $\ell_\X$ and the spectral term $\ell_\lambda$.  
The blocks $D$, $E$, $\pi$, and $\rho$ are learnable and parametrized by a neural network (see the supplementary material for implementation details). 
%
Eq.~\eqref{eq:llambda} enforces $\rho \approx \pi^{-1}$; in other words, $\pi$ and $\rho$ form a translation block between the latent vector and the spectral encoding of the shape.

\new{At test time, we recover a shape from the spectrum $\mathrm{Spec}$ simply via the composition $D(\pi(\mathrm{Spec}))$ (Section \ref{sec:results}). For additional applications we refer to Section \ref{sec:apps}.}

\begin{center}
\begin{figure}[t!]
\centering
\setlength{\tabcolsep}{0pt}


    \hspace{0.17cm}
        \begin{overpic}[trim=0cm -4.5cm 0cm 0cm,clip,width=0.63\linewidth]{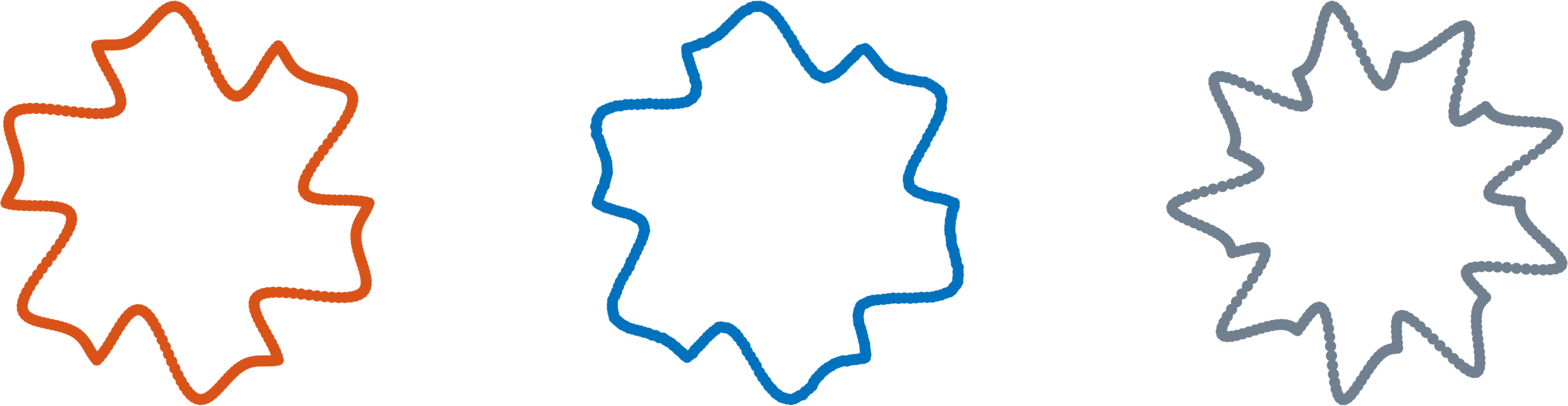}
            \put(6,39){\footnotesize \style{Target}}
            \put(44,39){\footnotesize \textbf{\our{Ours}}}
            \put(83.5,39){\footnotesize NN}
        \end{overpic}%
            \vspace{-0.22cm}
        \hspace{0.2cm}%
%
%
\definecolor{mycolor1}{rgb}{0.85000,0.32500,0.09800}%
\definecolor{mycolor2}{rgb}{0.4375,0.5,0.5625}%
\definecolor{mycolor3}{rgb}{0.00000,0.44700,0.74100}%
\begin{tikzpicture}

\begin{axis}[%
width=0.24\linewidth,
height=0.24\linewidth,
scale only axis,
xmin=1,
xmax=30,
ymin=0,
ymax=50,
ylabel style={yshift=-2.8em},
xtick={10, 20, 30},
xticklabels={},
ytick={12.5, 25, 37.5, 50},
yticklabels={,  ,  },
xmajorgrids,
ymajorgrids,
every x tick label/.append style={font=\color{black}, font=\tiny},
every y tick label/.append style={font=\color{black}, font=\tiny},
axis background/.style={fill=white},
legend columns=-1,
legend style={column sep=1ex,
at={(-2.5,-0.4)},anchor=south west,legend cell align=left,align=left,draw=white!15!black}
]
\addplot [color=mycolor1,solid,line width=1.5pt]
  table[row sep=crcr]{%
1	1.37872231006622\\
2	1.37890422344208\\
3	2.7665741443634\\
4	2.7677218914032\\
5	3.34282541275024\\
6	6.24738645553589\\
7	10.4842185974121\\
8	10.4901294708252\\
9	13.4960765838623\\
10	13.4985866546631\\
11	14.1016464233398\\
12	16.8866195678711\\
13	17.2387828826904\\
14	17.2456836700439\\
15	18.6004486083984\\
16	18.7306976318359\\
17	18.7523250579834\\
18	21.7695770263672\\
19	26.8373050689697\\
20	26.8572940826416\\
21	26.9531917572021\\
22	30.5512752532959\\
23	30.5710945129395\\
24	36.9584655761719\\
25	39.3228034973145\\
26	39.3857078552246\\
27	39.5388946533203\\
28	39.546817779541\\
29	44.1793518066406\\
30	47.423885345459\\
};
\addlegendentry{target};

\addplot [color=mycolor2,solid,line width=1.5pt]
  table[row sep=crcr]{%
1	1.30389818236728\\
2	1.30399550980237\\
3	2.77937834705462\\
4	2.78067564065808\\
5	3.8902424651995\\
6	3.89284962386677\\
7	4.45065626958973\\
8	4.45436687805225\\
9	5.88764191296082\\
10	9.74676028101391\\
11	9.74878144741579\\
12	13.5752955302473\\
13	13.5795065639385\\
14	15.2069559148011\\
15	17.762782470684\\
16	17.766622037998\\
17	21.5234480551866\\
18	21.5404868448217\\
19	21.6930670648003\\
20	21.7021006728999\\
21	28.0583835129949\\
22	28.0857552319879\\
23	29.463366679118\\
24	29.4712761553423\\
25	30.2172390372834\\
26	31.3237189600088\\
27	31.3498903073025\\
28	37.9544327245578\\
29	37.9924001569834\\
30	40.1913967597477\\
};
\addlegendentry{nn 1.3};

\addplot [color=mycolor3,solid,line width=1.5pt]
  table[row sep=crcr]{%
1	1.35993523817493\\
2	1.36188792346868\\
3	3.09666753618706\\
4	3.10430806479455\\
5	4.02373140512349\\
6	6.09792756880833\\
7	10.3046456960492\\
8	10.3070704785903\\
9	11.4576633889389\\
10	11.9138218717559\\
11	11.9385146387911\\
12	16.2024357601944\\
13	17.1433394786193\\
14	17.2185988160487\\
15	19.2785575943143\\
16	19.313754268478\\
17	19.4195518308542\\
18	22.1420794522962\\
19	25.2735054399402\\
20	25.315094043943\\
21	28.0152805193488\\
22	29.9864465679727\\
23	30.0633826485017\\
24	33.8980776931697\\
25	37.2196923355586\\
26	37.2907963960102\\
27	38.7540750187577\\
28	38.8596277139314\\
29	40.9409123351988\\
30	44.500534912163\\
};
\addlegendentry{Our 0.41};

\legend{}
\end{axis}

\end{tikzpicture}%
    \begin{minipage}{\linewidth}
        \begin{overpic}[trim=0cm -1cm 0cm 0cm,clip,width=0.7\linewidth]{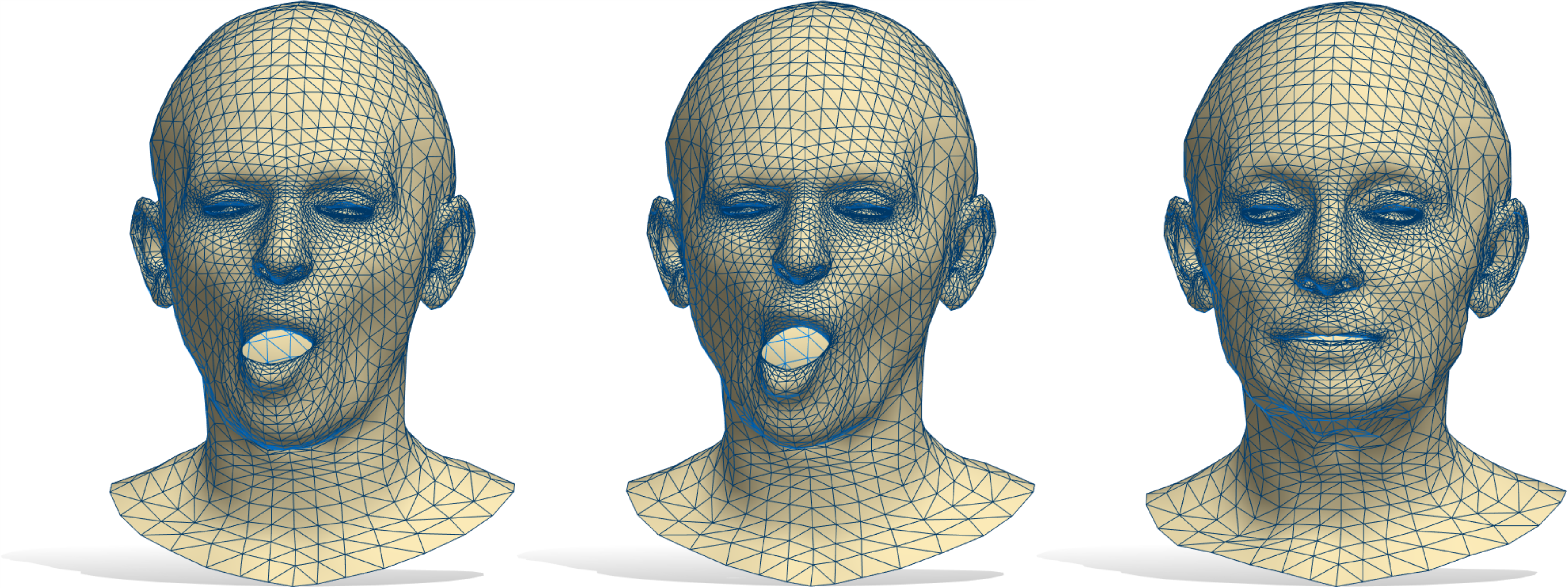}
        \end{overpic}
        \hspace{-0.1cm}
%
%
\definecolor{mycolor1}{rgb}{0.85000,0.32500,0.09800}%
\definecolor{mycolor2}{rgb}{0.4375,0.5,0.5625}%
\definecolor{mycolor3}{rgb}{0.00000,0.44700,0.74100}%
\begin{tikzpicture}

\begin{axis}[%
width=0.24\linewidth,
height=0.24\linewidth,
at={(0.758in,0.481in)},
scale only axis,
xmin=1,
xmax=30,
ymin=0,
ymax=2000,
xtick={10, 20, 30},
xticklabels={, , },
ytick={500, 1000, 1500, 2000},
yticklabels={,  ,  },
xmajorgrids,
ymajorgrids,
every x tick label/.append style={font=\color{black}, font=\tiny},
every y tick label/.append style={font=\color{black}, font=\tiny},
axis background/.style={fill=white},
legend columns=-1,
legend style={column sep=1ex,
at={(-2.5,-0.4)},anchor=south west,legend cell align=left,align=left,draw=white!15!black}
]
\addplot [color=mycolor1,solid,line width=1.5pt]
  table[row sep=crcr]{%
1	86.6273765305543\\
2	165.26117523603\\
3	185.196623689017\\
4	278.905115198141\\
5	300.009845384072\\
6	313.575211709381\\
7	474.882041758622\\
8	579.969742913116\\
9	584.757296781023\\
10	665.578379496478\\
11	689.423138608881\\
12	809.144394377749\\
13	827.345870708427\\
14	935.964786302512\\
15	982.064017343541\\
16	1054.94151469817\\
17	1172.98776706077\\
18	1177.95006863411\\
19	1229.89036679758\\
20	1329.04680304488\\
21	1465.27138364613\\
22	1545.39186958685\\
23	1619.56107074355\\
24	1632.4580320236\\
25	1704.23743287048\\
26	1738.40922563501\\
27	1792.30940064563\\
28	1911.31001937878\\
29	1936.43429277509\\
30	2096.95703422783\\
};
\addlegendentry{target};

\addplot [color=mycolor2,solid,line width=1.5pt]
  table[row sep=crcr]{%
1	91.7472381591797\\
2	159.667022705078\\
3	183.195938110352\\
4	276.535858154297\\
5	303.457275390625\\
6	333.736877441406\\
7	463.899108886719\\
8	583.762268066406\\
9	595.8525390625\\
10	661.053894042969\\
11	716.398681640625\\
12	782.897033691406\\
13	811.747680664063\\
14	966.503112792969\\
15	991.158264160156\\
16	1042.50915527344\\
17	1184.53698730469\\
18	1203.81127929688\\
19	1227.93994140625\\
20	1323.14086914063\\
21	1426.46008300781\\
22	1495.33776855469\\
23	1554.68530273438\\
24	1577.06042480469\\
25	1719.28259277344\\
26	1750.45520019531\\
27	1802.328125\\
28	1900.81665039063\\
29	1921.48620605469\\
30	2066.71411132813\\
};
\addlegendentry{nn 0.14};

\addplot [color=mycolor3,solid,line width=1.5pt]
  table[row sep=crcr]{%
1	87.3951072710542\\
2	165.227533716991\\
3	186.211979920195\\
4	279.93168571915\\
5	302.754577621947\\
6	319.66954938923\\
7	474.590954672879\\
8	581.889726483805\\
9	588.293598986608\\
10	665.267956492141\\
11	697.822117295295\\
12	818.073571899981\\
13	837.189685656274\\
14	951.106869086531\\
15	988.227815849574\\
16	1057.31482326398\\
17	1182.1337341983\\
18	1189.45261042673\\
19	1233.36031711783\\
20	1330.76688632399\\
21	1481.99150407976\\
22	1546.93972886159\\
23	1636.64032711421\\
24	1651.18080993083\\
25	1718.82467687182\\
26	1754.89735652813\\
27	1800.61033297745\\
28	1919.94211980459\\
29	1944.11935489724\\
30	2102.42575040076\\
};
\addlegendentry{Our 0.046};
\legend{}
\end{axis}
\end{tikzpicture}%
    \end{minipage}%
    \vspace{-0.22cm}
    \begin{minipage}{\linewidth}
        \begin{overpic}[trim=0cm -1cm 0cm 0cm,clip,width=0.7\linewidth]{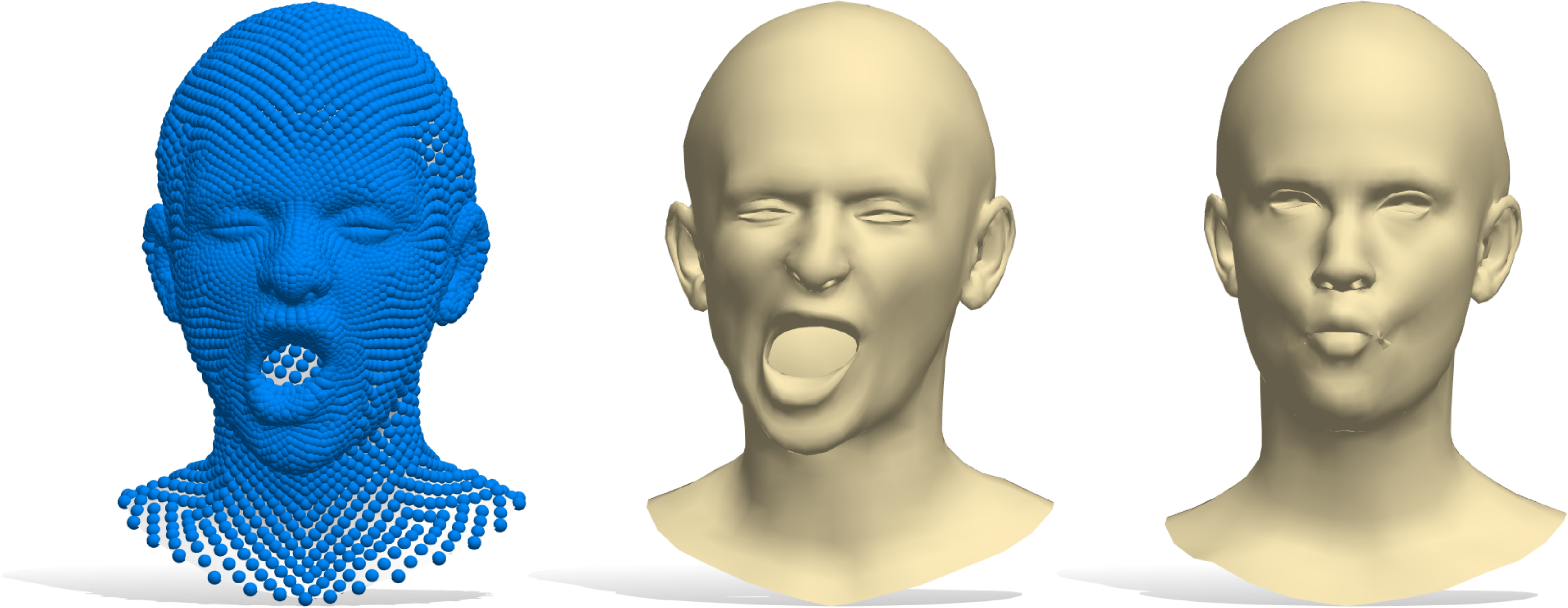}
        \end{overpic}
        \hspace{-0.1cm}
%
%
\definecolor{mycolor1}{rgb}{0.85000,0.32500,0.09800}%
\definecolor{mycolor2}{rgb}{0.4375,0.5,0.5625}%
\definecolor{mycolor3}{rgb}{0.00000,0.44700,0.74100}%
\begin{tikzpicture}

\begin{axis}[%
width=0.24\linewidth,
height=0.24\linewidth,
at={(0.758in,0.481in)},
scale only axis,
xmin=1,
xmax=30,
ymin=0,
ymax=2000,
xtick={10, 20, 30},
xticklabels={10, 20, 30},
ytick={500, 1000, 1500, 2000},
yticklabels={,  ,  },
xmajorgrids,
ymajorgrids,
every x tick label/.append style={font=\color{black}, font=\tiny},
every y tick label/.append style={font=\color{black}, font=\tiny},
axis background/.style={fill=white},
legend columns=-1,
legend style={column sep=1ex,
at={(-2.5,-0.4)},anchor=south west,legend cell align=left,align=left,draw=white!15!black}
]
\addplot [color=mycolor1,solid,line width=1.5pt]
  table[row sep=crcr]{%
1	81.7468242665779\\
2	160.001777927347\\
3	182.33287931391\\
4	268.686712949558\\
5	293.694063418884\\
6	306.317147587917\\
7	447.347967407482\\
8	537.679655672947\\
9	563.693407002373\\
10	625.754944574352\\
11	646.408606519892\\
12	832.585227488915\\
13	843.96045918703\\
14	883.273765817712\\
15	979.726436407299\\
16	996.668232583789\\
17	1094.44885489741\\
18	1120.25843883546\\
19	1204.83603657615\\
20	1239.32318514121\\
21	1377.89439647058\\
22	1470.17443466589\\
23	1600.3074225639\\
24	1627.17847075765\\
25	1675.20559032388\\
26	1683.32931359758\\
27	1721.82064555315\\
28	1807.62696604107\\
29	1821.25672145306\\
30	1888.71081429853\\
};
\addlegendentry{target};

\addplot [color=mycolor2,solid,line width=1.5pt]
  table[row sep=crcr]{%
1	77.212043762207\\
2	158.178161621094\\
3	197.479965209961\\
4	267.014862060547\\
5	291.798919677734\\
6	302.148681640625\\
7	443.291961669922\\
8	542.051574707031\\
9	576.829040527344\\
10	651.288635253906\\
11	662.307434082031\\
12	825.58056640625\\
13	846.804321289063\\
14	885.995300292969\\
15	946.667541503906\\
16	948.783203125\\
17	1112.0888671875\\
18	1121.99133300781\\
19	1198.16796875\\
20	1272.44946289063\\
21	1339.20935058594\\
22	1435.93225097656\\
23	1580.69409179688\\
24	1601.80102539063\\
25	1613.31286621094\\
26	1653.27502441406\\
27	1711.01159667969\\
28	1792.04272460938\\
29	1844.7138671875\\
30	1951.77368164063\\
};
\addlegendentry{nn 0.15};

\addplot [color=mycolor3,solid,line width=1.5pt]
  table[row sep=crcr]{%
1	81.6813308342258\\
2	159.91971554999\\
3	185.467090680361\\
4	266.947571018599\\
5	298.081072713714\\
6	306.923500413909\\
7	456.78791648224\\
8	534.451928401728\\
9	571.061680626278\\
10	648.047120295929\\
11	650.143135044084\\
12	817.105611995323\\
13	818.650022709332\\
14	876.848903738785\\
15	961.495257571485\\
16	1023.55966864794\\
17	1140.47070864975\\
18	1146.99335406701\\
19	1201.01630205652\\
20	1288.93573255515\\
21	1379.11299032747\\
22	1490.95477739617\\
23	1621.08470746941\\
24	1625.33125983291\\
25	1639.67721727738\\
26	1709.22980074173\\
27	1747.89954632932\\
28	1852.35324753132\\
29	1872.36411434254\\
30	2020.01506084318\\
};
\addlegendentry{Our 0.13};
\legend{}
\end{axis}
\end{tikzpicture}%
    \end{minipage}%

\vspace{0.1cm}
\caption{\label{fig:repr}Shape reconstruction from eigenvalues using our approach on different representations (i.e. 2D contours, 3D meshes and point clouds). The eigenvalues of the shapes on the left are given to our network, which outputs the shapes in the middle. For each representation, the eigenvalues are computed on the appropriate Laplacian discretization as per Sec.~\ref{sec:bg}. The NN column shows the nearest-neighbor solution sought in the training set.
\vspace{-2mm}}
\end{figure}
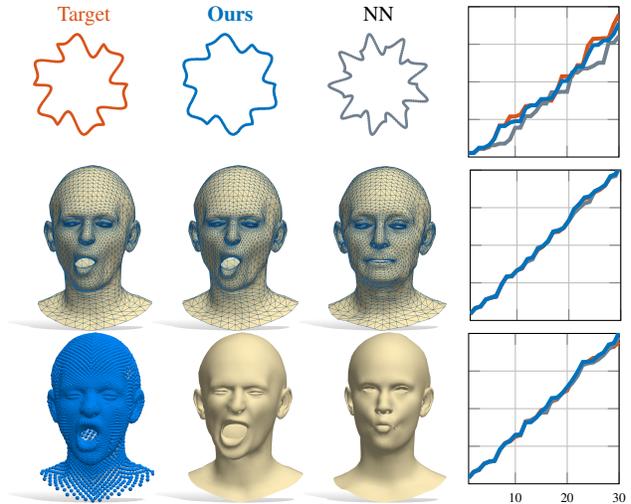
\end{center}
%

\vspace{-10mm}
\vspace{1ex}\noindent\textbf{Shape representation.}
We consider two different settings: triangle meshes in point-to-point correspondence {\em at training time} (typical in graphics and geometry processing), and unorganized point clouds {\em without} a consistent vertex labeling (typical in 3D computer vision). 

\vspace{1ex}\noindent\textbf{Autoencoder architecture.}
%
%
%
%
\new{Our model can be built with potentially any autoencoder. In our applications we chose relatively simple ones to deal with meshes and unorganized point clouds, although more powerful generative methods would be equally possible.}
The latent space dimension is fixed to $30$ (the same as $k$).
We refer to the supplementary material for details about the architecture, both in the case of meshes and point clouds. 
%


\vspace{1ex}\noindent\textbf{{\em Remark.}}
Our architecture takes $\mathrm{Spec}(\X)$ as an input, \ie, the eigenvalues are {\em not} computed at training time. By learning an {\em invertible} mapping to the latent space, we avoid expensive backpropagation steps through the spectral decomposition of the Laplacian $\Delta_\X$. In this sense, the mapping $\rho$ acts as an efficient proxy to differentiable eigendecomposition, which we exploit in several applications below.

Since eigenvalue computation is only incurred as an offline cost, it can be performed with arbitrary accuracy (we use cubic FEM, see Fig.~\ref{fig:FEM}) without sacrificing efficiency.
%
%
%
\section{Results}\label{sec:results}

In this section we report the results on our core application of shape from spectrum recovery.

To evaluate our method, we trained our model on 1,853 3D shapes from the COMA dataset~\cite{COMA} of human faces; 100 shapes of an unseen subject are used for the test set. We repeated this test at four different mesh resolutions: $\sim$4K (full resolution), 1K, 500 and 200 vertices respectively. 
For each resolution, we independently compute the Laplacian spectrum and use these spectra to recover the shape.

\vspace{1ex}\noindent\textbf{Comparison.}
We compared our method in terms of reconstruction accuracy to the state-of-the-art isospectralization method of Cosmo~\etal~\cite{isosp}, as well as to a nearest-neighbors baseline, consisting in picking the shape of the training set with the closest spectrum to the target one.
In addition, we trained two separate architectures (with and without the $\rho$ block) and compared them.
%
%
The test without this network component is an ablation study we carry out to validate the importance of the {\em invertible} module connecting the spectral encoding to the learned latent codes.

\begin{table}[]
\begin{center}
\begin{tabular}{l|llll}
          & {\small full res}        & {\small 1000}    & {\small 500}     & {\small 200}       \\ \hline
          {\small \textbf{Ours}} & {\small $\mathbf{1.61}$} &  {\small $\mathbf{1.62}$} &  {\small $\mathbf{1.71}$} &  {\small $\mathbf{2.13}$} \\
          {\small Ours without $\rho$}   &  {\small $1.89$} &  {\small $1.82$} &  {\small $2.06$} &  {\small $2.42$}  \\
          {\small NN} &  {\small $4.45$} &  {\small $4.63$} &  {\small $4.01$} &  {\small $2.65$} \\
          {\small Cosmo~\etal~\cite{isosp}} & $-$ &  {\small $16.4$} & {\small  $7.11$} &  {\small $4.08$}
\end{tabular}
\vspace{0.05cm}

\caption{\label{tab:tests}Shape-from-spectrum reconstruction comparisons with NN  (nearest neighbors between spectra) and the state of the art \cite{isosp}; we report average error over 100 shapes of an unseen subject from COMA~\cite{COMA}. Best results are obtained with our full pipeline. `$-$' denotes out of memory; all errors must be rescaled by $10^{-5}$.}
\end{center}
\end{table}

The quantitative results are reported in Table~\ref{tab:tests} as the mean squared error between the reconstructed shape and the ground-truth. Figures \ref{fig:isospec_bad} and \ref{fig:repr} further show qualitative comparisons with the different baselines involving different shape representations. In Fig.~\ref{fig:repr}, for the sake of illustration, similarly to \cite{isosp,hamiltonian}, we also include 2D contours, discretized as regular cycle graphs.

As the results suggest, the $\rho$ block both contributes to reduce the reconstruction error, and to enable novel applications (see in Sec.~\ref{sec:apps}). Note that our method achieves a significant improvement over nearest neighbors in terms of accuracy, and an order of magnitude improvement over isospectralization. Also, the latter approach consists in an expensive optimization which requires hours to run, while our method is instantaneous at test time.

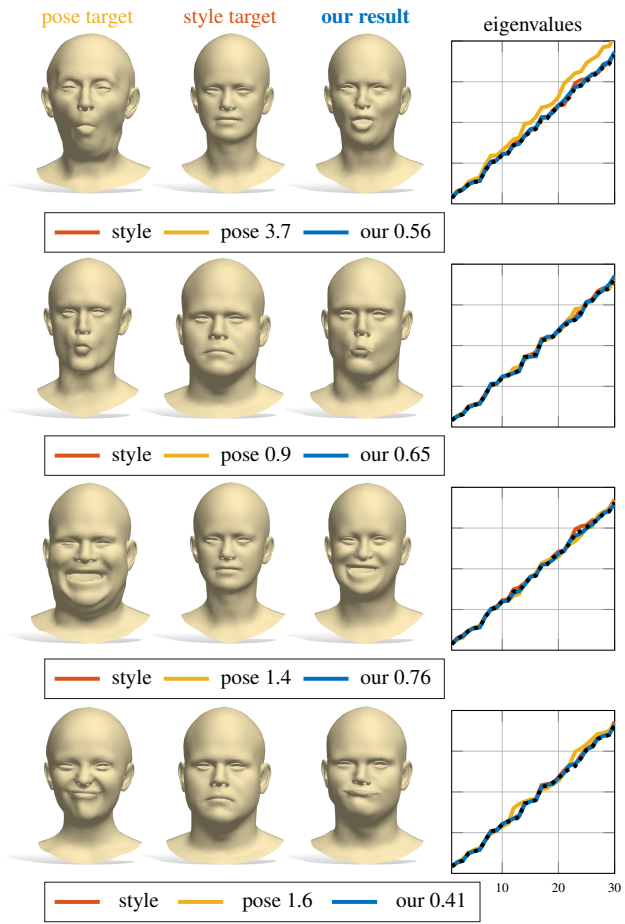
\begin{figure}[!t]
\centering
\vspace{-0.4cm}

 \setlength{\tabcolsep}{0pt}
 \begin{tabular}{l r}
 \hspace{-4.15cm}
 
    \begin{minipage}{0.74\linewidth}
\vspace{-0.4cm}

    \begin{overpic}[trim=0cm 0cm 0cm 0cm,clip,width=0.9\linewidth]{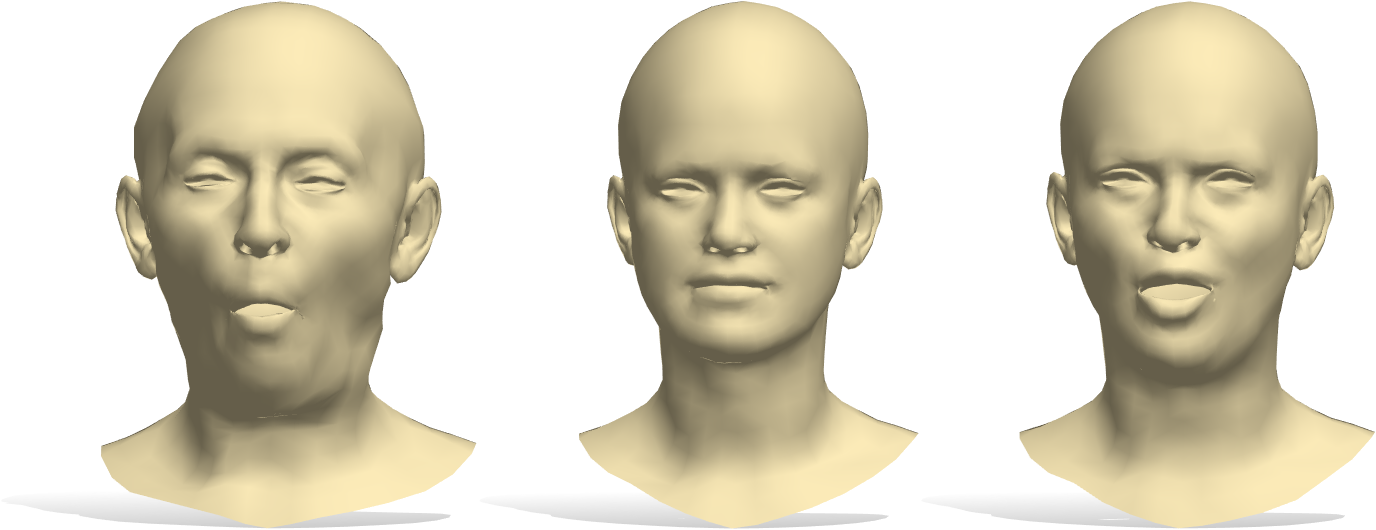}
        \put(8.5,41){\footnotesize \pose{pose target}}
        \put(42.5,41){\footnotesize \style{style target}}
        \put(75.5,41){\footnotesize \textbf{\our{our result}}}
    
  \end{overpic}

\vspace{0.85cm}

\begin{overpic}[trim=0cm 0cm 0cm 0cm,clip,width=0.9\linewidth]{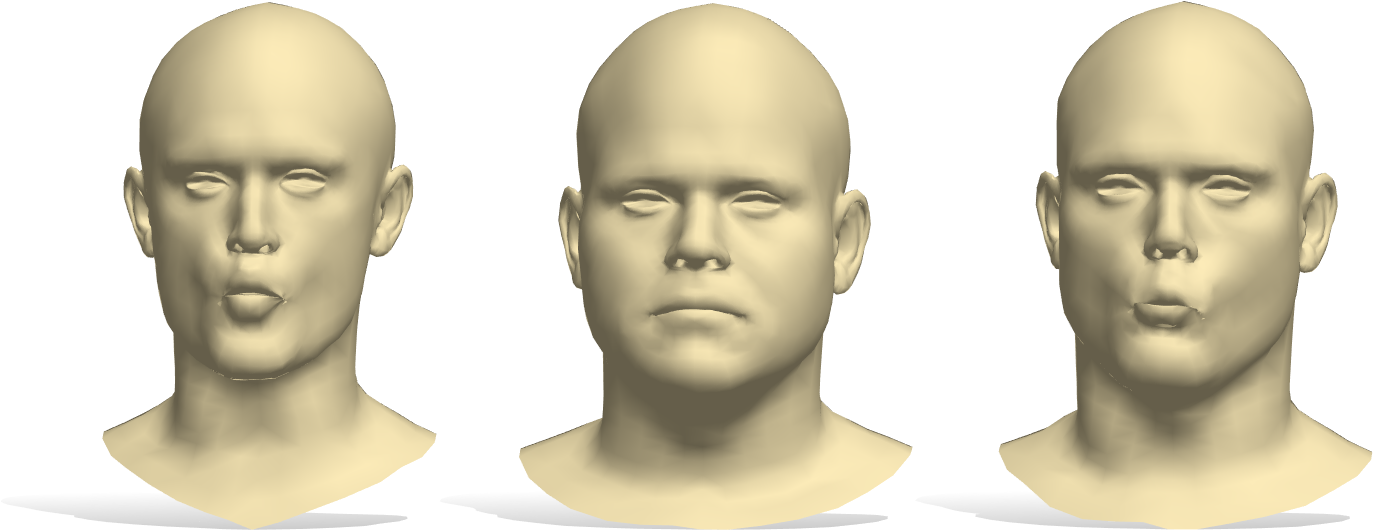}
       
  \end{overpic}
  
  \vspace{0.85cm}

     

\begin{overpic}[trim=0cm 0cm 0cm 0cm,clip,width=0.9\linewidth]{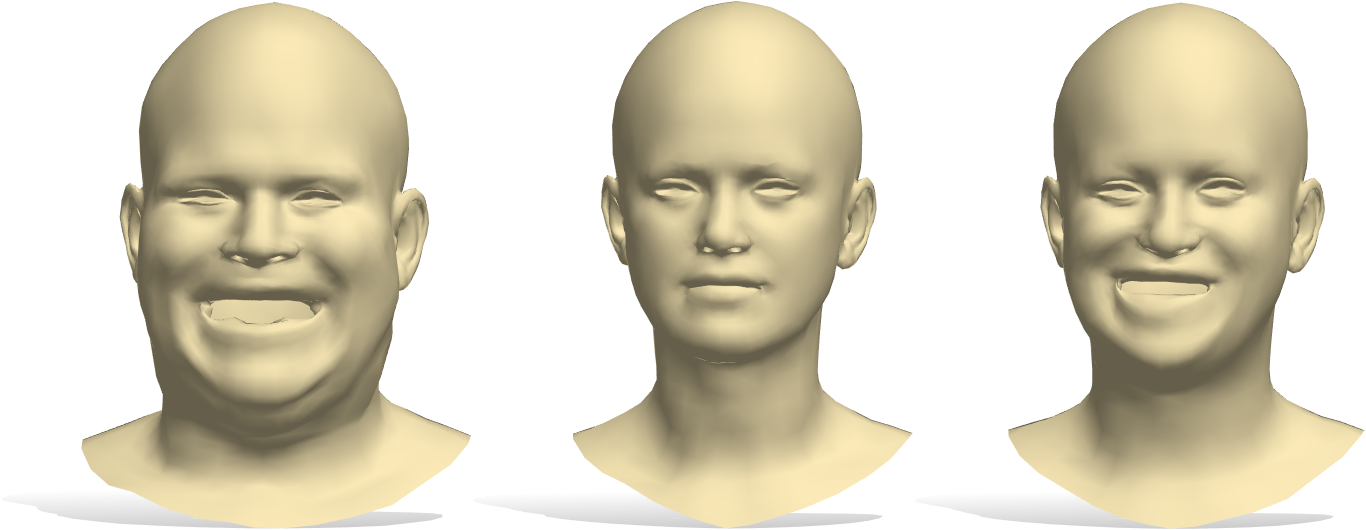}
     
  \end{overpic}

  \vspace{0.85cm}

\begin{overpic}[trim=0cm 0cm 0cm 0cm,clip,width=0.9\linewidth]{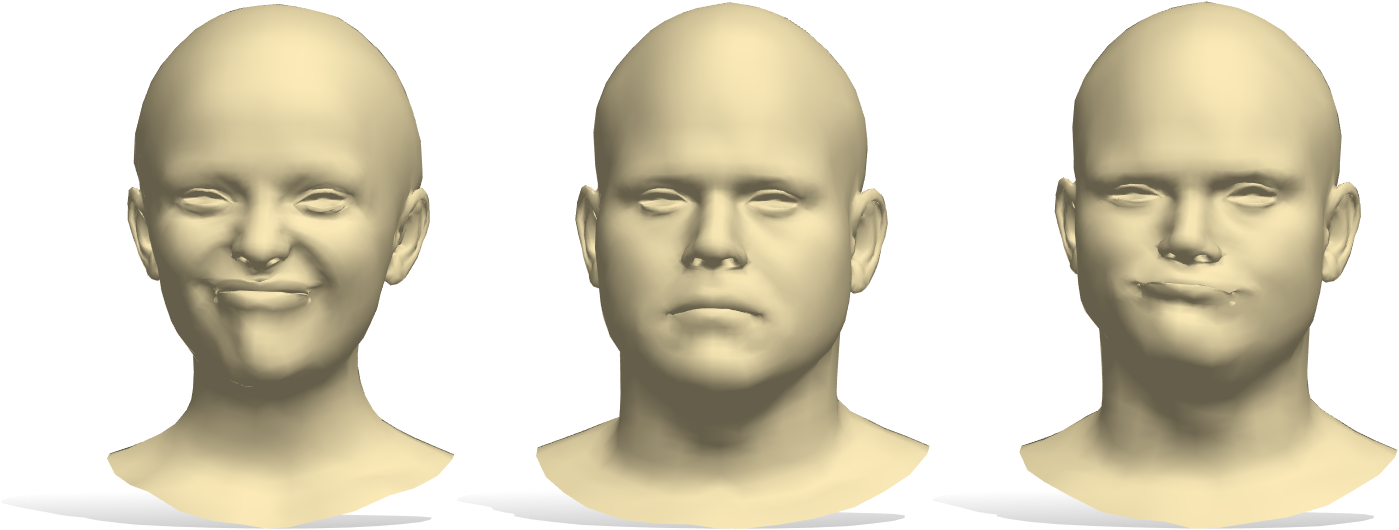}
       
  \end{overpic}

    \end{minipage}%
    &
\hspace{-5.75cm}
 
 \begin{minipage}{0.26\linewidth}
%
%
\definecolor{mycolor1}{rgb}{0.85000,0.32500,0.09800}%
\definecolor{mycolor2}{rgb}{0.92900,0.69400,0.12500}%
\definecolor{mycolor3}{rgb}{0.00000,0.44700,0.74100}%
\begin{tikzpicture}

\begin{axis}[%
width=1\linewidth,
height=\linewidth,
at={(0.758in,0.481in)},
scale only axis,
xmin=1,
xmax=30,
ymin=0,
ymax=2000,
xtick={10, 20, 30},
xticklabels={, , },
ytick={500, 1000, 1500, 2000},
yticklabels={,  ,  },
xmajorgrids,
ymajorgrids,
every x tick label/.append style={font=\color{black}, font=\tiny},
every y tick label/.append style={font=\color{black}, font=\tiny},
title={\footnotesize eigenvalues},
title style={yshift=-0.75em},
axis background/.style={fill=white},
legend columns=-1,
legend style={column sep=0.4ex,
at={(-2.5,-0.3)},anchor=south west,legend cell align=left,align=left,draw=white!15!black}
]
\addplot [color=mycolor1,solid,line width=1.5pt]
  table[row sep=crcr]{%
1	70.9235542737846\\
2	147.436062294176\\
3	184.943500103574\\
4	238.631310945706\\
5	270.236324904701\\
6	285.67486205612\\
7	414.353272027636\\
8	511.178885716692\\
9	536.383337607052\\
10	610.962817483952\\
11	622.961311688964\\
12	746.225645580279\\
13	769.998534011358\\
14	828.342954853275\\
15	901.121059523321\\
16	930.494708516844\\
17	1045.18319185844\\
18	1063.59381678575\\
19	1146.58681903968\\
20	1211.66920811416\\
21	1220.79530386579\\
22	1331.02921594494\\
23	1484.37412424063\\
24	1521.29181120117\\
25	1534.31125833918\\
26	1594.69770070404\\
27	1607.86908048197\\
28	1688.66268871575\\
29	1735.64104032083\\
30	1846.97929533746\\
};
\addlegendentry{\footnotesize style};

\addplot [color=mycolor2,solid,line width=1.5pt]
  table[row sep=crcr]{%
1	88.7378291009176\\
2	163.508146679178\\
3	183.138276160415\\
4	272.143636275263\\
5	302.106692742309\\
6	328.870489576818\\
7	480.231737270804\\
8	591.913896297595\\
9	594.583579145372\\
10	656.819586231973\\
11	718.782659929056\\
12	791.431829529083\\
13	813.363026684108\\
14	984.412949353734\\
15	995.83719021016\\
16	1068.74325342979\\
17	1183.15974739912\\
18	1197.77803407979\\
19	1233.96859144574\\
20	1314.48571093143\\
21	1479.08514174256\\
22	1540.33213489265\\
23	1606.08055530042\\
24	1623.93050140876\\
25	1731.85365067636\\
26	1778.14606125494\\
27	1821.08158513367\\
28	1925.48909406826\\
29	1937.82727111396\\
30	2054.36811377637\\
};
\addlegendentry{\footnotesize pose 3.7};

\addplot [color=mycolor3,solid,line width=1.5pt]
  table[row sep=crcr]{%
1	75.1387695322419\\
2	146.737807665859\\
3	175.733100967529\\
4	240.302394767571\\
5	273.620369448073\\
6	279.919314460181\\
7	425.776401110583\\
8	517.471864559576\\
9	536.886713143006\\
10	611.026232844199\\
11	619.981564477517\\
12	716.434333660626\\
13	736.436920753121\\
14	839.203580089967\\
15	915.458939218127\\
16	936.874696378215\\
17	1064.13385250436\\
18	1069.68476993982\\
19	1120.92593058191\\
20	1204.88099866669\\
21	1270.71290756113\\
22	1356.94159732659\\
23	1437.03570760604\\
24	1446.75628756583\\
25	1538.88674530452\\
26	1583.31920610928\\
27	1622.24842673741\\
28	1713.10862666622\\
29	1741.67201276234\\
30	1865.97282399714\\
};
\addlegendentry{\footnotesize our 0.56};

\addplot [color=black,dotted,line width=1.5pt]
  table[row sep=crcr]{%
1	73.1501998901367\\
2	144.237701416016\\
3	175.611389160156\\
4	238.284973144531\\
5	264.317413330078\\
6	284.289703369141\\
7	410.019195556641\\
8	505.585235595703\\
9	527.867553710938\\
10	599.11962890625\\
11	617.174133300781\\
12	720.756225585938\\
13	740.120971679688\\
14	822.50927734375\\
15	889.450927734375\\
16	919.438415527344\\
17	1032.05871582031\\
18	1049.93542480469\\
19	1134.94921875\\
20	1189.06640625\\
21	1243.24084472656\\
22	1311.39379882813\\
23	1428.67553710938\\
24	1456.47033691406\\
25	1507.62866210938\\
26	1569.17297363281\\
27	1606.70788574219\\
28	1680.70483398438\\
29	1716.28308105469\\
30	1825.37036132813\\
};

\end{axis}
\end{tikzpicture}%
    
%
%
\definecolor{mycolor1}{rgb}{0.85000,0.32500,0.09800}%
\definecolor{mycolor2}{rgb}{0.92900,0.69400,0.12500}%
\definecolor{mycolor3}{rgb}{0.00000,0.44700,0.74100}%
\begin{tikzpicture}

\begin{axis}[%
width=\linewidth,
height=\linewidth,
at={(0.758in,0.481in)},
scale only axis,
xmin=1,
xmax=30,
ymin=0,
ymax=2000,
xtick={10, 20, 30},
xticklabels={ , , },
ytick={500, 1000, 1500, 2000},
yticklabels={,  ,  },
xmajorgrids,
ymajorgrids,
every x tick label/.append style={font=\color{black}, font=\tiny},
every y tick label/.append style={font=\color{black}, font=\tiny},
axis background/.style={fill=white},
legend columns=-1,
legend style={column sep=0.4ex,
at={(-2.5,-0.3)},anchor=south west,legend cell align=left,align=left,draw=white!15!black}
]
\addplot [color=mycolor1,solid,line width=1.5pt]
  table[row sep=crcr]{%
1	88.3991523402342\\
2	132.877175166086\\
3	157.533858589753\\
4	254.018259526285\\
5	277.000743109669\\
6	293.626727038495\\
7	410.179936684996\\
8	519.651989988256\\
9	529.364972645713\\
10	608.429572535167\\
11	637.275530022479\\
12	674.444189589679\\
13	712.990219448453\\
14	863.66149700102\\
15	885.297316186071\\
16	917.83830134501\\
17	1082.40772865865\\
18	1095.97085519538\\
19	1110.15434484794\\
20	1179.61286353169\\
21	1261.40147031271\\
22	1334.80606661506\\
23	1343.76075356628\\
24	1447.58295048208\\
25	1533.40967717059\\
26	1570.07209606373\\
27	1667.58581976392\\
28	1705.3784390348\\
29	1714.43273379947\\
30	1861.40993421404\\
};
\addlegendentry{\footnotesize style};

\addplot [color=mycolor2,solid,line width=1.5pt]
  table[row sep=crcr]{%
1	77.837568412641\\
2	137.767452761683\\
3	171.165650226323\\
4	242.483856183389\\
5	273.680325652848\\
6	279.684403193034\\
7	392.750614408827\\
8	512.50445785702\\
9	531.346897497063\\
10	613.738306800066\\
11	624.525347179465\\
12	721.016393207625\\
13	736.196300023015\\
14	826.030532718814\\
15	874.058333176616\\
16	878.494779055351\\
17	1057.80895736137\\
18	1076.71903267607\\
19	1106.18395578841\\
20	1173.65252449723\\
21	1236.99683212043\\
22	1325.07772857825\\
23	1448.48637644823\\
24	1456.78045374554\\
25	1525.06928777969\\
26	1545.82081585922\\
27	1636.71446827039\\
28	1690.00134846893\\
29	1694.02888158604\\
30	1814.58176262219\\
};
\addlegendentry{\footnotesize pose 0.9}; 

\addplot [color=mycolor3,solid,line width=1.5pt]
  table[row sep=crcr]{%
1	82.3632780644817\\
2	132.30029996309\\
3	161.790989184955\\
4	242.38652332426\\
5	270.032230372707\\
6	287.900270009343\\
7	391.365809583266\\
8	530.978382118559\\
9	534.109385044209\\
10	618.92549881098\\
11	639.916558331831\\
12	659.913971857606\\
13	684.620635182612\\
14	854.618541256558\\
15	864.791534935415\\
16	868.77793991337\\
17	1061.2603903302\\
18	1108.64706758176\\
19	1112.01623666162\\
20	1192.49484890369\\
21	1260.53345826653\\
22	1299.89656115527\\
23	1321.24397909805\\
24	1378.05818654206\\
25	1535.28095908758\\
26	1550.13761373886\\
27	1633.2287818596\\
28	1681.71872175391\\
29	1749.07912801425\\
30	1848.24083674814\\
};
\addlegendentry{\footnotesize our 0.65};

\addplot [color=black,dotted,line width=1.5pt]
  table[row sep=crcr]{%
1	85.8543090820313\\
2	130.913925170898\\
3	156.50715637207\\
4	246.120620727539\\
5	270.960998535156\\
6	289.811828613281\\
7	400.185089111328\\
8	516.315979003906\\
9	528.979370117188\\
10	603.090515136719\\
11	634.789611816406\\
12	657.401611328125\\
13	693.221984863281\\
14	859.8046875\\
15	865.892150878906\\
16	894.774353027344\\
17	1060.79064941406\\
18	1078.85205078125\\
19	1107.50036621094\\
20	1171.25769042969\\
21	1241.70202636719\\
22	1287.54919433594\\
23	1317.81726074219\\
24	1405.14538574219\\
25	1512.44555664063\\
26	1550.85009765625\\
27	1646.43957519531\\
28	1679.025390625\\
29	1698.2265625\\
30	1827.87805175781\\
};

\end{axis}
\end{tikzpicture}%

%
%
\definecolor{mycolor1}{rgb}{0.85000,0.32500,0.09800}%
\definecolor{mycolor2}{rgb}{0.92900,0.69400,0.12500}%
\definecolor{mycolor3}{rgb}{0.00000,0.44700,0.74100}%
\begin{tikzpicture}

\begin{axis}[%
width=\linewidth,
height=\linewidth,
at={(0.758in,0.481in)},
scale only axis,
xmin=1,
xmax=30,
ymin=0,
ymax=2000,
xtick={10, 20, 30},
xticklabels={ , , },
ytick={500, 1000, 1500, 2000},
yticklabels={,  ,  },
xmajorgrids,
ymajorgrids,
every x tick label/.append style={font=\color{black}, font=\tiny},
every y tick label/.append style={font=\color{black}, font=\tiny},
axis background/.style={fill=white},
legend columns=-1,
legend style={column sep=0.4ex,
at={(-2.5,-0.3)},anchor=south west,legend cell align=left,align=left,draw=white!15!black}
]
\addplot [color=mycolor1,solid,line width=1.5pt]
  table[row sep=crcr]{%
1	70.9235542737846\\
2	147.436062294176\\
3	184.943500103574\\
4	238.631310945706\\
5	270.236324904701\\
6	285.67486205612\\
7	414.353272027636\\
8	511.178885716692\\
9	536.383337607052\\
10	610.962817483952\\
11	622.961311688964\\
12	746.225645580279\\
13	769.998534011358\\
14	828.342954853275\\
15	901.121059523321\\
16	930.494708516844\\
17	1045.18319185844\\
18	1063.59381678575\\
19	1146.58681903968\\
20	1211.66920811416\\
21	1220.79530386579\\
22	1331.02921594494\\
23	1484.37412424063\\
24	1521.29181120117\\
25	1534.31125833918\\
26	1594.69770070404\\
27	1607.86908048197\\
28	1688.66268871575\\
29	1735.64104032083\\
30	1846.97929533746\\
};
\addlegendentry{\footnotesize style};

\addplot [color=mycolor2,solid,line width=1.5pt]
  table[row sep=crcr]{%
1	86.6308618593242\\
2	131.280512137045\\
3	154.978014591937\\
4	237.94942860164\\
5	280.978403661111\\
6	297.204473719039\\
7	417.969435014621\\
8	498.970839604211\\
9	515.690328558454\\
10	595.519700058097\\
11	605.710124833354\\
12	666.203837127314\\
13	689.838808752756\\
14	820.2641508014\\
15	889.32443797106\\
16	947.738865733597\\
17	1067.3516065659\\
18	1101.03095990987\\
19	1118.80107164776\\
20	1146.48382484362\\
21	1231.68100446436\\
22	1300.63210380728\\
23	1340.31215678275\\
24	1408.06802644794\\
25	1506.36007191323\\
26	1515.64328542532\\
27	1652.10967505449\\
28	1700.88668058532\\
29	1718.41873928453\\
30	1832.2195408963\\
};
\addlegendentry{\footnotesize pose 1.4};

\addplot [color=mycolor3,solid,line width=1.5pt]
  table[row sep=crcr]{%
1	76.5103611879836\\
2	141.192383660605\\
3	171.295720416137\\
4	237.619901775739\\
5	273.362851031173\\
6	279.062620821916\\
7	415.762649883029\\
8	501.320306953945\\
9	523.313751008843\\
10	603.561899846347\\
11	605.442803161204\\
12	704.135720316171\\
13	725.108017755053\\
14	802.871447331285\\
15	899.854168307903\\
16	921.760205284582\\
17	1043.98924763377\\
18	1063.99486685333\\
19	1107.7530015362\\
20	1184.71067576321\\
21	1216.37155815945\\
22	1338.62354624892\\
23	1404.9051551712\\
24	1435.98399250706\\
25	1493.30664864687\\
26	1542.98727345136\\
27	1613.05935857664\\
28	1682.72033718681\\
29	1705.3338035537\\
30	1799.94893961672\\
};
\addlegendentry{\footnotesize our 0.76};

\addplot [color=black,dotted,line width=1.5pt]
  table[row sep=crcr]{%
1	73.0873794555664\\
2	143.035659790039\\
3	175.984176635742\\
4	237.78337097168\\
5	265.330169677734\\
6	283.674926757813\\
7	408.456481933594\\
8	499.192993164063\\
9	528.542114257813\\
10	601.162719726563\\
11	614.808898925781\\
12	720.045471191406\\
13	735.386474609375\\
14	808.244567871094\\
15	891.824645996094\\
16	915.676086425781\\
17	1036.01953125\\
18	1055.21936035156\\
19	1133.38366699219\\
20	1184.09118652344\\
21	1225.47436523438\\
22	1312.64245605469\\
23	1430.8017578125\\
24	1464.98498535156\\
25	1509.36560058594\\
26	1566.0703125\\
27	1596.94018554688\\
28	1678.20068359375\\
29	1702.21472167969\\
30	1815.94763183594\\
};

\end{axis}
\end{tikzpicture}%

%
%
\definecolor{mycolor1}{rgb}{0.85000,0.32500,0.09800}%
\definecolor{mycolor2}{rgb}{0.92900,0.69400,0.12500}%
\definecolor{mycolor3}{rgb}{0.00000,0.44700,0.74100}%
\begin{tikzpicture}

\begin{axis}[%
width=\linewidth,
height=\linewidth,
at={(0.758in,0.481in)},
scale only axis,
xmin=1,
xmax=30,
ymin=0,
ymax=2000,
xtick={10, 20, 30},
xticklabels={10, 20, 30},
ytick={500, 1000, 1500, 2000},
yticklabels={,  ,  },
xmajorgrids,
ymajorgrids,
every x tick label/.append style={font=\color{black}, font=\tiny},
every y tick label/.append style={font=\color{black}, font=\tiny},
axis background/.style={fill=white},
legend columns=-1,
legend style={column sep=1ex,
at={(-2.5,-0.3)},anchor=south west,legend cell align=left,align=left,draw=white!15!black}
]
\addplot [color=mycolor1,solid,line width=1.5pt]
  table[row sep=crcr]{%
1	88.3991523402342\\
2	132.877175166086\\
3	157.533858589753\\
4	254.018259526285\\
5	277.000743109669\\
6	293.626727038495\\
7	410.179936684996\\
8	519.651989988256\\
9	529.364972645713\\
10	608.429572535167\\
11	637.275530022479\\
12	674.444189589679\\
13	712.990219448453\\
14	863.66149700102\\
15	885.297316186071\\
16	917.83830134501\\
17	1082.40772865865\\
18	1095.97085519538\\
19	1110.15434484794\\
20	1179.61286353169\\
21	1261.40147031271\\
22	1334.80606661506\\
23	1343.76075356628\\
24	1447.58295048208\\
25	1533.40967717059\\
26	1570.07209606373\\
27	1667.58581976392\\
28	1705.3784390348\\
29	1714.43273379947\\
30	1861.40993421404\\
};
\addlegendentry{\footnotesize style};

\addplot [color=mycolor2,solid,line width=1.5pt]
  table[row sep=crcr]{%
1	76.9590454002175\\
2	143.277887832105\\
3	177.205757874137\\
4	251.760753567211\\
5	288.63798918366\\
6	308.844661519225\\
7	412.142381883317\\
8	487.113000424898\\
9	519.78136892596\\
10	576.763310380284\\
11	606.396566207865\\
12	800.506764029323\\
13	851.940482599876\\
14	883.04951141771\\
15	907.025469008872\\
16	912.718238775442\\
17	1029.53812452443\\
18	1046.9270120251\\
19	1096.5945315683\\
20	1135.82195218613\\
21	1222.26037998677\\
22	1334.36887102476\\
23	1494.83140754792\\
24	1533.75957384106\\
25	1585.38048043316\\
26	1648.27811286271\\
27	1713.02856193456\\
28	1723.43055278928\\
29	1750.1703862155\\
30	1829.56149084174\\
};
\addlegendentry{\footnotesize pose 1.6};

\addplot [color=mycolor3,solid,line width=1.5pt]
  table[row sep=crcr]{%
1	83.740325281962\\
2	133.942561523856\\
3	160.933094496405\\
4	244.140510550459\\
5	271.284488311698\\
6	288.493137953973\\
7	406.721655560963\\
8	517.012677147384\\
9	528.477709984352\\
10	601.865960283388\\
11	641.962591100306\\
12	671.035186754101\\
13	700.22238038576\\
14	850.14577219832\\
15	883.270744622515\\
16	897.803109760493\\
17	1061.85745583939\\
18	1091.33425388779\\
19	1096.29181554601\\
20	1170.66169278862\\
21	1236.31783558386\\
22	1326.50904102621\\
23	1343.97815389714\\
24	1410.42078481259\\
25	1523.73672007127\\
26	1550.02716497271\\
27	1656.15987587799\\
28	1687.55311540196\\
29	1700.59688278773\\
30	1827.69697869886\\
};
\addlegendentry{\footnotesize  our 0.41};

\addplot [color=black,dotted,line width=1.5pt]
  table[row sep=crcr]{%
1	85.8610763549805\\
2	131.121932983398\\
3	156.214477539063\\
4	244.282943725586\\
5	269.728729248047\\
6	293.384368896484\\
7	405.604431152344\\
8	517.650695800781\\
9	527.174865722656\\
10	601.577880859375\\
11	632.419189453125\\
12	658.095092773438\\
13	691.782470703125\\
14	860.320068359375\\
15	875.055847167969\\
16	903.124694824219\\
17	1058.85827636719\\
18	1079.548828125\\
19	1113.32507324219\\
20	1171.01586914063\\
21	1235.77673339844\\
22	1285.47814941406\\
23	1318.94274902344\\
24	1415.14318847656\\
25	1514.57739257813\\
26	1549.19006347656\\
27	1647.51525878906\\
28	1677.83190917969\\
29	1699.48571777344\\
30	1823.00317382813\\
};

\end{axis}
\end{tikzpicture}%
    \end{minipage}%
 \end{tabular}
\vspace{0.2cm}

\caption{\label{fig:style}Examples of style transfer. The target style (middle) is applied to the target pose (left) by solving problem~\eqref{eq:style} and then decoding the resulting latent vector (right). For each example we also report the corresponding eigenvalue alignment (rightmost plots). The black dotted line is the image of $\rho$. The numbers in the legend denote the distance from the target ``style'' spectrum to the source pose and to our generated shape; a small number suggests near-isometry between the generated shape and the style target. }
\end{figure}

\vspace{1ex}\noindent\textbf{Spectral bandwidth} has a direct effect on reconstruction accuracy, since increasing this number brings more high-frequency detail into the representation. Following  \cite{isosp,hamiltonian,roufosse2019unsupervised}, in all our experiments we use $k=30$. In the supplementary material we report results for different $k$.

\section{Additional applications}\label{sec:apps}
\vspace{-0.1cm}

Our general model enables several additional applications, by exploiting the connection between spectral properties and shape generation. Due to the limited space, we collect in the supplementary materials the details of the training and test sets and the parameters used in our experiments.
%

%
%

\subsection{Style transfer}\label{sec:style}
As shown in Fig.~\ref{fig:teaser}, we can use our trained network to transfer the style of a shape $\X_\mathrm{style}$ to another shape $\X_\mathrm{pose}$ having both a different style and pose. This is done by a search in the latent space, phrased as:
\begin{align}
    \min_{\mathbf{v}} \| \mathrm{Spec}(\X_\mathrm{style}) \hspace{-0.05cm} - \hspace{-0.05cm} \rho(\mathbf{v}) \|_2^2 + w \| \mathbf{v} \hspace{-0.05cm} - \hspace{-0.05cm} E(\X_\mathrm{pose}) \|_2^2 \label{eq:style}
\end{align}

\begin{figure}[t!]
\begin{center}
\vspace{0.33cm}

  \begin{overpic}
  [trim=0cm 0cm 0cm 0cm,clip,width=0.85\linewidth]{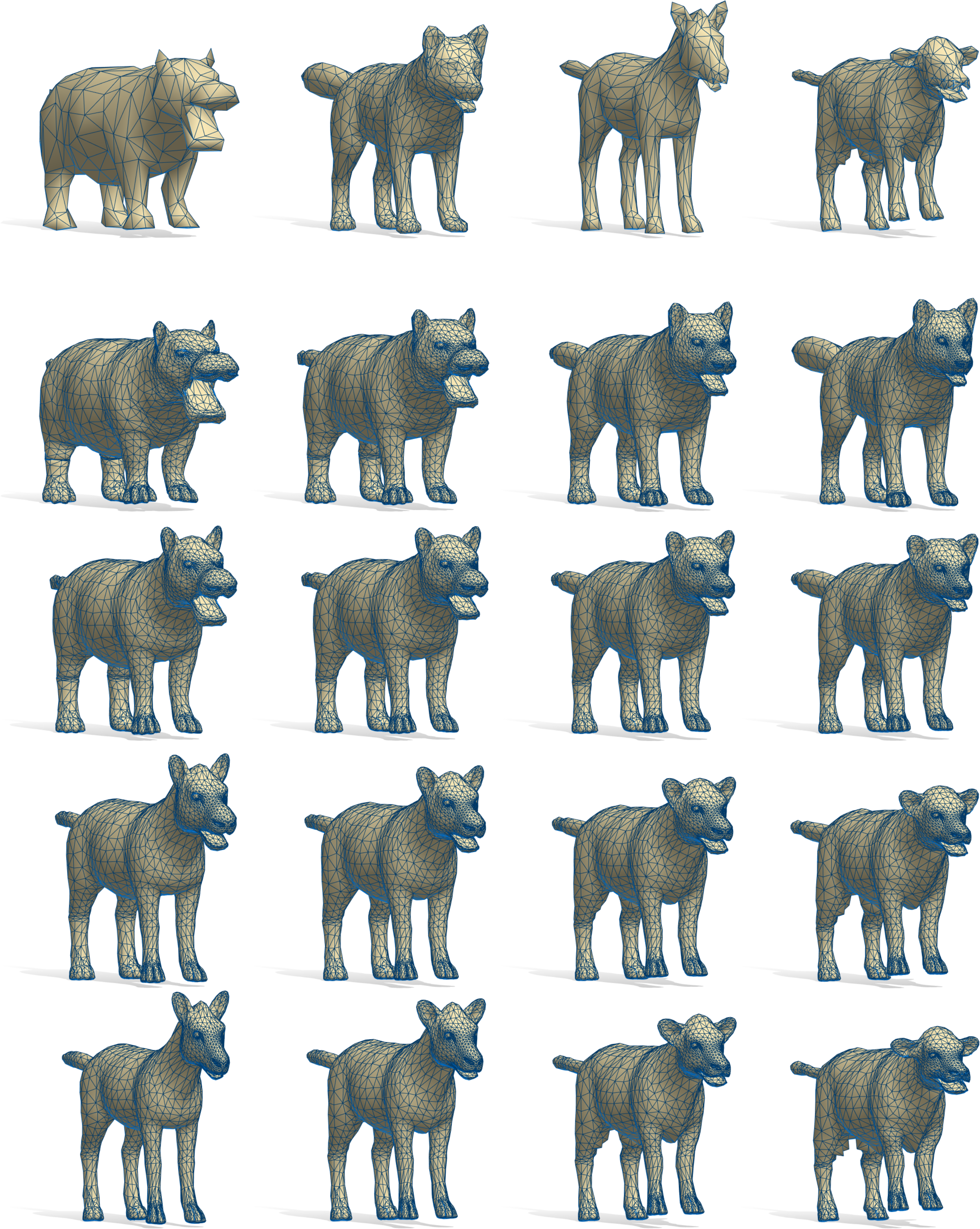}
  \put(27,101){\footnotesize \textbf{Input: low resolution shapes}}
  \put(26,76){\footnotesize \textbf{Interpolation of latent vectors}}
  \end{overpic}
  \caption{\label{fig:interp}Latent space interpolation of four low-resolution shapes with different connectivity (top row, unseen at training). The spectra of the input shapes are mapped via $\pi$ to the latent space, where they are bilinearly interpolated and then decoded to $\mathbb{R}^3$. The reconstructions of the input are depicted at the corners of the grid.\vspace{-2mm}}
  \end{center}
\end{figure}

Here, the first term seeks a latent vector whose associated spectrum aligns with the eigenvalues of $\X_\mathrm{style}$; in other words, we regard style as an intrinsic property of the shape, and exploit the fact that the Laplacian spectrum is invariant to pose deformations. The second term keeps the latent vector close to that of the input pose (we initialize with $\mathbf{v}_\mathrm{init}=E(\X_\mathrm{pose})$). \new{We solve the optimization problem by back-propagating the gradient of the cost function of Eq.~\eqref{eq:style} with respect to $\mathbf{v}$ through $\rho$.}
The sought shape is then given by a forward pass on the resulting minimizer.
In Fig.~\ref{fig:style}, we show four examples \new{(others can be found in the supplementary material)}. 
%
%
We emphasize here that the style is purely encoded in the input eigenvalues, therefore it does not rely on the test shapes being in point-to-point correspondence with the training set. This leads to the following:

\vspace{-1ex}
\begin{property}
Our method can be used in a \textbf{correspondence-free} scenario. By taking eigenvalues as input, it enables applications that traditionally require a correspondence, but side-steps this requirement.
\end{property}

\vspace{-1ex}
This observation was also mentioned in other spectrum-based approaches \cite{isosp,hamiltonian}. However, the data-driven nature of our method makes it more robust, efficient and accurate, therefore greatly improving its practical utility.

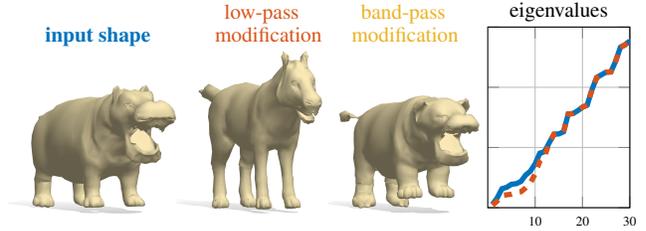
\begin{figure}[t]
\centering
\begin{minipage}{1.03\linewidth}
    \begin{overpic}[trim=0cm -2.5cm 0cm 0cm,clip,width=0.73\linewidth]{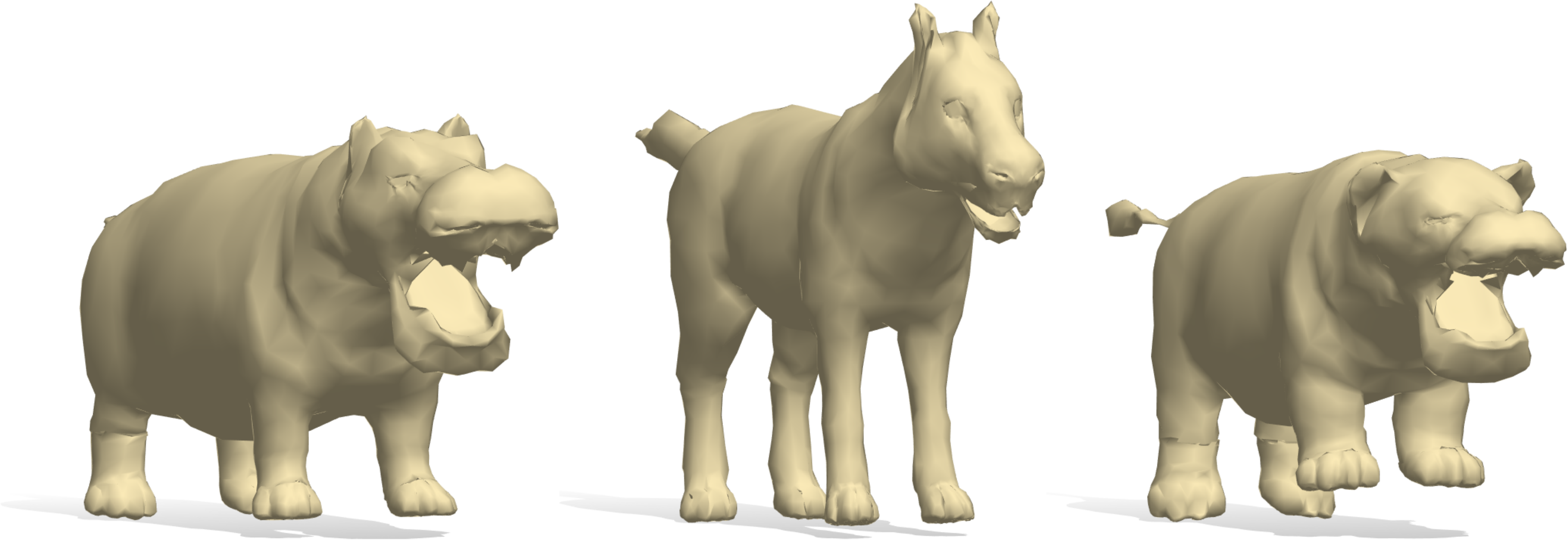}
        \put(8,41){\footnotesize \textbf{\our{input shape}}}
        \put(46,46){\footnotesize \style{low-pass}}
        \put(44,41){\footnotesize \style{modification}}
        \put(75,46){\footnotesize \pose{band-pass}}
        \put(73,41){\footnotesize \pose{modification}}
    \end{overpic}
    \hspace{-0.3cm}
%
%
\definecolor{mycolor1}{rgb}{0.00000,0.44700,0.74100}%
\definecolor{mycolor2}{rgb}{0.85000,0.32500,0.09800}%
\definecolor{mycolor3}{rgb}{0.92900,0.69400,0.12500}%
\begin{tikzpicture}

\begin{axis}[%
width=0.22\linewidth,
height=0.28\linewidth,
at={(0.758in,0.481in)},
scale only axis,
xmin=0,
xmax=30,
ymin=0,
ymax=400,
xtick={10, 20, 30},
xticklabels={ 10, 20, 30},
ytick={0, 133.3333, 266.6666, 400},
yticklabels={,  ,  },
xmajorgrids,
ymajorgrids,
every x tick label/.append style={font=\color{black}, font=\tiny},
every y tick label/.append style={font=\color{black}, font=\tiny},
title={\footnotesize eigenvalues},
title style={yshift=-0.75em},
axis background/.style={fill=white},
legend style={legend cell align=left,align=left,draw=white!15!black}
]
\addplot [color=mycolor1,solid,line width=2.0pt]
  table[row sep=crcr]{%
1	7.37898445129395\\
2	21.9582252502441\\
3	42.0739288330078\\
4	45.1457290649414\\
5	51.9622993469238\\
6	53.1369514465332\\
7	58.022705078125\\
8	72.1887054443359\\
9	80.8922500610352\\
10	97.2623291015625\\
11	120.760025024414\\
12	123.989433288574\\
13	144.22412109375\\
14	162.674057006836\\
15	163.391052246094\\
16	167.98356628418\\
17	207.54833984375\\
18	208.394897460938\\
19	214.714126586914\\
20	222.51301574707\\
21	227.165313720703\\
22	261.218811035156\\
23	288.130096435547\\
24	294.3681640625\\
25	300.084747314453\\
26	300.589080810547\\
27	324.27001953125\\
28	355.186553955078\\
29	360.746826171875\\
30	369.354614257813\\
};
\addlegendentry{initial};

\addplot [color=mycolor2,dashed,line width=2.0pt]
  table[row sep=crcr]{%
1	5.33413884\\
2	14.77864192\\
3	27.36219334\\
4	28.50442698\\
5	29.99401624\\
6	33.44427288\\
7	35.20821661\\
8	43.85499265\\
9	54.6418985\\
10	74.69792434\\
11	103.48415263\\
12	113.60467882\\
13	144.22412109\\
14	162.67405701\\
15	163.39105225\\
16	167.98356628\\
17	207.54833984\\
18	208.39489746\\
19	214.71412659\\
20	222.51301575\\
21	227.16531372\\
22	261.21881104\\
23	288.13009644\\
24	294.36816406\\
25	300.08474731\\
26	300.58908081\\
27	324.27001953\\
28	355.18655396\\
29	360.74682617\\
30	369.35461426\\
};
\addlegendentry{low-pass deformation};

\legend{}
\end{axis}
\end{tikzpicture}%
\end{minipage}
\caption{\label{fig:freq}Exploring the space of shapes in real time via manipulation of the spectrum. The low-pass modification (middle) decreases the first 12 eigenvalues of the input shape; the band-pass modification (right) amplifies the last 12 eigenvalues. The damping of low eigenvalues leads to more pronounced geometric features (e.g. longer legs and snout), while amplification of mid-range eigenvalues affects the high-frequency details (e.g. the ears and fingers); see the supplementary video for a wall-clock demo.
}
\end{figure}

%


\subsection{Shape exploration}

%
The results of Sec.~\ref{sec:style} suggest that eigenvalues can be used to drive the exploration of the AE's latent space toward a desired direction.
Another possibility is to regard {\em the eigenvalues themselves} as a parametric model for isometry classes, and explore the ``space of spectra'' as is typically done with latent spaces. 
Our bi-directional coupling between spectra and latent codes makes this exploration feasible, as remarked by the following property:
\vspace{-0.1cm}

\begin{property}
Latent space connections provide both a means for \textbf{controlling} the latent space, and vice-versa, enable \textbf{exploration} of the space of Laplacian spectra.
\end{property}
\vspace{-0.1cm}

Since eigenvalues change continuously with the manifold metric~\cite{bando1983generic}, a small variation in the spectrum will give rise to a small change in the geometry. We can visualize such variations in shape directly, by first deforming a given spectrum (\eg, by a simple linear interpolation between two spectra) to obtain the new eigenvalue sequence $\bm{\mu}$, and then directly computing $D(\pi(\bm{\mu}))$.

In Fig.~\ref{fig:interp} we show a related experiment. Here we train the network on 4,430 animal meshes generated with the SMAL parametric model following the official protocol~\cite{SMAL}. Given four {\em low-resolution} shapes $\X_i$ as input, we first compute their spectra $\mathrm{Spec}(\X_i)$, map these to the latent space via $\pi(\mathrm{Spec}(\X_i))$, perform a bilinear interpolation of the resulting latent vectors, and finally reconstruct the corresponding shapes.
Finally, in Fig.~\ref{fig:freq} we show an example of interactive spectrum-driven shape exploration. Given a shape and its Laplacian eigenvalues as input, we  navigate the space of shapes by directly modifying different frequency bands with the aid of a simple user interface. The modified spectra are then decoded by our network in {\em real time}. The interactive nature of this application is enabled by the efficiency of our shape from spectrum recovery (obtained in a single forward pass) and would not be possible with previous methods \cite{isosp} that rely on costly test-time optimization. We refer to the accompanying video and the supplementary materials for additional illustrations.

\subsection{Super-resolution}
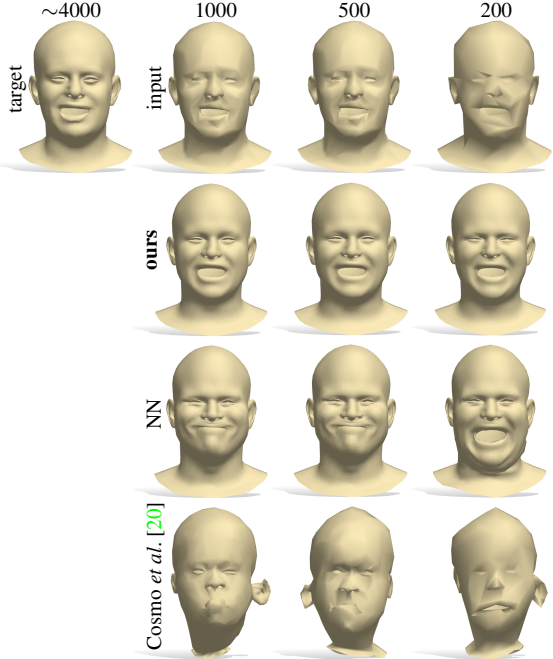
\begin{figure}[!t]
\centering
\vspace{0.3cm}

  \begin{overpic}[trim=0cm 0cm 0cm 0cm,clip,width=0.89\linewidth]{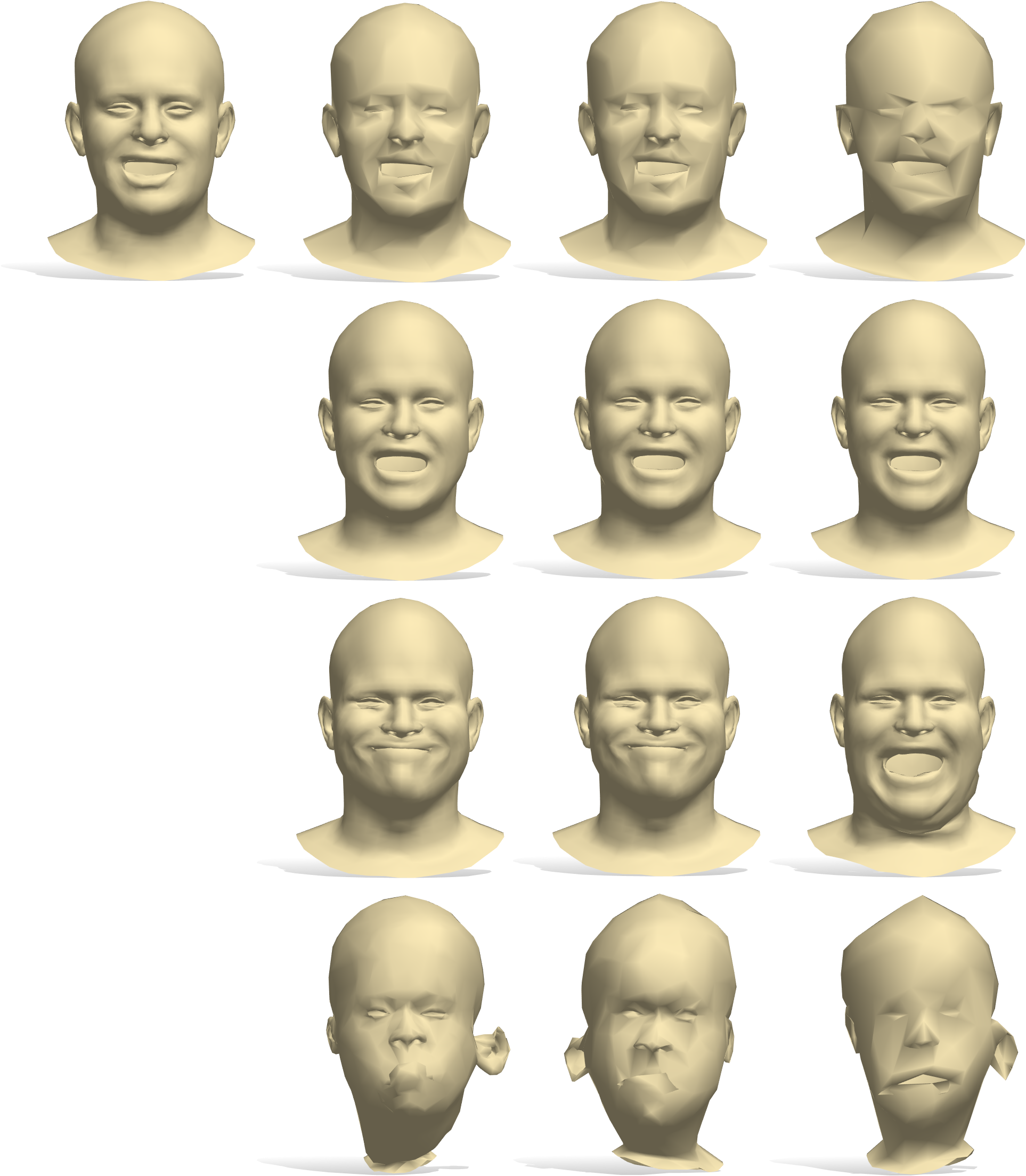}
    \put(2.5,85.5){\rotatebox{90}{\footnotesize target}}
    \put(7.5,100.5){\footnotesize $\sim$4000}
    \put(31.5,100.5){\footnotesize 1000}
    \put(54,100.5){\footnotesize 500}
    \put(76.2,100.5){\footnotesize 200}
    \put(24,86){\rotatebox{90}{\footnotesize input}}
    \put(24,1.5){\rotatebox{90}{\footnotesize Cosmo~\etal~\cite{isosp}}}
    \put(24,36){\rotatebox{90}{\footnotesize NN}}
    \put(24,61){\rotatebox{90}{\footnotesize \textbf{ours}}}
  \end{overpic}

\caption{\label{fig:different_resolution}Mesh super-resolution for inputs at decreasing resolution (top row, left to right). Our method fits closely the original input shapes (top left), while other approaches either predict the wrong pose (NN baseline) or generate an unrealistic shape (Cosmo~\etal).}
\end{figure}

A key feature that emerges from the experiment in Fig.~\ref{fig:interp} is the perfect reconstruction of the low-resolution shapes once their eigenvalues are mapped to the latent space via $\pi$. This brings us to a fundamental property of our approach:

\vspace{-1ex}
\begin{property}
Since eigenvalues are largely \textbf{insensitive to mesh resolution and sampling}, so is our trained network.
\end{property}

\vspace{-1ex}
This fact is especially evident when using cubic FEM discretization, as we do in all our tests, since it more closely approximates the continuous setting and is thus much less affected by the surface discretization.

\vspace{1ex}\noindent\textbf{\em Remark.} It is worth mentioning that existing methods can employ cubic FEM as well; however, this soon becomes prohibitively expensive due to the differentiation of spectral decomposition required by their optimizations \cite{isosp,hamiltonian}.



\definecolor{mycolor4}{rgb}{0.00000,0.44700,0.74100}%
\definecolor{mycolor5}{rgb}{0.85000,0.32500,0.09800}%
\definecolor{mycolor3}{rgb}{0.92900,0.69400,0.12500}%
\definecolor{mycolor1}{rgb}{0.49400,0.18400,0.55600}%
\definecolor{mycolor2}{rgb}{0.46600,0.67400,0.18800}%

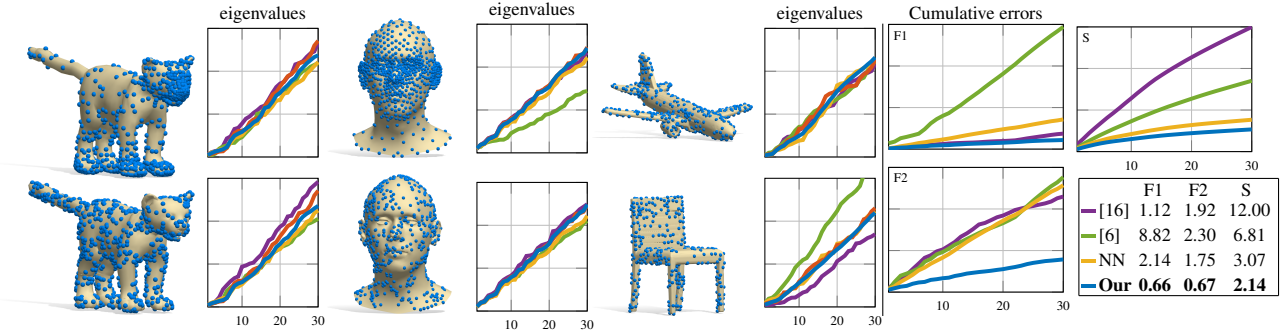
\begin{figure*}[!t]
\centering
\setlength{\tabcolsep}{0pt}
\begin{tabular}{l c c c c c c r}
\hspace{-0.4cm}

    \begin{minipage}{0.165\linewidth}   
    \vspace{0.0cm}        

        \vspace{0.2cm}
        
        \begin{overpic}[trim=0cm 0cm 0cm 0cm,clip,width=0.95\linewidth]{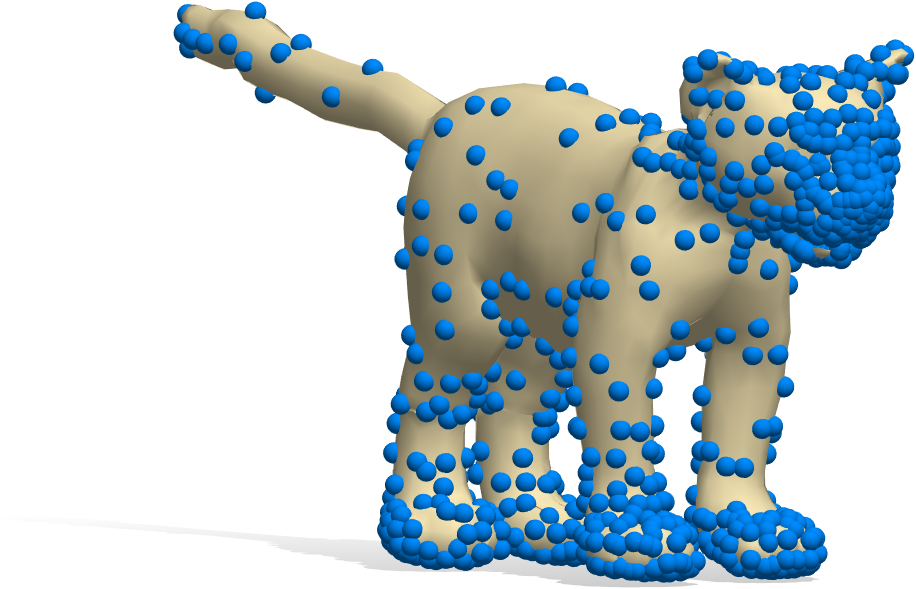}
        \end{overpic}
        
        \vspace{0.0cm}
        
        \begin{overpic}[trim=0cm 0cm 0cm 0cm,clip,width=0.95\linewidth]{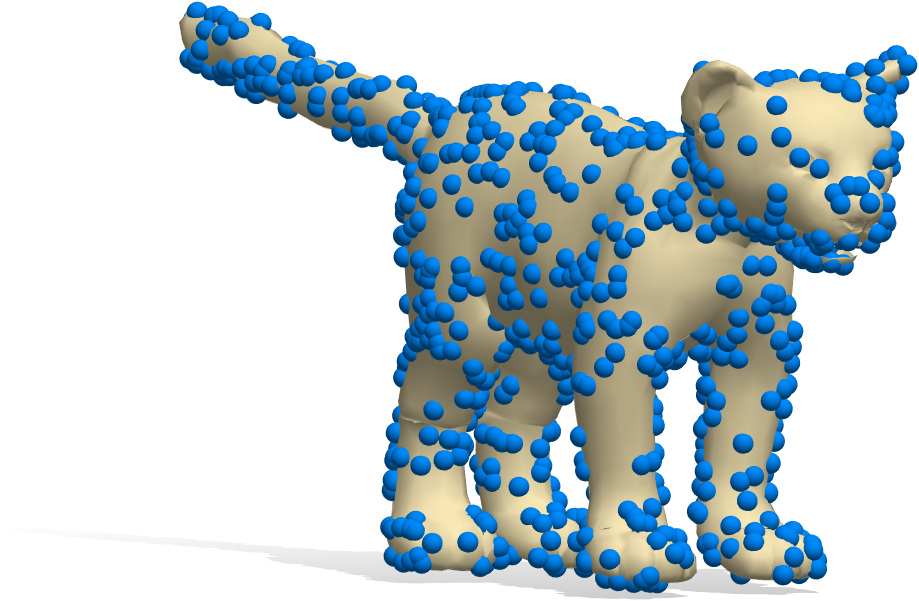}
        \end{overpic}
    
        \vspace{1.8cm}

    \end{minipage}%

    &
    
    \hspace{-0.3cm}
    
    \begin{minipage}{0.14\linewidth}     
    \vspace{-2cm}   
        
%
%
\definecolor{mycolor4}{rgb}{0.00000,0.44700,0.74100}%
\definecolor{mycolor5}{rgb}{0.85000,0.32500,0.09800}%
\definecolor{mycolor3}{rgb}{0.92900,0.69400,0.12500}%
\definecolor{mycolor1}{rgb}{0.49400,0.18400,0.55600}%
\definecolor{mycolor2}{rgb}{0.46600,0.67400,0.18800}%
\begin{tikzpicture}

\begin{axis}[%
width=0.6\linewidth,
height=0.7\linewidth,
at={(0.758in,0.481in)},
scale only axis,
xmin=1,
xmax=30,
ymin=0,
ymax=450,
xtick={10, 20, 30},
xticklabels={ , , },
ytick={0, 150, 300, 450},
yticklabels={,  ,  },
xmajorgrids,
ymajorgrids,
every x tick label/.append style={font=\color{black}, font=\tiny},
every y tick label/.append style={font=\color{black}, font=\tiny},
title={\scriptsize eigenvalues},
title style={yshift=-0.75em},
axis background/.style={fill=white},
legend columns=-1,
legend style={column sep=0.4ex,at={(-10,0)}, legend cell align=left,align=left,draw=white!15!black, anchor=north west}
]
\addplot [color=mycolor1,solid,line width=1.5pt]
  table[row sep=crcr]{%
1	8.83655984975501\\
2	16.887471188897\\
3	20.9309360520312\\
4	24.2805528411727\\
5	38.7429899610807\\
6	64.9887161910212\\
7	71.9729840794523\\
8	97.9751790390073\\
9	102.022495922192\\
10	117.114177989232\\
11	133.996415155265\\
12	141.824055000121\\
13	154.080503261979\\
14	163.634168372756\\
15	175.211424470613\\
16	180.102826258908\\
17	201.632855606516\\
18	224.991274794123\\
19	244.89082672274\\
20	256.945841329575\\
21	262.387442807982\\
22	280.607668040615\\
23	290.120065432639\\
24	323.910523151011\\
25	328.395550598792\\
26	335.178014074253\\
27	350.410552248559\\
28	362.710510068445\\
29	369.177250309392\\
30	398.268009735001\\
};
\addlegendentry{LBpcl: 6.0052};

\addplot [color=mycolor2,solid,line width=1.5pt]
  table[row sep=crcr]{%
1	0.719462717827213\\
2	5.67286386640713\\
3	12.0388078227936\\
4	29.1391546902907\\
5	36.4458660005631\\
6	40.301597972764\\
7	48.8009366410366\\
8	49.6286191647832\\
9	62.6379912981646\\
10	69.56637078478\\
11	84.3795990028258\\
12	97.8949930752274\\
13	113.340812560449\\
14	151.089919656174\\
15	154.275211455901\\
16	172.620438309728\\
17	179.51841274109\\
18	201.051166658605\\
19	217.16603708976\\
20	223.907941291992\\
21	225.136703796275\\
22	234.512837206608\\
23	250.820541198101\\
24	261.641264446979\\
25	287.503956548139\\
26	300.454796138961\\
27	304.672565465316\\
28	315.919902853056\\
29	322.055347506835\\
30	328.333421108143\\
};
\addlegendentry{Belkin et al.: 6.6625};

\addplot [color=mycolor3,solid,line width=1.5pt]
  table[row sep=crcr]{%
1	7.05580186843872\\
2	16.3496017456055\\
3	18.3307456970215\\
4	25.4276714324951\\
5	29.2366695404053\\
6	30.8236961364746\\
7	50.0562324523926\\
8	67.5597610473633\\
9	70.453125\\
10	77.0609588623047\\
11	93.2853012084961\\
12	105.756698608398\\
13	112.739921569824\\
14	117.419937133789\\
15	144.628631591797\\
16	147.532180786133\\
17	161.896942138672\\
18	177.563537597656\\
19	197.257873535156\\
20	199.364868164063\\
21	221.972579956055\\
22	224.937133789063\\
23	242.79280090332\\
24	255.210601806641\\
25	262.310516357422\\
26	270.743011474609\\
27	291.780395507813\\
28	303.935028076172\\
29	318.474822998047\\
30	332.164947509766\\
};
\addlegendentry{NN: 5.526};

\addplot [color=mycolor5,solid,line width=1.5pt]
  table[row sep=crcr]{%
1	6.3412299156189\\
2	10.6845026016235\\
3	14.7073593139648\\
4	18.7813739776611\\
5	20.0568065643311\\
6	32.3215103149414\\
7	59.304817199707\\
8	83.2574462890625\\
9	91.3701553344727\\
10	91.6595230102539\\
11	96.6064453125\\
12	120.607841491699\\
13	131.256744384766\\
14	136.730072021484\\
15	140.92658996582\\
16	178.923080444336\\
17	194.991409301758\\
18	217.861907958984\\
19	227.014022827148\\
20	247.164566040039\\
21	264.173126220703\\
22	282.719604492188\\
23	290.822784423828\\
24	303.717132568359\\
25	328.449798583984\\
26	331.951263427734\\
27	337.549377441406\\
28	357.251220703125\\
29	389.849761962891\\
30	406.436279296875\\
};
\addlegendentry{GT};

\addplot [color=mycolor4,solid,line width=1.5pt]
  table[row sep=crcr]{%
1	6.60021305084229\\
2	13.8827610015869\\
3	19.3910713195801\\
4	22.7062721252441\\
5	27.4157962799072\\
6	34.3898048400879\\
7	57.6000099182129\\
8	74.5008926391602\\
9	81.8639221191406\\
10	87.9896926879883\\
11	98.9813537597656\\
12	107.834083557129\\
13	120.94554901123\\
14	128.309188842773\\
15	146.204010009766\\
16	171.639831542969\\
17	175.724334716797\\
18	192.817459106445\\
19	215.021453857422\\
20	224.909133911133\\
21	239.276443481445\\
22	251.359222412109\\
23	266.654632568359\\
24	276.947357177734\\
25	299.613525390625\\
26	313.099975585938\\
27	327.289337158203\\
28	338.550354003906\\
29	349.742126464844\\
30	355.920532226563\\
};
\addlegendentry{Our: 3.1129};

\legend{}
\end{axis}
\end{tikzpicture}%
%
%
\definecolor{mycolor4}{rgb}{0.00000,0.44700,0.74100}%
\definecolor{mycolor5}{rgb}{0.85000,0.32500,0.09800}%
\definecolor{mycolor3}{rgb}{0.92900,0.69400,0.12500}%
\definecolor{mycolor1}{rgb}{0.49400,0.18400,0.55600}%
\definecolor{mycolor2}{rgb}{0.46600,0.67400,0.18800}%
\begin{tikzpicture}

\begin{axis}[%
width=0.6\linewidth,
height=0.7\linewidth,
at={(0.758in,0.481in)},
scale only axis,
xmin=1,
xmax=30,
ymin=0,
ymax=450,
xtick={10, 20, 30},
xticklabels={ 10, 20, 30},
ytick={0, 150, 300, 4500},
yticklabels={,  ,  },
xmajorgrids,
ymajorgrids,
every x tick label/.append style={font=\color{black}, font=\tiny},
every y tick label/.append style={font=\color{black}, font=\tiny},
title={},
title style={yshift=-0.75em},
axis background/.style={fill=white},
legend columns=-1,
legend style={column sep=0.4ex,at={(-1,-0.2)}, legend cell align=left,align=left,draw=white!15!black, anchor=north west}
]
\addplot [color=mycolor1,solid,line width=1.5pt]
  table[row sep=crcr]{%
1	8.01832671752451\\
2	12.41981321863\\
3	18.9805698256434\\
4	33.8107921361998\\
5	61.4886300502186\\
6	76.249195131724\\
7	89.9888508767576\\
8	100.57328962324\\
9	106.423621992056\\
10	140.97433900234\\
11	144.902278265288\\
12	151.399405103139\\
13	154.98738706465\\
14	169.485073699466\\
15	182.270563695038\\
16	206.009919211805\\
17	232.498077454689\\
18	240.282344871922\\
19	264.955072897543\\
20	280.5119341352\\
21	289.943360885064\\
22	316.487359994087\\
23	319.036818771362\\
24	342.665172951057\\
25	350.660103277527\\
26	380.056826221249\\
27	391.080182493149\\
28	414.341732689592\\
29	426.165928214944\\
30	437.82651879426\\
};
\addlegendentry{LB\_pcl: 9.7318};

\addplot [color=mycolor2,solid,line width=1.5pt]
  table[row sep=crcr]{%
1	6.76212363081583\\
2	11.4992960671273\\
3	14.857973517182\\
4	20.7395696118137\\
5	28.0573943324999\\
6	55.7884055949469\\
7	68.3988102393552\\
8	76.0993635941437\\
9	80.8692436370111\\
10	87.1952163017375\\
11	108.866802690847\\
12	117.767778693885\\
13	118.273619455123\\
14	134.873274919415\\
15	137.111017532545\\
16	155.046905620703\\
17	170.251921040704\\
18	188.322815994867\\
19	195.944873172931\\
20	207.897906033139\\
21	213.278985303569\\
22	218.759471848622\\
23	227.521182739889\\
24	247.325358201194\\
25	254.360589861854\\
26	266.249522330414\\
27	289.185588901904\\
28	293.851867938144\\
29	300.487392468149\\
30	308.039700150451\\
};
\addlegendentry{Belkin et al.: 4.8045};

\addplot [color=mycolor3,solid,line width=1.5pt]
  table[row sep=crcr]{%
1	7.26620483398438\\
2	14.8603763580322\\
3	18.9689903259277\\
4	25.5136566162109\\
5	31.0934467315674\\
6	31.233081817627\\
7	51.2986297607422\\
8	69.304443359375\\
9	70.7731475830078\\
10	79.1353607177734\\
11	92.8870162963867\\
12	103.469596862793\\
13	113.317779541016\\
14	121.039749145508\\
15	144.625030517578\\
16	158.452438354492\\
17	159.744766235352\\
18	186.37385559082\\
19	197.759216308594\\
20	202.333633422852\\
21	219.99462890625\\
22	224.219497680664\\
23	244.742340087891\\
24	253.304626464844\\
25	269.560760498047\\
26	279.966949462891\\
27	297.73388671875\\
28	304.043151855469\\
29	328.239868164063\\
30	337.545532226563\\
};
\addlegendentry{NN: 5.2706};

\addplot [color=mycolor5,solid,line width=1.5pt]
  table[row sep=crcr]{%
1	6.3412299156189\\
2	10.6845026016235\\
3	14.7073593139648\\
4	18.7813739776611\\
5	20.0568065643311\\
6	32.3215103149414\\
7	59.304817199707\\
8	83.2574462890625\\
9	91.3701553344727\\
10	91.6595230102539\\
11	96.6064453125\\
12	120.607841491699\\
13	131.256744384766\\
14	136.730072021484\\
15	140.92658996582\\
16	178.923080444336\\
17	194.991409301758\\
18	217.861907958984\\
19	227.014022827148\\
20	247.164566040039\\
21	264.173126220703\\
22	282.719604492188\\
23	290.822784423828\\
24	303.717132568359\\
25	328.449798583984\\
26	331.951263427734\\
27	337.549377441406\\
28	357.251220703125\\
29	389.849761962891\\
30	406.436279296875\\
};
\addlegendentry{GT};

\addplot [color=mycolor4,solid,line width=1.5pt]
  table[row sep=crcr]{%
1	6.0541090965271\\
2	12.6984615325928\\
3	16.626070022583\\
4	20.6391696929932\\
5	24.0273857116699\\
6	30.7026462554932\\
7	53.9190101623535\\
8	73.8370056152344\\
9	82.1690673828125\\
10	86.5674591064453\\
11	96.8115692138672\\
12	105.238716125488\\
13	114.997856140137\\
14	129.127639770508\\
15	145.831497192383\\
16	165.867630004883\\
17	172.930114746094\\
18	187.383850097656\\
19	209.237899780273\\
20	220.06706237793\\
21	235.531127929688\\
22	243.544448852539\\
23	260.082702636719\\
24	274.584899902344\\
25	300.706726074219\\
26	315.605041503906\\
27	326.143585205078\\
28	334.049621582031\\
29	344.557647705078\\
30	353.640747070313\\
};
\addlegendentry{Our: 2.8563};

\legend{}
\end{axis}
\end{tikzpicture}%
        
    \end{minipage}%
    
    &
    
    \hspace{-0.70cm}
     
    \begin{minipage}{0.135\linewidth} 
    
    \vspace{-2cm}   
                
        \begin{overpic}[trim=0cm 0cm 0cm 0cm,clip,width=0.7\linewidth]{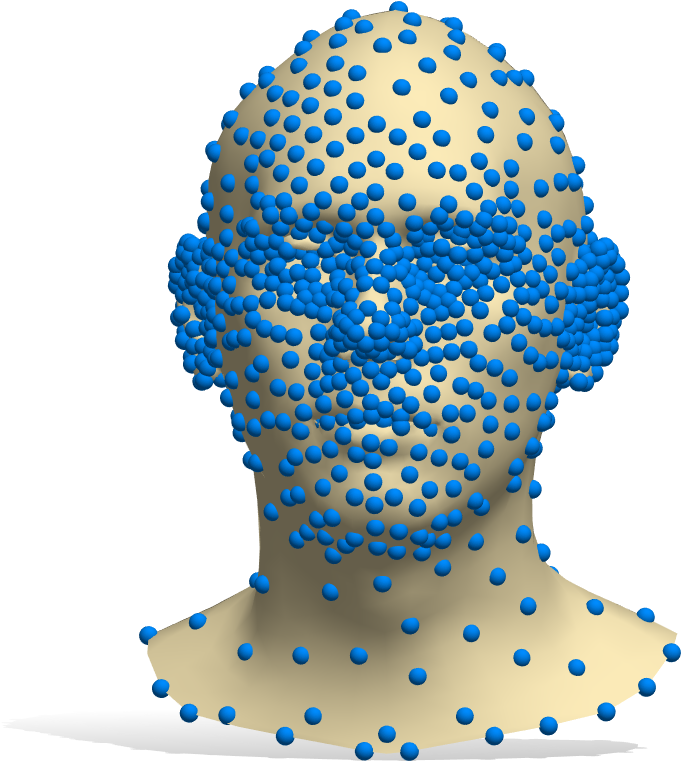}
        \end{overpic}
        
        \vspace{0.2cm}
        
        \begin{overpic}[trim=0cm 0cm 0cm 0cm,clip,width=0.7\linewidth]{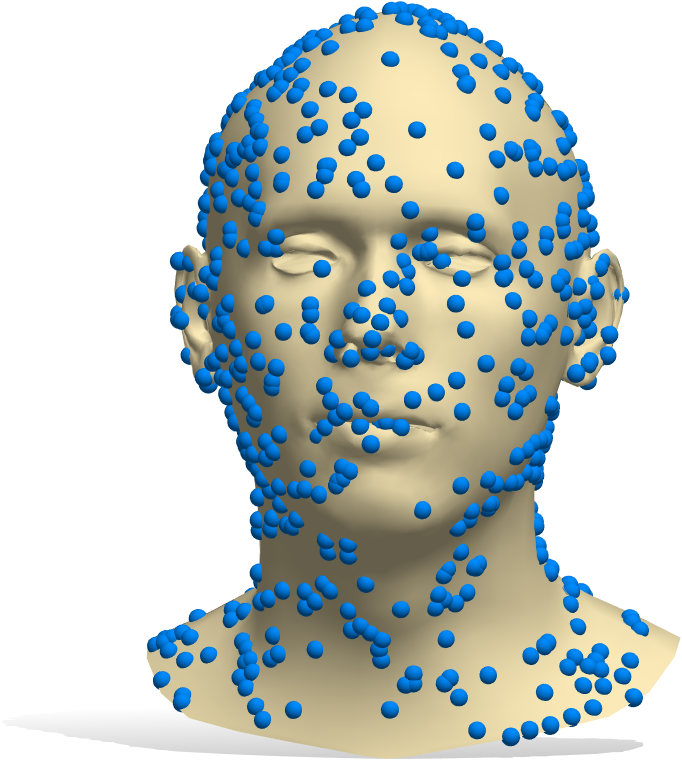}
        \end{overpic}
        
    \end{minipage}%
    
    &
    
    \hspace{-0.7cm}
    
    \begin{minipage}{0.14\linewidth}
        \vspace{-2cm}   
        
%
%
\definecolor{mycolor4}{rgb}{0.00000,0.44700,0.74100}%
\definecolor{mycolor5}{rgb}{0.85000,0.32500,0.09800}%
\definecolor{mycolor3}{rgb}{0.92900,0.69400,0.12500}%
\definecolor{mycolor1}{rgb}{0.49400,0.18400,0.55600}%
\definecolor{mycolor2}{rgb}{0.46600,0.67400,0.18800}%
\begin{tikzpicture}

\begin{axis}[%
width=0.6\linewidth,
height=0.7\linewidth,
at={(0.758in,0.481in)},
scale only axis,
xmin=1,
xmax=30,
ymin=0,
ymax=2500,
xtick={10, 20, 30},
xticklabels={ , , },
ytick={0, 833.3333, 1666.666, 2500},
yticklabels={ ,  ,  , },
xmajorgrids,
ymajorgrids,
every x tick label/.append style={font=\color{black}, font=\tiny},
every y tick label/.append style={font=\color{black}, font=\tiny},
title={\scriptsize eigenvalues},
title style={yshift=-0.75em},
axis background/.style={fill=white},
legend style={legend cell align=left,align=left,draw=white!15!black}
]
\addplot [color=mycolor1,solid,line width=1.5pt]
  table[row sep=crcr]{%
1	92.051994484157\\
2	157.286469120153\\
3	205.466451499584\\
4	283.670274672259\\
5	315.122383382319\\
6	339.580314733565\\
7	504.583322103653\\
8	589.474259207706\\
9	611.94273245642\\
10	673.545563157824\\
11	732.852153344336\\
12	770.915095423238\\
13	846.232416526637\\
14	966.804160119655\\
15	1025.86484320966\\
16	1064.25760268795\\
17	1145.92168538717\\
18	1213.52866563909\\
19	1232.39312830124\\
20	1303.13961572959\\
21	1367.87968671591\\
22	1440.77380809352\\
23	1452.92189157386\\
24	1577.23110539671\\
25	1632.27372494628\\
26	1698.38533405393\\
27	1733.43480654627\\
28	1764.00817368824\\
29	1877.77332365268\\
30	1946.96745741342\\
};
\addlegendentry{LB\_pcl: 1.1225};

\addplot [color=mycolor2,solid,line width=1.5pt]
  table[row sep=crcr]{%
1	106.583482877004\\
2	119.168959031623\\
3	187.734967847965\\
4	209.44871160566\\
5	228.814700096358\\
6	238.709727467693\\
7	255.224653339459\\
8	256.18258174827\\
9	293.452326834652\\
10	383.113756062705\\
11	425.260674769105\\
12	464.343184781822\\
13	476.606307047211\\
14	510.807831616899\\
15	557.887210643717\\
16	573.507222890432\\
17	644.674495867817\\
18	665.493104160244\\
19	740.96535308185\\
20	756.817313226782\\
21	791.897309439033\\
22	825.319221712003\\
23	876.344336718611\\
24	917.643970352255\\
25	1029.5754982514\\
26	1063.66357860041\\
27	1122.9888634008\\
28	1136.277769845\\
29	1185.77333140891\\
30	1201.08342743094\\
};
\addlegendentry{Belkin et al.: 11.6329};

\addplot [color=mycolor3,solid,line width=1.5pt]
  table[row sep=crcr]{%
1	78.5449523925781\\
2	146.275619506836\\
3	169.493041992188\\
4	247.047180175781\\
5	279.651458740234\\
6	297.846282958984\\
7	433.332092285156\\
8	522.576232910156\\
9	525.571716308594\\
10	606.163513183594\\
11	609.152587890625\\
12	746.075744628906\\
13	761.97802734375\\
14	848.941650390625\\
15	885.933349609375\\
16	943.347961425781\\
17	1052.53161621094\\
18	1066.84936523438\\
19	1105.38623046875\\
20	1168.45031738281\\
21	1288.27917480469\\
22	1359.29150390625\\
23	1488.48107910156\\
24	1524.375\\
25	1568.11779785156\\
26	1595.97766113281\\
27	1666.89025878906\\
28	1692.62573242188\\
29	1741.00708007813\\
30	1853.79699707031\\
};
\addlegendentry{NN: 2.8486};

\addplot [color=mycolor5,solid,line width=1.5pt]
  table[row sep=crcr]{%
1	90.2838897705078\\
2	155.843948364258\\
3	182.884643554688\\
4	276.514678955078\\
5	290.470764160156\\
6	341.724945068359\\
7	474.112487792969\\
8	578.340637207031\\
9	597.244323730469\\
10	672.757385253906\\
11	695.014465332031\\
12	789.047973632813\\
13	823.812255859375\\
14	942.249450683594\\
15	1004.369140625\\
16	1035.16564941406\\
17	1188.37219238281\\
18	1200.55419921875\\
19	1263.35498046875\\
20	1311.52221679688\\
21	1400.71374511719\\
22	1497.25183105469\\
23	1566.72192382813\\
24	1626.78527832031\\
25	1739.22607421875\\
26	1747.44482421875\\
27	1890.65368652344\\
28	1898.29187011719\\
29	1928.70153808594\\
30	2053.88940429688\\
};
\addlegendentry{GT};

\addplot [color=mycolor4,solid,line width=1.5pt]
  table[row sep=crcr]{%
1	87.8590927124023\\
2	158.79655456543\\
3	182.205612182617\\
4	265.063385009766\\
5	298.372985839844\\
6	334.126525878906\\
7	478.982116699219\\
8	554.211181640625\\
9	587.085144042969\\
10	642.929321289063\\
11	687.055358886719\\
12	786.818176269531\\
13	812.489318847656\\
14	928.718872070313\\
15	982.303771972656\\
16	1059.12268066406\\
17	1164.39794921875\\
18	1183.33703613281\\
19	1230.18151855469\\
20	1280.53161621094\\
21	1402.89099121094\\
22	1494.82934570313\\
23	1568.98107910156\\
24	1627.56677246094\\
25	1714.17297363281\\
26	1753.05541992188\\
27	1809.80700683594\\
28	1890.89929199219\\
29	1928.14208984375\\
30	2037.3662109375\\
};
\addlegendentry{Our: 0.50331};

\legend{}
\end{axis}
\end{tikzpicture}%
        \vspace{0.0cm} 
        
%
%
\definecolor{mycolor4}{rgb}{0.00000,0.44700,0.74100}%
\definecolor{mycolor5}{rgb}{0.85000,0.32500,0.09800}%
\definecolor{mycolor3}{rgb}{0.92900,0.69400,0.12500}%
\definecolor{mycolor1}{rgb}{0.49400,0.18400,0.55600}%
\definecolor{mycolor2}{rgb}{0.46600,0.67400,0.18800}%
\begin{tikzpicture}

\begin{axis}[%
width=0.6\linewidth,
height=0.7\linewidth,
at={(0.758in,0.481in)},
scale only axis,
xmin=1,
xmax=30,
ymin=0,
ymax=2500,
xtick={10, 20, 30},
xticklabels={ 10, 20, 30},
ytick={0, 833.3333, 1666.666, 2500},
yticklabels={,  ,  },
xmajorgrids,
ymajorgrids,
every x tick label/.append style={font=\color{black}, font=\tiny},
every y tick label/.append style={font=\color{black}, font=\tiny},
title={},
title style={yshift=-0.75em},
axis background/.style={fill=white},
legend columns=-1,
legend style={column sep=0.4ex,at={(-1,-0.2)}, legend cell align=left,align=left,draw=white!15!black, anchor=north west}
]
\addplot [color=mycolor1,solid,line width=1.5pt]
  table[row sep=crcr]{%
1	99.3568480998126\\
2	169.130853482553\\
3	203.398667410628\\
4	302.908729769149\\
5	332.779323007417\\
6	383.630465066739\\
7	537.718007465723\\
8	606.819020567604\\
9	653.039935611672\\
10	689.347059559769\\
11	759.428082297524\\
12	884.875264292755\\
13	934.808098825316\\
14	1070.97768227628\\
15	1080.48374670894\\
16	1107.47072753509\\
17	1224.66563802573\\
18	1280.46202638876\\
19	1304.45524866182\\
20	1325.54180842116\\
21	1552.49287864604\\
22	1566.5088042944\\
23	1652.14796265825\\
24	1743.88924370439\\
25	1810.3157928994\\
26	1846.84459038754\\
27	1900.99601298941\\
28	1976.98444438376\\
29	2025.56440813363\\
30	2077.34818712606\\
};
\addlegendentry{LB\_pcl: 2.222};

\addplot [color=mycolor2,solid,line width=1.5pt]
  table[row sep=crcr]{%
1	80.0041153809453\\
2	161.800147587812\\
3	173.509527905282\\
4	250.521632881678\\
5	278.022013909585\\
6	337.255190102892\\
7	469.467903829029\\
8	533.229771834999\\
9	579.797152586631\\
10	601.95333005334\\
11	653.055253674373\\
12	683.275391832638\\
13	798.582896333466\\
14	871.267045797802\\
15	929.503913427783\\
16	979.480592858579\\
17	1042.28048148935\\
18	1116.90304826332\\
19	1154.92014958949\\
20	1200.97304753156\\
21	1225.19195994882\\
22	1318.86245480618\\
23	1366.67987330457\\
24	1445.05532235393\\
25	1481.7532419613\\
26	1528.18891381018\\
27	1572.07199417154\\
28	1595.95750287393\\
29	1705.73226757386\\
30	1749.89996739744\\
};
\addlegendentry{Belkin et al.: 2.7157};

\addplot [color=mycolor3,solid,line width=1.5pt]
  table[row sep=crcr]{%
1	78.5449523925781\\
2	146.275619506836\\
3	169.493041992188\\
4	247.047180175781\\
5	279.651458740234\\
6	297.846282958984\\
7	433.332092285156\\
8	522.576232910156\\
9	525.571716308594\\
10	606.163513183594\\
11	609.152587890625\\
12	746.075744628906\\
13	761.97802734375\\
14	848.941650390625\\
15	885.933349609375\\
16	943.347961425781\\
17	1052.53161621094\\
18	1066.84936523438\\
19	1105.38623046875\\
20	1168.45031738281\\
21	1288.27917480469\\
22	1359.29150390625\\
23	1488.48107910156\\
24	1524.375\\
25	1568.11779785156\\
26	1595.97766113281\\
27	1666.89025878906\\
28	1692.62573242188\\
29	1741.00708007813\\
30	1853.79699707031\\
};
\addlegendentry{NN: 2.8486};

\addplot [color=mycolor5,solid,line width=1.5pt]
  table[row sep=crcr]{%
1	90.2838897705078\\
2	155.843948364258\\
3	182.884643554688\\
4	276.514678955078\\
5	290.470764160156\\
6	341.724945068359\\
7	474.112487792969\\
8	578.340637207031\\
9	597.244323730469\\
10	672.757385253906\\
11	695.014465332031\\
12	789.047973632813\\
13	823.812255859375\\
14	942.249450683594\\
15	1004.369140625\\
16	1035.16564941406\\
17	1188.37219238281\\
18	1200.55419921875\\
19	1263.35498046875\\
20	1311.52221679688\\
21	1400.71374511719\\
22	1497.25183105469\\
23	1566.72192382813\\
24	1626.78527832031\\
25	1739.22607421875\\
26	1747.44482421875\\
27	1890.65368652344\\
28	1898.29187011719\\
29	1928.70153808594\\
30	2053.88940429688\\
};
\addlegendentry{GT};

\addplot [color=mycolor4,solid,line width=1.5pt]
  table[row sep=crcr]{%
1	87.6991882324219\\
2	160.825500488281\\
3	183.257171630859\\
4	272.912078857422\\
5	301.779724121094\\
6	334.066223144531\\
7	483.564392089844\\
8	573.681823730469\\
9	592.963256835938\\
10	658.059387207031\\
11	700.948486328125\\
12	799.8759765625\\
13	821.489196777344\\
14	947.59375\\
15	991.08251953125\\
16	1060.97912597656\\
17	1179.40600585938\\
18	1196.43811035156\\
19	1239.60485839844\\
20	1309.05773925781\\
21	1441.88037109375\\
22	1514.07543945313\\
23	1589.564453125\\
24	1642.59191894531\\
25	1734.73852539063\\
26	1764.708984375\\
27	1822.32067871094\\
28	1909.42517089844\\
29	1949.95288085938\\
30	2070.02880859375\\
};
\addlegendentry{Our: 0.43285};

\legend{}
\end{axis}
\end{tikzpicture}%
    \end{minipage}%
    
    &
    
    \hspace{-0.75cm}
     
    \begin{minipage}{0.135\linewidth}
        \vspace{-2cm}

        \vspace{0.5cm}    
    
        \begin{overpic}[trim=0cm 0cm 0cm 0cm,clip,width=0.9\linewidth]{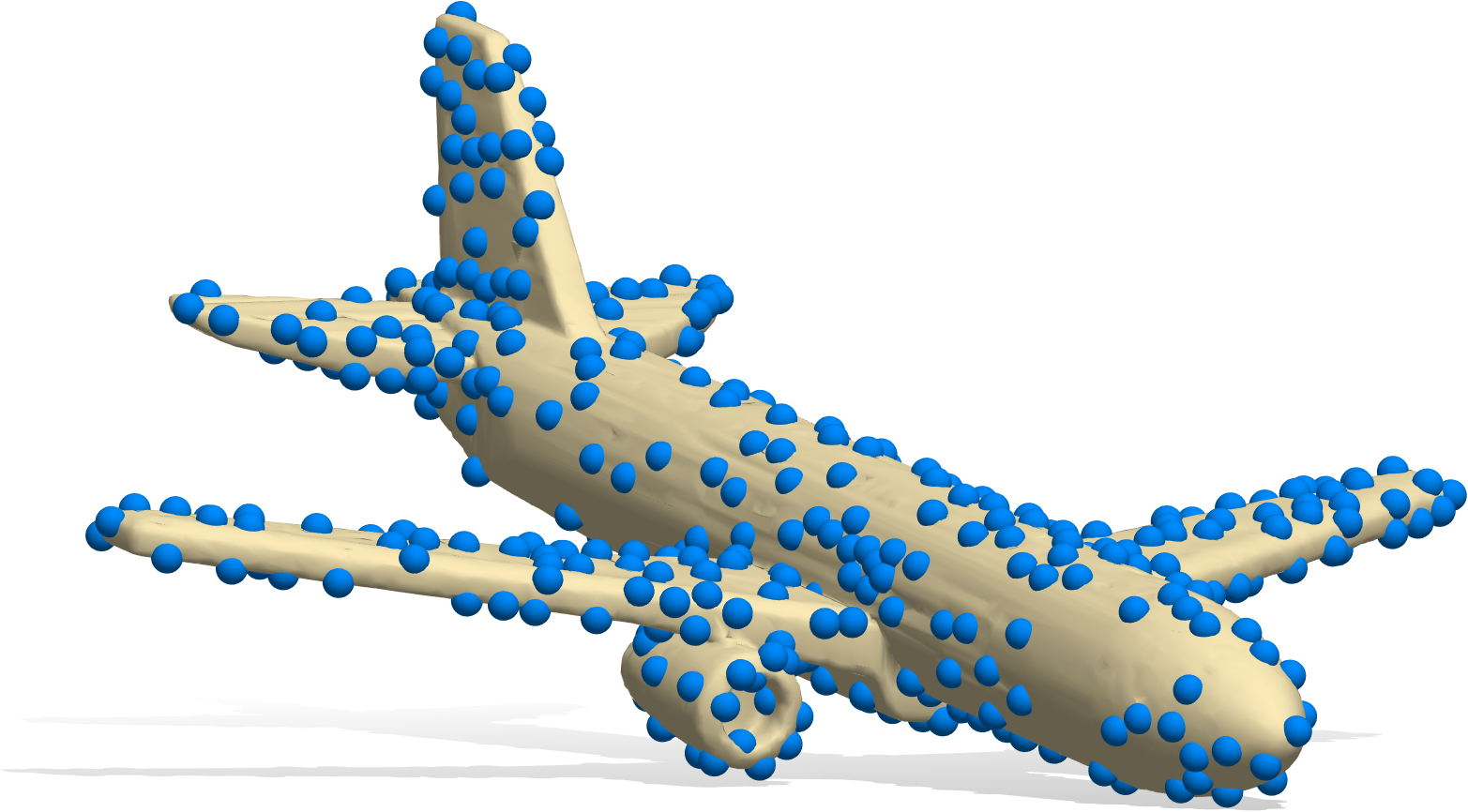}
        \end{overpic}
        
        \vspace{0.7cm}
        
        \begin{overpic}[trim=0cm 0cm 0cm 0cm,clip,width=0.75\linewidth]{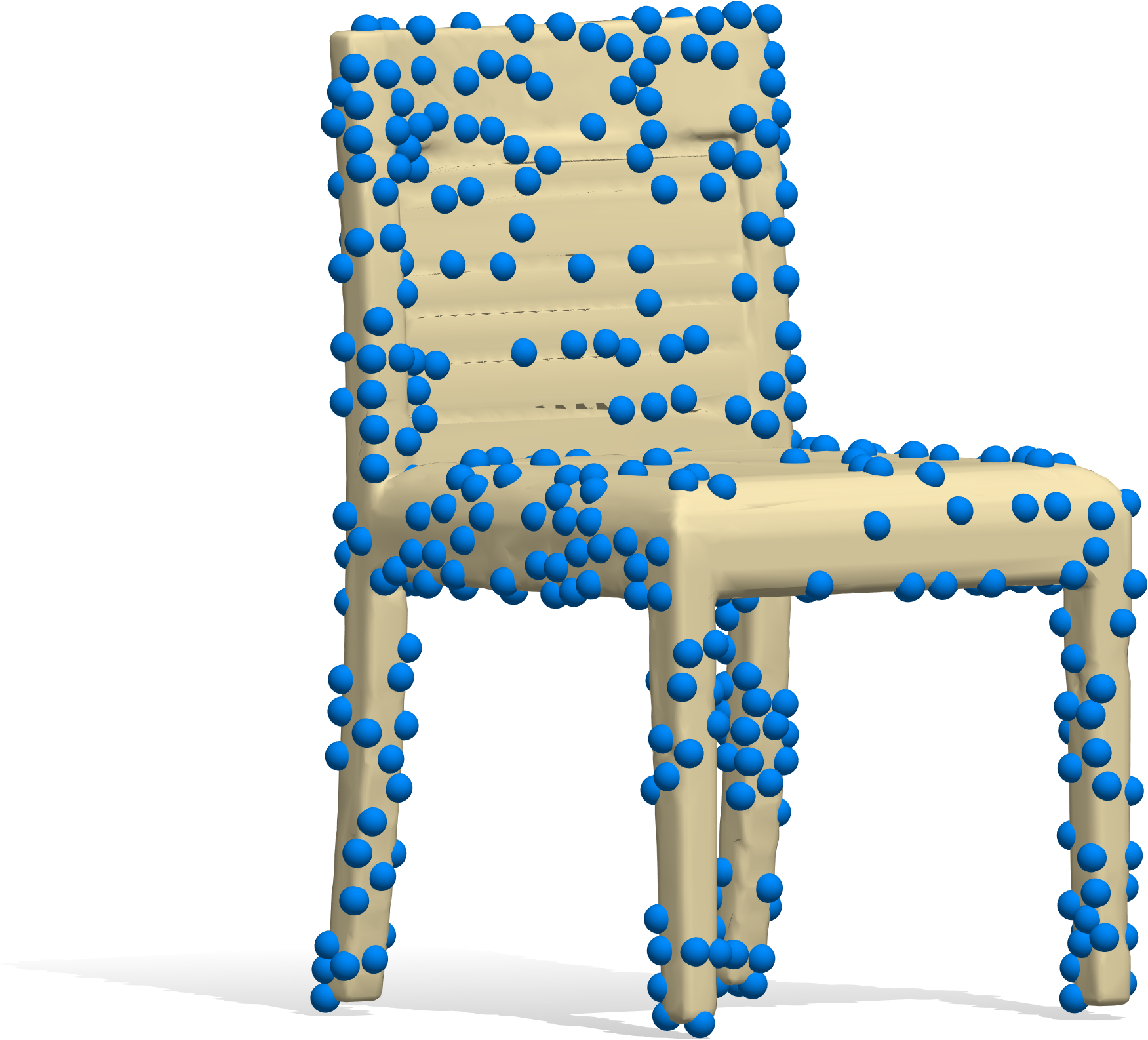}
        \put(218,0){\line(0,1){220}}
        
        \end{overpic}
    
    \end{minipage}%
    
    &
    
    \hspace{-0.4cm}
    
    \begin{minipage}{0.14\linewidth}
        \vspace{-2cm}   
        
%
%
\definecolor{mycolor4}{rgb}{0.00000,0.44700,0.74100}%
\definecolor{mycolor5}{rgb}{0.85000,0.32500,0.09800}%
\definecolor{mycolor3}{rgb}{0.92900,0.69400,0.12500}%
\definecolor{mycolor1}{rgb}{0.49400,0.18400,0.55600}%
\definecolor{mycolor2}{rgb}{0.46600,0.67400,0.18800}%
\begin{tikzpicture}

\begin{axis}[%
width=0.6\linewidth,
height=0.7\linewidth,
at={(0.758in,0.481in)},
scale only axis,
xmin=1,
xmax=30,
ymin=0,
ymax=450,
xtick={10, 20, 30},
xticklabels={ , , },
ytick={0, 150, 300, 450},
yticklabels={,  ,  },
xmajorgrids,
ymajorgrids,
every x tick label/.append style={font=\color{black}, font=\tiny},
every y tick label/.append style={font=\color{black}, font=\tiny},
title={\scriptsize eigenvalues},
title style={yshift=-0.75em},
axis background/.style={fill=white},
legend style={legend cell align=left,align=left,draw=white!15!black}
]
\addplot [color=mycolor1, line width=1.5pt]
  table[row sep=crcr]{%
1	4.12483010488421\\
2	9.96635535386622\\
3	14.0769775163384\\
4	24.6048243590736\\
5	36.9136321744448\\
6	61.3634007489572\\
7	70.4686328334626\\
8	71.9971084895203\\
9	83.2643775037119\\
10	94.3138020699138\\
11	101.381353726075\\
12	114.822851552383\\
13	131.478669033942\\
14	136.171492105954\\
15	152.198566823874\\
16	161.481345614885\\
17	173.930070132745\\
18	177.501178613242\\
19	181.250200823257\\
20	202.736041076824\\
21	218.839226039249\\
22	235.427838145646\\
23	253.294619602762\\
24	266.169850526431\\
25	268.959436565446\\
26	277.81050410079\\
27	290.085795312634\\
28	294.144831141115\\
29	300.621728408876\\
30	325.38177508625\\
};
\addlegendentry{LB pcl}

\addplot [color=mycolor2, line width=1.5pt]
  table[row sep=crcr]{%
1	5.79823510886524\\
2	14.7280802267529\\
3	16.5799301651024\\
4	30.672670790851\\
5	38.9386122895532\\
6	52.1453869089443\\
7	62.8219239268123\\
8	73.8404656752901\\
9	92.8320018178055\\
10	105.557973191398\\
11	106.973180053367\\
12	129.929999327779\\
13	145.052315420209\\
14	159.579118679247\\
15	166.258529472128\\
16	176.616754194974\\
17	177.153447002496\\
18	187.367047116999\\
19	194.053555363918\\
20	203.181934463226\\
21	211.623327130883\\
22	218.901283708053\\
23	230.943915530667\\
24	243.999898930378\\
25	250.872528768016\\
26	271.297483561897\\
27	277.461961196706\\
28	301.433581093398\\
29	317.48730456522\\
30	335.759387206086\\
};
\addlegendentry{Belkin et al.}

\addplot [color=mycolor3, line width=1.5pt]
  table[row sep=crcr]{%
1	4.30435180664063\\
2	9.374267578125\\
3	9.63174438476563\\
4	23.4203491210938\\
5	25.9844360351563\\
6	30.5943298339844\\
7	40.7926330566406\\
8	45.2403869628906\\
9	51.7097778320313\\
10	57.1986083984375\\
11	59.4768371582031\\
12	85.2870483398438\\
13	94.4889526367188\\
14	111.114837646484\\
15	138.746368408203\\
16	146.892883300781\\
17	168.650726318359\\
18	181.013793945313\\
19	214.045776367188\\
20	250.585723876953\\
21	256.880920410156\\
22	258.808624267578\\
23	278.732330322266\\
24	285.042114257813\\
25	298.521667480469\\
26	303.364318847656\\
27	311.628814697266\\
28	317.026000976563\\
29	321.021820068359\\
30	342.057678222656\\
};
\addlegendentry{NN}

\addplot [color=mycolor5, line width=1.5pt]
  table[row sep=crcr]{%
1	4.53631591796875\\
2	10.6399841308594\\
3	11.4770812988281\\
4	25.9176025390625\\
5	35.7257385253906\\
6	40.0388793945313\\
7	58.650634765625\\
8	62.4209899902344\\
9	69.4515075683594\\
10	99.0671691894531\\
11	101.049377441406\\
12	120.920166015625\\
13	139.412567138672\\
14	140.841125488281\\
15	162.084838867188\\
16	182.9912109375\\
17	212.537628173828\\
18	221.343780517578\\
19	225.9150390625\\
20	232.616973876953\\
21	234.391143798828\\
22	240.260833740234\\
23	247.053771972656\\
24	248.593841552734\\
25	250.641021728516\\
26	274.813781738281\\
27	288.627319335938\\
28	298.173522949219\\
29	312.607849121094\\
30	340.905700683594\\
};
\addlegendentry{GT}

\addplot [color=mycolor4, line width=1.5pt]
  table[row sep=crcr]{%
1	5.37255859375\\
2	8.19601440429688\\
3	10.8541259765625\\
4	22.9943237304688\\
5	32.6732177734375\\
6	39.8045349121094\\
7	51.7733764648438\\
8	59.1537475585938\\
9	65.9798889160156\\
10	74.4645690917969\\
11	82.4432067871094\\
12	99.188720703125\\
13	111.304931640625\\
14	126.301086425781\\
15	146.494964599609\\
16	163.563629150391\\
17	179.154510498047\\
18	197.431701660156\\
19	213.057556152344\\
20	230.179107666016\\
21	243.920715332031\\
22	256.360290527344\\
23	267.95654296875\\
24	279.352905273438\\
25	290.307495117188\\
26	301.025299072266\\
27	311.897979736328\\
28	323.713073730469\\
29	334.715789794922\\
30	348.475494384766\\
};
\addlegendentry{Our}

\legend{}
\end{axis}
\end{tikzpicture}%
%
%
\definecolor{mycolor4}{rgb}{0.00000,0.44700,0.74100}%
\definecolor{mycolor5}{rgb}{0.85000,0.32500,0.09800}%
\definecolor{mycolor3}{rgb}{0.92900,0.69400,0.12500}%
\definecolor{mycolor1}{rgb}{0.49400,0.18400,0.55600}%
\definecolor{mycolor2}{rgb}{0.46600,0.67400,0.18800}%
\begin{tikzpicture}

\begin{axis}[%
width=0.6\linewidth,
height=0.7\linewidth,
at={(0.758in,0.481in)},
scale only axis,
xmin=1,
xmax=30,
ymin=0,
ymax=450,
xtick={10, 20, 30},
xticklabels={ 10, 20, 30},
ytick={0, 150, 300, 4500},
yticklabels={,  ,  },
xmajorgrids,
ymajorgrids,
every x tick label/.append style={font=\color{black}, font=\tiny},
every y tick label/.append style={font=\color{black}, font=\tiny},
title={},
title style={yshift=-0.75em},
axis background/.style={fill=white},
legend style={legend cell align=left,align=left,draw=white!15!black}
]
\addplot [color=mycolor1, line width=1.5pt]
  table[row sep=crcr]{%
1	4.99816438035231\\
2	7.61488154835084\\
3	7.8285121149228\\
4	13.6491094957267\\
5	13.8095954379987\\
6	16.7436739471216\\
7	26.1171964900544\\
8	32.1678115922769\\
9	32.8841526227675\\
11	50.471423966222\\
12	54.5606986674833\\
13	68.8671486078183\\
14	74.4226867091131\\
15	89.0289186692688\\
16	106.863794370689\\
18	126.668260456498\\
19	137.061764037877\\
20	159.88829926777\\
21	169.075771620758\\
22	178.976017436808\\
23	195.48511648562\\
24	198.798864029986\\
25	222.09910704749\\
26	224.050437291207\\
27	232.002290787768\\
28	241.001977287068\\
29	246.119740303404\\
30	254.342320675673\\
};
\addlegendentry{LB pcl}

\addplot [color=mycolor2, line width=1.5pt]
  table[row sep=crcr]{%
1	13.246100631391\\
2	28.6794465161137\\
3	32.7258327628697\\
4	44.4317853555874\\
5	61.0782307722583\\
6	77.0270817339418\\
7	93.6593413901787\\
8	112.864680374114\\
9	137.815111074723\\
10	151.837464054344\\
11	164.269759727737\\
12	171.062186554363\\
13	189.657290321428\\
14	211.985937176151\\
15	230.774623825608\\
16	262.46550139107\\
17	285.351298381492\\
18	299.384450555173\\
19	317.376173774745\\
20	335.852476950913\\
21	360.109320254446\\
22	373.12436137513\\
23	385.942318637026\\
24	395.31075828333\\
25	407.23583694485\\
26	422.920172802559\\
27	462.160863582653\\
28	472.255945620748\\
29	481.361305372524\\
30	502.140956601544\\
};
\addlegendentry{Belkin et al.}

\addplot [color=mycolor3, line width=1.5pt]
  table[row sep=crcr]{%
1	13.697021484375\\
2	22.7979736328125\\
3	27.5061340332031\\
4	29.9098815917969\\
5	40.50244140625\\
6	41.0823059082031\\
7	50.2127075195313\\
8	64.6213989257813\\
9	64.8377380371094\\
10	79.6309814453125\\
11	82.0702819824219\\
12	95.9877319335938\\
13	112.424865722656\\
14	129.582641601563\\
15	154.754974365234\\
16	179.416381835938\\
17	181.715209960938\\
18	189.109680175781\\
19	189.384582519531\\
20	221.075042724609\\
21	223.870483398438\\
22	230.85009765625\\
23	241.994964599609\\
24	249.833831787109\\
25	256.000122070313\\
26	272.574890136719\\
27	301.547180175781\\
28	316.669494628906\\
29	328.038543701172\\
30	335.717041015625\\
};
\addlegendentry{NN}

\addplot [color=mycolor5, line width=1.5pt]
  table[row sep=crcr]{%
1	10.8077392578125\\
2	15.2053833007813\\
3	17.3267517089844\\
4	17.9339599609375\\
5	27.1559143066406\\
6	50.6089782714844\\
7	51.5133361816406\\
8	63.1691284179688\\
9	67.0074462890625\\
10	89.1207580566406\\
11	101.923095703125\\
12	124.060760498047\\
13	136.859741210938\\
14	147.668029785156\\
15	155.24658203125\\
16	163.293304443359\\
17	169.527648925781\\
18	189.9248046875\\
19	191.67041015625\\
20	216.286102294922\\
21	218.580474853516\\
22	229.947448730469\\
23	242.288421630859\\
24	243.414215087891\\
25	261.035400390625\\
26	277.932373046875\\
27	306.433776855469\\
28	311.656646728516\\
29	333.735504150391\\
30	345.724395751953\\
};
\addlegendentry{GT}

\addplot [color=mycolor4, line width=1.5pt]
  table[row sep=crcr]{%
1	7.80828857421875\\
2	12.5137329101563\\
3	17.9447021484375\\
4	21.8641357421875\\
5	28.839599609375\\
6	40.5272521972656\\
7	51.39697265625\\
8	63.1459350585938\\
9	73.1322326660156\\
10	84.1802978515625\\
11	95.8323059082031\\
12	105.386108398438\\
13	118.292419433594\\
14	130.367218017578\\
15	141.317077636719\\
16	151.972320556641\\
17	164.701263427734\\
18	178.153930664063\\
19	190.032989501953\\
20	201.076232910156\\
21	211.740753173828\\
23	237.112945556641\\
24	250.323791503906\\
26	276.258056640625\\
27	290.526763916016\\
28	303.509368896484\\
29	318.4296875\\
30	328.986907958984\\
};
\addlegendentry{Our}

\legend{}
\end{axis}
\end{tikzpicture}%
    \end{minipage}%
    
    \hspace{-0.8cm}
    
    &
 
    \begin{minipage}{0.19\linewidth}
    
        \vspace{-2.2cm}
        
%
%
\definecolor{mycolor4}{rgb}{0.00000,0.44700,0.74100}%
\definecolor{mycolor5}{rgb}{0.85000,0.32500,0.09800}%
\definecolor{mycolor3}{rgb}{0.92900,0.69400,0.12500}%
\definecolor{mycolor1}{rgb}{0.49400,0.18400,0.55600}%
\definecolor{mycolor2}{rgb}{0.46600,0.67400,0.18800}%
\begin{tikzpicture}

\begin{axis}[%
width=0.7\linewidth,
height=0.5\linewidth,
at={(0.758in,0.481in)},
scale only axis,
xmin=1,
xmax=30,
ymin=0,
ymax=9,
xtick={10, 20, 30},
xticklabels={ , , },
ytick={0, 3, 6, 9},
yticklabels={ , ,  ,  },
ylabel style={yshift=-0.6em},
xmajorgrids,
ymajorgrids,
every x tick label/.append style={font=\color{black}, font=\tiny},
every y tick label/.append style={font=\color{black}, font=\tiny},
title={\scriptsize Cumulative errors},
title style={yshift=-0.75em},
axis background/.style={fill=white},
legend columns=-1,
legend style={column sep=2ex,at={(-10,-0.4)}, cells={align=right},align=left,draw=white!15!black, anchor=north east}
]
\addplot [color=mycolor1,solid,line width=1.5pt]
  table[row sep=crcr]{%
1	0.0146822178261252\\
2	0.0223174932220845\\
3	0.152926022487319\\
4	0.172360335464275\\
5	0.20390748998127\\
6	0.219698536998641\\
7	0.282752381786953\\
8	0.303247501545514\\
9	0.349054096016085\\
10	0.364200020746868\\
11	0.405300525400988\\
12	0.427463655308102\\
13	0.444929425914104\\
14	0.461908348597879\\
15	0.474816066081164\\
16	0.494897047440486\\
17	0.525914096520977\\
18	0.551512423986162\\
19	0.571045179812855\\
20	0.583565527099942\\
21	0.606257116360248\\
22	0.632879129936419\\
23	0.698906283717591\\
24	0.773923418313301\\
25	0.836816094419999\\
26	0.900972760619078\\
27	0.960338193831921\\
28	1.0245306812147\\
29	1.0794197143401\\
30	1.11685950012511\\
};
\addlegendentry{\scriptsize \cite{clarenz2004finite}: 1.12 \\ \scriptsize 1.92 \\ \scriptsize 12.0};

\addplot [color=mycolor2,solid,line width=1.5pt]
  table[row sep=crcr]{%
1	0.422494377440255\\
2	0.540181865054805\\
3	0.785552109709892\\
4	0.882507160740607\\
5	0.993317851125222\\
6	1.10948469317802\\
7	1.43982905564895\\
8	1.85797693801043\\
9	2.16921186813594\\
10	2.40264851816942\\
11	2.59531857819227\\
12	2.89506045814498\\
13	3.17872328272305\\
14	3.51336004334369\\
15	3.8233094755867\\
16	4.13498803391853\\
17	4.47561429140614\\
18	4.80176402839508\\
19	5.11709488867887\\
20	5.43887749580976\\
21	5.78476568619668\\
22	6.11973028829773\\
23	6.50036111610416\\
24	6.85739044996355\\
25	7.20624067557698\\
26	7.54420149577404\\
27	7.87274015563996\\
28	8.19745939692343\\
29	8.50208952684645\\
30	8.82480532579289\\
};
\addlegendentry{\scriptsize \cite{belkin2009constructing}: 8.82 \\ \scriptsize 2.30 \\ \scriptsize 6.81};

\addplot [color=mycolor3,solid,line width=1.5pt]
  table[row sep=crcr]{%
1	0.0633715018630028\\
2	0.11784852296114\\
3	0.205927446484566\\
4	0.273435175418854\\
5	0.346421092748642\\
6	0.436775058507919\\
7	0.492882013320923\\
8	0.568726539611816\\
9	0.630772590637207\\
10	0.692314565181732\\
11	0.760969698429108\\
12	0.840890049934387\\
13	0.925047278404236\\
14	1.00753748416901\\
15	1.08194637298584\\
16	1.15346896648407\\
17	1.22329390048981\\
18	1.28947007656097\\
19	1.36600410938263\\
20	1.43039906024933\\
21	1.49612128734589\\
22	1.54785215854645\\
23	1.617884516716\\
24	1.70040988922119\\
25	1.78223586082458\\
26	1.86034095287323\\
27	1.93552827835083\\
28	2.01321029663086\\
29	2.0690484046936\\
30	2.14226651191711\\
};
\addlegendentry{\scriptsize NN: 2.14 \\ \scriptsize 1.75 \\ \scriptsize 3.07};

\addplot [color=mycolor4,solid,line width=1.5pt]
  table[row sep=crcr]{%
1	0.0458786264061928\\
2	0.0780017077922821\\
3	0.108483962714672\\
4	0.124371893703938\\
5	0.167396619915962\\
6	0.192936167120934\\
7	0.204498112201691\\
8	0.244601801037788\\
9	0.25893771648407\\
10	0.27800714969635\\
11	0.289399445056915\\
12	0.32318514585495\\
13	0.344560235738754\\
14	0.371626317501068\\
15	0.387164235115051\\
16	0.402303457260132\\
17	0.414364695549011\\
18	0.428028672933578\\
19	0.445232331752777\\
20	0.460014820098877\\
21	0.482685655355453\\
22	0.503156363964081\\
23	0.528767764568329\\
24	0.561595439910889\\
25	0.585793077945709\\
26	0.603423893451691\\
27	0.621609032154083\\
28	0.631681561470032\\
29	0.642870128154755\\
30	0.655016899108887\\
};
\addlegendentry{\scriptsize \textbf{Ours: 0.66} \\ \scriptsize \textbf{0.67} \\ \scriptsize \textbf{2.14}};

\legend{}
\node[text width=0.5cm] at  (40,820) {\tiny F1};

\end{axis}

\end{tikzpicture}%
        \vspace{-0.15cm}
        
%
%
\definecolor{mycolor4}{rgb}{0.00000,0.44700,0.74100}%
\definecolor{mycolor5}{rgb}{0.85000,0.32500,0.09800}%
\definecolor{mycolor3}{rgb}{0.92900,0.69400,0.12500}%
\definecolor{mycolor1}{rgb}{0.49400,0.18400,0.55600}%
\definecolor{mycolor2}{rgb}{0.46600,0.67400,0.18800}%
\begin{tikzpicture}

\begin{axis}[%
width=0.7\linewidth,
height=0.5\linewidth,
at={(0.758in,0.481in)},
scale only axis,
xmin=1,
xmax=30,
ymin=0,
ymax=2.5,
xtick={10, 20, 30},
xticklabels={ 10, 20, 30},
ytick={0, 0.833333, 1.66666, 2.5},
yticklabels={ , ,  ,  },
xmajorgrids,
ymajorgrids,
every x tick label/.append style={font=\color{black}, font=\tiny},
every y tick label/.append style={font=\color{black}, font=\tiny},
title style={yshift=-0.75em},
axis background/.style={fill=white},
]
\addplot [color=mycolor1,solid,line width=1.5pt]
  table[row sep=crcr]{%
1	0.0841755138147081\\
2	0.167307688361937\\
3	0.268083537188502\\
4	0.344718580763538\\
5	0.461742012745632\\
6	0.572779638060551\\
7	0.688290379318268\\
8	0.737764838681336\\
9	0.817000584283037\\
10	0.855277315234956\\
11	0.928566553399615\\
12	1.0056139804555\\
13	1.11117644273502\\
14	1.17803774806323\\
15	1.26096397454056\\
16	1.3555667106492\\
17	1.38670063211502\\
18	1.44209150767746\\
19	1.5005282323546\\
20	1.53523103590473\\
21	1.58846978518251\\
22	1.64574536933273\\
23	1.67611597053592\\
24	1.70556298773359\\
25	1.72596247354702\\
26	1.75365715978092\\
27	1.77920823006571\\
28	1.8166726909337\\
29	1.8815589606885\\
30	1.91527576628003\\
};
\addlegendentry{\scriptsize \cite{clarenz2004finite}: 1.92};

\addplot [color=mycolor2,solid,line width=1.5pt]
  table[row sep=crcr]{%
1	0.0501855697541725\\
2	0.204561052862513\\
3	0.329167373493437\\
4	0.371595639502269\\
5	0.417712856046237\\
6	0.477531955998226\\
7	0.609384598971955\\
8	0.66530862005449\\
9	0.758369875477777\\
10	0.805297765386184\\
11	0.87232816841505\\
12	0.950027717213208\\
13	1.00526586457911\\
14	1.06546349643463\\
15	1.12418140023175\\
16	1.18382573160651\\
17	1.24110479858312\\
18	1.28699479521857\\
19	1.33351861593521\\
20	1.38224677825362\\
21	1.44114637217874\\
22	1.50383193063571\\
23	1.61289462476488\\
24	1.72492761641507\\
25	1.83428729722142\\
26	1.93479438895978\\
27	2.03091987094645\\
28	2.1231109013876\\
29	2.20009662284012\\
30	2.30448640001648\\
};
\addlegendentry{\scriptsize \cite{belkin2009constructing}: 2.30};

\addplot [color=mycolor3,solid,line width=1.5pt]
  table[row sep=crcr]{%
1	0.0633715018630028\\
2	0.11784852296114\\
3	0.205927446484566\\
4	0.273435175418854\\
5	0.346421092748642\\
6	0.436775058507919\\
7	0.492882013320923\\
8	0.568726539611816\\
9	0.630772590637207\\
10	0.692314565181732\\
11	0.760969698429108\\
12	0.840890049934387\\
13	0.925047278404236\\
14	1.00753748416901\\
15	1.08194637298584\\
16	1.15346896648407\\
17	1.22329390048981\\
18	1.28947007656097\\
19	1.36600410938263\\
20	1.43039906024933\\
21	1.49612128734589\\
22	1.54785215854645\\
23	1.617884516716\\
24	1.70040988922119\\
25	1.78223586082458\\
26	1.86034095287323\\
27	1.93552827835083\\
28	2.01321029663086\\
29	2.0690484046936\\
30	2.14226651191711\\
};
\addlegendentry{\scriptsize NN: 1.75};

\addplot [color=mycolor4,solid,line width=1.5pt]
  table[row sep=crcr]{%
1	0.0458786264061928\\
2	0.0780017077922821\\
3	0.108483962714672\\
4	0.124371893703938\\
5	0.167396619915962\\
6	0.192936167120934\\
7	0.204498112201691\\
8	0.244601801037788\\
9	0.25893771648407\\
10	0.27800714969635\\
11	0.289399445056915\\
12	0.32318514585495\\
13	0.344560235738754\\
14	0.371626317501068\\
15	0.387164235115051\\
16	0.402303457260132\\
17	0.414364695549011\\
18	0.428028672933578\\
19	0.445232331752777\\
20	0.460014820098877\\
21	0.482685655355453\\
22	0.503156363964081\\
23	0.528767764568329\\
24	0.561595439910889\\
25	0.585793077945709\\
26	0.603423893451691\\
27	0.621609032154083\\
28	0.631681561470032\\
29	0.642870128154755\\
30	0.655016899108887\\
};
\addlegendentry{\scriptsize \textbf{Ours: 0.67}};

\legend{}
\node[text width=0.5cm] at  (40,230) {\tiny F2};
\end{axis}
\end{tikzpicture}%
    \end{minipage}%

&
  \hspace{-0.9cm}
    \begin{minipage}{0.19\linewidth}
        \vspace{-0.25cm}
        
%
%
\definecolor{mycolor4}{rgb}{0.00000,0.44700,0.74100}%
\definecolor{mycolor5}{rgb}{0.85000,0.32500,0.09800}%
\definecolor{mycolor3}{rgb}{0.92900,0.69400,0.12500}%
\definecolor{mycolor1}{rgb}{0.49400,0.18400,0.55600}%
\definecolor{mycolor2}{rgb}{0.46600,0.67400,0.18800}%
\begin{tikzpicture}

\begin{axis}[%
width=0.7\linewidth,
height=0.5\linewidth,
at={(0.758in,0.481in)},
scale only axis,
xmin=1,
xmax=30,
ymin=0,
ymax=12,
xtick={10, 20, 30},
xticklabels={ 10, 20, 30},
ytick={0, 4, 8, 12},
yticklabels={ , ,  ,  },
xmajorgrids,
ymajorgrids,
every x tick label/.append style={font=\color{black}, font=\tiny},
every y tick label/.append style={font=\color{black}, font=\tiny},
title style={yshift=-0.75em},
axis background/.style={fill=white},
]
\addplot [color=mycolor1, line width=1.5pt]
  table[row sep=crcr]{%
1	0.693729400634766\\
2	1.2174072265625\\
3	1.77809715270996\\
4	2.28475570678711\\
5	2.77515602111816\\
6	3.25922584533691\\
7	3.70697212219238\\
8	4.17295837402344\\
9	4.65995025634766\\
10	5.13254356384277\\
11	5.61997413635254\\
12	6.08212661743164\\
13	6.5367374420166\\
14	6.96297073364258\\
15	7.34439086914063\\
16	7.71457290649414\\
17	8.06819534301758\\
18	8.41020393371582\\
19	8.74635314941406\\
20	9.06955146789551\\
21	9.38372039794922\\
22	9.69252967834473\\
24	10.2904186248779\\
26	10.8738441467285\\
30	11.9922180175781\\
};
\addlegendentry{LB\_pcl: 11.9922}

\addplot [color=mycolor2, line width=1.5pt]
  table[row sep=crcr]{%
1	0.372882843017578\\
2	0.721593856811523\\
3	1.03390693664551\\
4	1.34048652648926\\
5	1.63061141967773\\
7	2.17037200927734\\
9	2.67048454284668\\
11	3.17227363586426\\
14	3.86924743652344\\
16	4.2992000579834\\
18	4.69692802429199\\
20	5.08813858032227\\
21	5.27824401855469\\
22	5.46296119689941\\
23	5.64282989501953\\
24	5.81725692749023\\
27	6.32034111022949\\
29	6.65133094787598\\
30	6.8129711151123\\
};
\addlegendentry{Belkin et al.: 6.813}

\addplot [color=mycolor3, line width=1.5pt]
  table[row sep=crcr]{%
1	0.298196792602539\\
2	0.537958145141602\\
3	0.735723495483398\\
4	0.912574768066406\\
6	1.23458862304688\\
7	1.36493873596191\\
9	1.58269691467285\\
13	2.01141738891602\\
16	2.27913475036621\\
19	2.49384117126465\\
21	2.61651992797852\\
23	2.72602272033691\\
26	2.87446784973145\\
30	3.06630706787109\\
};
\addlegendentry{NN: 3.0663}

\addplot [color=mycolor4, line width=1.5pt]
  table[row sep=crcr]{%
1	0.236431121826172\\
2	0.402402877807617\\
3	0.547103881835938\\
4	0.681756973266602\\
5	0.799503326416016\\
6	0.89936637878418\\
7	0.985252380371094\\
10	1.20551490783691\\
12	1.34458160400391\\
13	1.41507339477539\\
15	1.53652572631836\\
16	1.59061050415039\\
19	1.73249435424805\\
22	1.85479736328125\\
26	2.00057792663574\\
30	2.13624000549316\\
};
\addlegendentry{Our: 2.1362}

\legend{}
\node[text width=0.5cm] at (40,110) {\tiny S};
\end{axis}
\end{tikzpicture}%

    \vspace{2.7cm}
    
      \hspace{-4cm}
      
        \begin{overpic}[trim=-100cm 100cm 0cm 0cm,clip,width=2\linewidth]{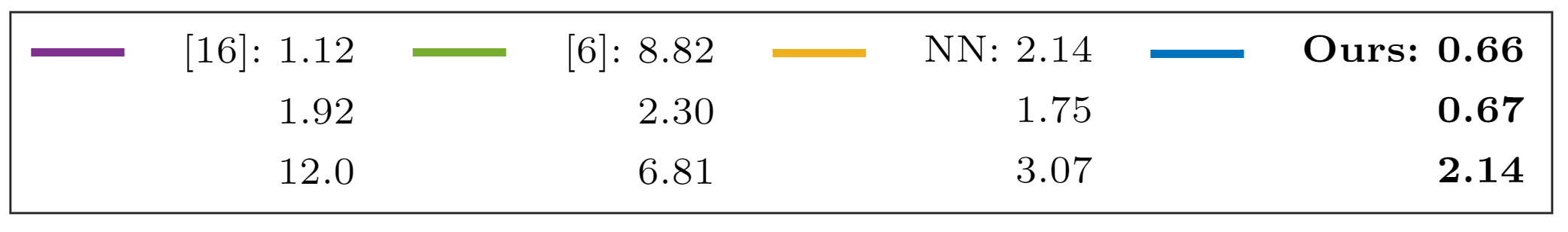}
        \put(4,53.5){\line(1,0){40}}
        \put(4,30){\line(0,1){23.5}}
        \put(4,30){\line(1,0){40}}
        \put(44,30){\line(0,1){23.5}}

        \put(4.5,47){\linethickness{1.5pt}\color{mycolor1}\line(2,0){3}}
        \put(4.5,42){\linethickness{1.5pt}\color{mycolor2}\line(2,0){3}}
        \put(4.5,37){\linethickness{1.5pt}\color{mycolor3}\line(2,0){3}}
        \put(4.5,32){\linethickness{1.5pt}\color{mycolor4}\line(2,0){3}}
        
        \put(8,46){\scriptsize [16]}
        \put(8,41){\scriptsize [6]}
        \put(8,36){\scriptsize NN}
        \put(8,31){\scriptsize \textbf{Our}}
        
        \put(17,50){\scriptsize F1}
        \put(26,50){\scriptsize F2}
        \put(36.5,50){\scriptsize S}
        
        \put(16,46){\scriptsize 1.12}
        \put(25,46){\scriptsize 1.92}
        \put(34,46){\scriptsize 12.00}
 
        \put(16,41){\scriptsize 8.82}
        \put(25,41){\scriptsize 2.30}
        \put(35,41){\scriptsize 6.81}       

        \put(16,36){\scriptsize 2.14}
        \put(25,36){\scriptsize 1.75}
        \put(35,36){\scriptsize 3.07} 
        
        \put(16,31){\scriptsize \textbf{0.66}}
        \put(25,31){\scriptsize \textbf{0.67}}
        \put(35,31){\scriptsize \textbf{2.14} }  
        \end{overpic}
        
    \end{minipage}%

\end{tabular}
\vspace{-1.4cm}

\caption{\label{fig:pcl}
Qualitative and quantitative evaluation of point cloud spectra estimation. On the left we show the qualitative comparison for different samplings on three classes (animals, human faces and objects). We show the eigenvalues estimations alongside the input point cloud (depicted as surface samplings), and the ground truth spectrum (in red). On the last two columns, we report the average cumulative error curves evaluated on the FLAME dataset for the two different distributions (F1 and F2) and on ShapeNet (S).
}
\end{figure*}
%

These properties allow us to use our network for the task of mesh super-resolution. Given a low-resolution mesh as input, our aim is to recover a higher resolution counterpart of it. Furthermore, while the input mesh has {\em arbitrary} resolution and is unknown to the network (and a correspondence with the training models is {\em not} given), an additional desideratum is for the new shape to be in dense point-to-point correspondence with models from the training set. We do so in a single shot, by predicting the decoded shape as:
\begin{align}\label{eq:hires}
    \X_\mathrm{hires} = D(\pi(\mathrm{Spec}(\X_\mathrm{lowres})))\,.
\end{align}
This simple approach exploits the resolution-independent geometric information encoded in the spectrum along with the power of a data-driven generative model. 

In Fig.~\ref{fig:different_resolution} we show a comparison with nearest-neighbors between eigenvalues (among shapes in the training set), and the isospectralization method of Cosmo~\etal~\cite{isosp}.
Our solution closely reproduces the high-resolution target. Isospectralization correctly aligns the eigenvalues, but it recovers unrealistic shapes due to ineffective regularization. This phenomenon highlights the following

\vspace{-1ex}
\begin{property}
Our data-driven approach replaces ad-hoc regularizers, that are difficult to model axiomatically, with \textbf{realistic priors} learned from examples.
\end{property}

\vspace{-1ex}
This is especially important for deformable objects; shapes falling into the same isometry class are often hard to disambiguate without using geometric priors.

\subsection{Estimating point cloud spectra}
\new{As an additional experiment}, we show how our network can directly predict Laplacian eigenvalues for unorganized point clouds. This task is particularly challenging due to the lack of a structure in the point set, and existing approaches such as \cite{clarenz2004finite,belkin2009constructing} often fail at approximating the eigenvalues of the underlying surface accurately. The difficulty is even more pronounced when the point sets are irregularly sampled, as we empirically show here.
In our case, estimation of the spectrum boils down to the single forward pass:
%
%
\begin{align}
    \widetilde{\mathrm{Spec}}(\X)=\rho(E(\X))\,.
\end{align}
%

To address this task we train our network by feeding unorganized point clouds as input, together with the spectra computed from the corresponding meshes (which are available at training time). As described in the supplementary materials, for this setting we use a PointNet \cite{qi2017pointnet} encoder and a fully connected decoder, and we replace the reconstruction loss of Eq.~\eqref{eq:rloss} with the Chamfer distance. \new{This application highlights the generality of our model, which can accommodate different representations of geometric data.}


We consider two types of point clouds: (1) with similar point density and regularity as in the training set {(shown in the supplementary materials)}, and (2) with randomized non-uniform sampling.
We compare the  spectrum  estimated via $\rho(E(\X))$ to axiomatic methods \cite{clarenz2004finite,belkin2009constructing}, and to the NN baseline (applied in the latent space); see Fig.~\ref{fig:pcl}.
The qualitative results are obtained by training on SMAL~\cite{SMAL} (left), COMA~\cite{COMA} (middle) and \new{ShapeNet watertight ~\cite{huang2018robust} (right)}. To highlight its generalization capability, the network trained on COMA is tested on point clouds from the FLAME dataset, while on ShapeNet we consider 4 different classes (airplanes, boats, screens and chairs). \new{We compute the cumulative error curves of the distance between the eigenvalues from the meshes corresponding to the test point clouds. 
 The mean error across all test sets is also reported in the legend. Our method leads to a significant improvement over the closest state-of-the-art baseline \cite{belkin2009constructing}. }
%
%
%

%
\begin{figure}[!t]
\centering
  \setlength{\tabcolsep}{0pt}
  \begin{tabular}{l c c c c r}
  \hspace{-0.4cm}
    \vspace{-0.2cm}
    
    \begin{minipage}{0.2\linewidth}        
    \input{figures/tikz/matching.tikz}
    
     \end{minipage}%
      &

 \hspace{0.7cm}
 
      \begin{minipage}{0.19\linewidth} 
       \vspace{0.4cm}
       
     \begin{overpic}[trim=0cm 0cm 0cm 0cm,clip,width=0.85\linewidth]{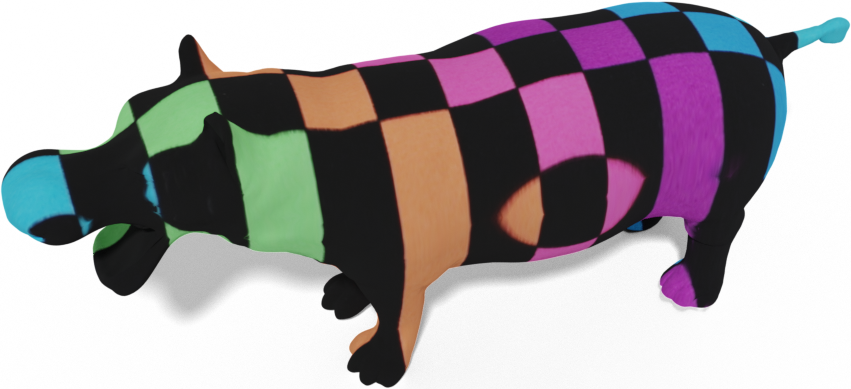}
     \put(-3.5,-100){\line(0,1){170}}
     \put(25,68){\footnotesize{Source}}
   \end{overpic}
   
     \vspace{0.6cm}
     
 \begin{overpic}[trim=0cm 0cm 0cm 0cm,clip,width=0.85\linewidth]{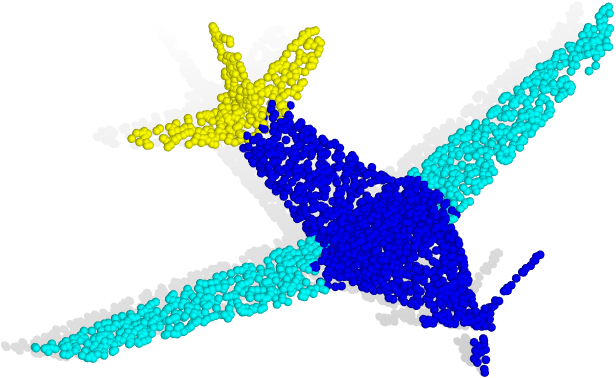}
   \end{overpic}
   
    \end{minipage}%
     
     &

    \begin{minipage}{0.19\linewidth}

     \begin{overpic}[trim=0cm 0cm 0cm 0cm,clip,width=0.95\linewidth]{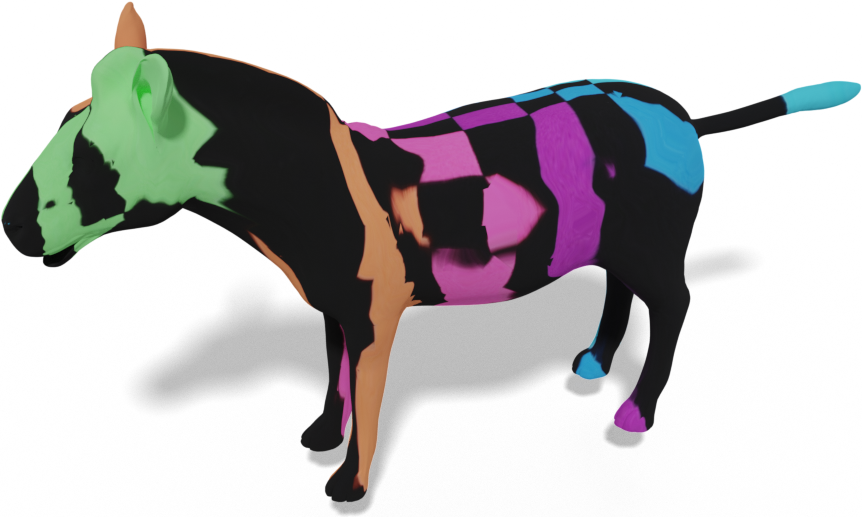}
     \put(33,60){\footnotesize{ICP}}
   \end{overpic}
 
 \begin{overpic}[trim=0cm 0cm 0cm 0cm,clip,width=0.95\linewidth]{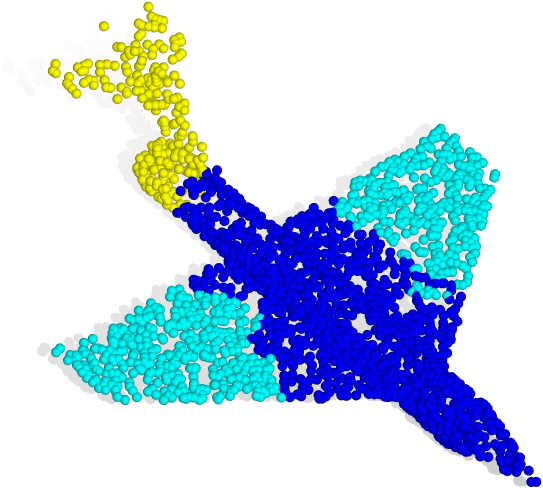}
 \put(20,-8){\footnotesize $75,18\%$}
 \put(19,-20.5){\footnotesize accuracy}

   \end{overpic}
   
 \end{minipage}%
 
       &

  \begin{minipage}{0.19\linewidth}
     \begin{overpic}[trim=0cm 0cm 0cm 0cm,clip,width=0.95\linewidth]{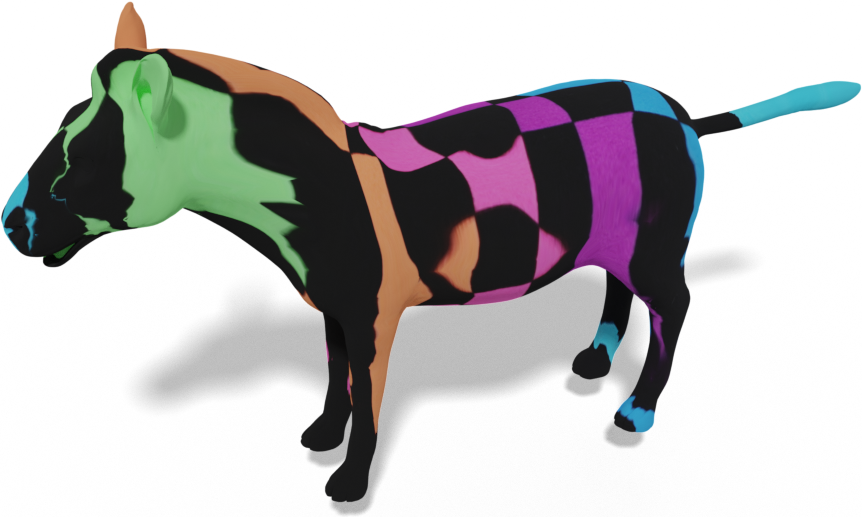}
     \put(33,60){\footnotesize{\textbf{Ours}}}
   \end{overpic}
 
 \begin{overpic}[trim=0cm 0cm 0cm 0cm,clip,width=0.95\linewidth]{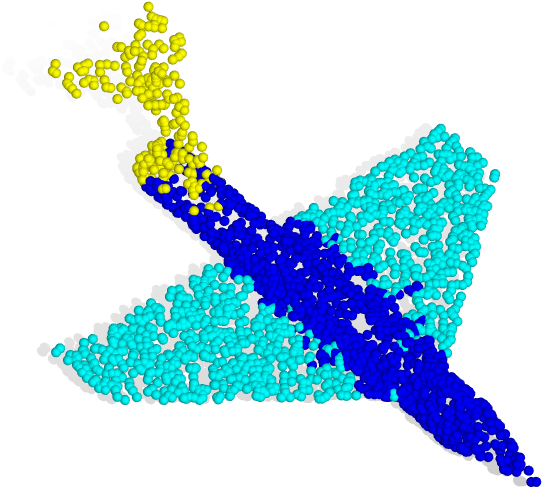}
\put(20,-8){\footnotesize $\mathbf{86,88\%}$}
 \put(19,-20.5){\footnotesize \textbf{accuracy}}

   \end{overpic}

    \end{minipage}%
 
    &
 
    \begin{minipage}{0.19\linewidth}
     
     \begin{overpic}[trim=0cm 0cm 0cm 0cm,clip,width=0.95\linewidth]{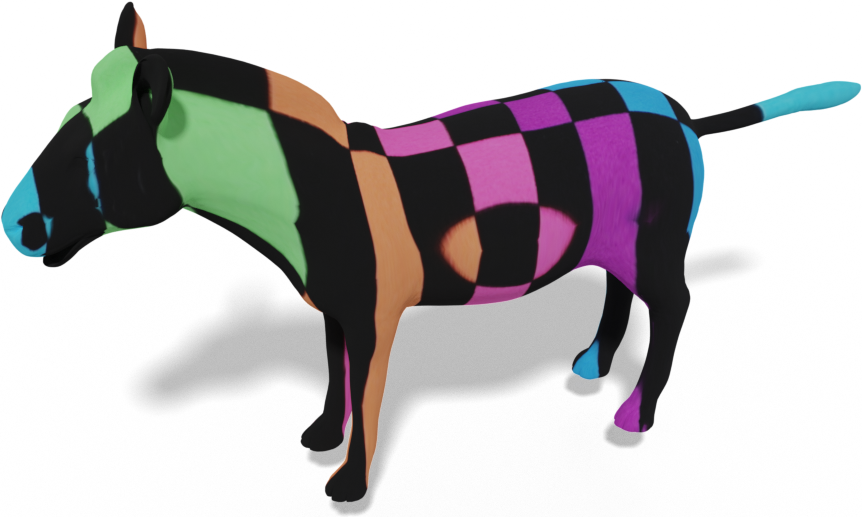}
     \put(33,60){\footnotesize{GT}}
   \end{overpic}
   
 \begin{overpic}[trim=0cm 0cm 0cm 0cm,clip,width=0.95\linewidth]{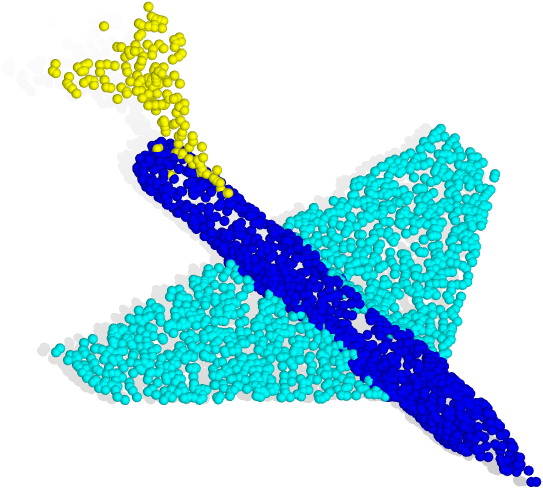}
   \end{overpic}
   
     \end{minipage}%
    
  \end{tabular}
  \vspace{0.4cm}
  
\caption{\label{fig:match}
On the left, quantitative evaluation of matching~\cite{kim2011blended} between 100 pairs of animals. On the right, the qualitative comparison on texture and segmentation transfer.
}

\end{figure}

\subsection{\new{Matching from spectrum}}
\vspace{-0.1cm}

\new{Finally, we compute dense correspondences between shape pairs using only their spectra. These are fed into our network; since the output points are naturally ordered by the decoder, we exploit this to establish a sparse correspondence. In the case of meshes, we extend it to a dense one by using the functional maps framework~\cite{ovsjanikov2012functional}. In the case of point clouds, we can propagate a semantic segmentation using nearest neighbors. 
We perform a quantitative evaluation on SMAL~\cite{SMAL}, testing on 100 non-isometric  pairs of animals from different classes. Two applications that benefit from our approach are texture and segmentation transfer; we tested them respectively on animals and segmented ShapeNet ~\cite{yi2017large}. The comparison baseline consists of 100 iterations of ICP~\cite{ICP} to rigidly align the two shapes followed by nearest-neighbor assignment as correspondence. See Fig.~\ref{fig:match} and the supplementary for further details.
}

\vspace{-0.1cm}

\section{Conclusions}
\vspace{-0.2cm}

We introduced the first data-driven method for shape generation from Laplacian spectra. Our approach consists in enriching a standard AE with a pair of cycle-consistent maps, associating ordered sequences of eigenvalues to latent codes and vice-versa. This explicit coupling brings forth key advantages of spectral methods to generative models, enabling novel applications and a significant improvement over existing approaches.
%
%
Our limitations are shared with other spectral methods in the computation of a robust Laplacian discretization.
Adopting the recent approach~\cite{Sharp:2019:NIT:3306346.3322979} for such borderline cases is a promising possibility.
%
%
Further, while the Laplacian is a classical choice due to its Fourier-like properties, spectra of other operators with different properties may lead to other promising applications.


%

{\small \vspace{1ex}\noindent\textbf{Acknowledgements} 
We gratefully acknowledge Luca Moschella and Silvia Casola for the technical support, Nicholas Sharp for the useful suggestions about pointcloud spectra.
Parts of this work were supported by the KAUST OSR Award No. CRG-2017-3426, the ERC Starting Grant No. 758800 (EXPROTEA), the ERC Starting Grant No. 802554 (SPECGEO), the ANR AI Chair AIGRETTE, and the MIUR under grant ``Dipartimenti di eccellenza 2018-2022'' of the Department of Computer Science of Sapienza University and University of Verona.}

{\small
\bibliographystyle{ieee}
\bibliography{egbib}
}

\clearpage
\newpage

\appendix
\renewcommand*{\thesection}{\arabic{section}}
\null

\threedvfinalcopy 
\def\threedvPaperID{81} 
\def\httilde{\mbox{\tt\raisebox{-.5ex}{\symbol{126}}}}

\ifthreedvfinal\pagestyle{empty}\fi

\appendix
\renewcommand*{\thesection}{\arabic{section}}
\null
\begin{center}
      \begin{tabular}[!t]{c}
          \centering  
    {\Large \bf Supplementary Materials \par}
      \vspace*{24pt}
      \large
      \lineskip .5em
      \end{tabular}

      \par
      
      \vskip .5em
      \vspace*{12pt}
\end{center}

\begin{abstract}
In this document we collect some additional details about the proposed method, architecture and results, that due to lack of space were not included in the main manuscript.
\end{abstract}

\section{Architecture}

\vspace{1ex}\noindent\textbf{{\em Meshes.}}
When $\X$ is discretized as a mesh (typical in graphics and geometry processing applications), the AE architecture is a multilayer perceptron (MLP) with a bottleneck and dimensions: $n\times 3 \veryshortarrow 300 \veryshortarrow 200 \veryshortarrow  30 \veryshortarrow  200  \veryshortarrow n\times 3$; each layer, except for the last one, is followed by $\tanh$ activation. The maps $\pi$ and $\rho$ are both parametrized by a MLP with dimensions: $k \veryshortarrow 80 \veryshortarrow 160 \veryshortarrow 320 \veryshortarrow 640 \veryshortarrow 320 \veryshortarrow 160 \veryshortarrow 80 \veryshortarrow k$; each layer, except for the last one, is followed by batch norm and $\mathrm{SeLu}$ activation.
We use batch size 16, and the Adam optimizer with learning rate equal to $10^{-4}$. Note that we also use this architecture for 2D contours, which are discretized as (2-regular) cycle graphs.

\vspace{1ex}\noindent\textbf{{\em Unorganized point clouds.}}
In this case we do {\em not} assume the training data to have a consistent vertex labeling. To tackle this setting, we use a PointNet~\cite{qi2017pointnet} encoder and a fully connected decoder\footnote{As suggested by the authors of~\cite{qi2017pointnet} in  \url{https://github.com/charlesq34/pointnet-autoencoder}}; we use the MLP from the mesh case for the decoder.
%
The encoder consists in a shared MLP network on each point with layer output sizes 64 and 128; each layer is followed by batch norm.
A maxpool layer is then used to output a 128-dimensional vector, which is reduced with a $\tanh$-activated MLP to dimensions: $128\veryshortarrow 64 \veryshortarrow 30$.

The full implementation of the proposed architecture and the data that we used in our experiments can be retrived here: \url{https://github.com/riccardomarin/InstantRecoveryFromSpectrum}.

\section{Analysis of increasing bandwidth ($k$)}
\label{sec:different_K}

We performed an experimental analysis for different values of $k$ (the number of Laplacian eigenvalues given as input). This parameter affects both the amount of information used by the network (\ie, more eigenvalues) and the capability of the learned representation (\ie, higher-dimensional latent space).
We evaluate our model with $k=15, k=30$ and $k=60$ on the shape-from-spectrum reconstruction task, using exactly the same dataset and experimental setup as in Table~1 of the main manuscript. 

\begin{figure}[!t]
  \centering
  \begin{overpic}
  [trim=0cm -4cm 0cm 0cm,clip,width=\linewidth]{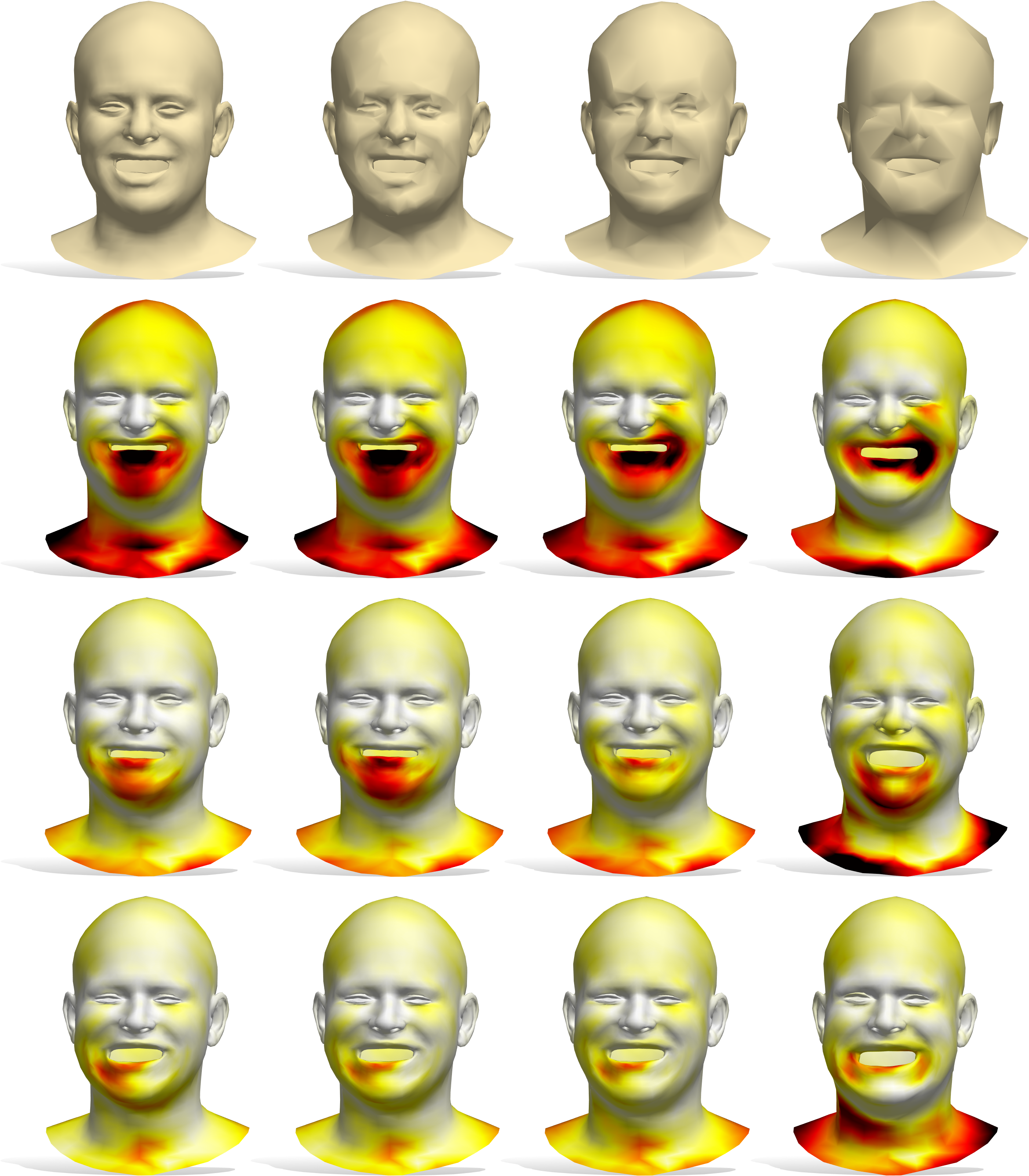}
    \put(8,101){\footnotesize full res}
    \put(30,101){\footnotesize 1000}
    \put(52,101){\footnotesize 500}
    \put(73.5,101){\footnotesize 200}
    
    \put(0,63){\footnotesize $15$}
    \put(0,38){\footnotesize $30$}
    \put(0,14){\footnotesize $60$}
  \end{overpic}
  
  \vspace{-0.0cm}
  
  \caption{\label{fig:K_qual}Examples of shape-from-spectrum reconstructions at different values for the spectral bandwidth $k$; see Table~\ref{tab:diff_k} for a quantitative evaluation.}
  
\end{figure}
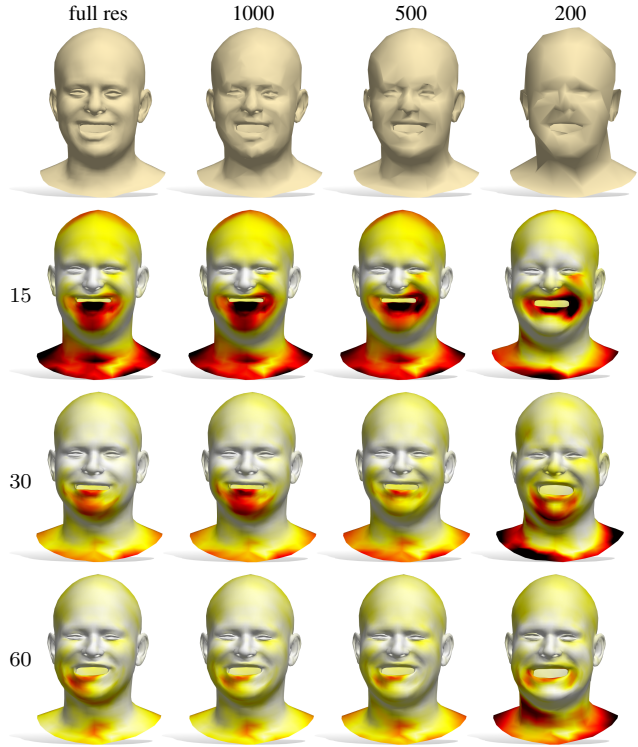

The quantitative results are reported in Table \ref{tab:diff_k}.
While cutting $k$ by half produces a sensible performance degradation, we observed that doubling it achieves just slightly better results. For this reason, in our experiments we settled for a value of $k=30$. From a qualitative perspective, our results suggest that without changing other architecture details, a larger value for $k$  can achieve better precision on the high-frequency details (\eg, the pose of the mouth); as shown in Fig.~\ref{fig:K_qual}.


\begin{table}[h!]
\begin{center}
\vspace{0cm}
\begin{tabular}{l|llll}
          & {\small full res}        & {\small 1000}    & {\small 500}     & {\small 200}       \\ \hline
          {\small $k  = 30$} & {\small $1.61$} &  {\small $1.62$} &  {\small $\mathbf{1.71}$} &  {\small $2.13$} \\
          {\small  $k=15$}   &  {\small $3.74$} &  {\small $3.78$} &  {\small $3.72$} &  {\small $3.59$}  \\
          {\small  $k=60$} &  {\small $\mathbf{1.60}$} &  {\small $\mathbf{1.52}$} &  {\small $1.79$} &  {\small $\mathbf{2.05}$}
\end{tabular}
\vspace{0.2cm}
\caption{\label{tab:diff_k}Shape-from-spectrum reconstruction comparisons with different $k$ trained for the same number of epochs; we report average error over 100 shapes of an unseen subject from the COMA dataset~\cite{COMA}. All errors must be rescaled by $10^{-5}$. }
\end{center}
\end{table}

\section{Laplace-Beltrami operator discretization}
\label{sec:FEM}

\begin{figure}[h!]
  \centering
  \begin{overpic}
  [trim=0cm -4cm 0cm 0cm,clip,width=\linewidth]{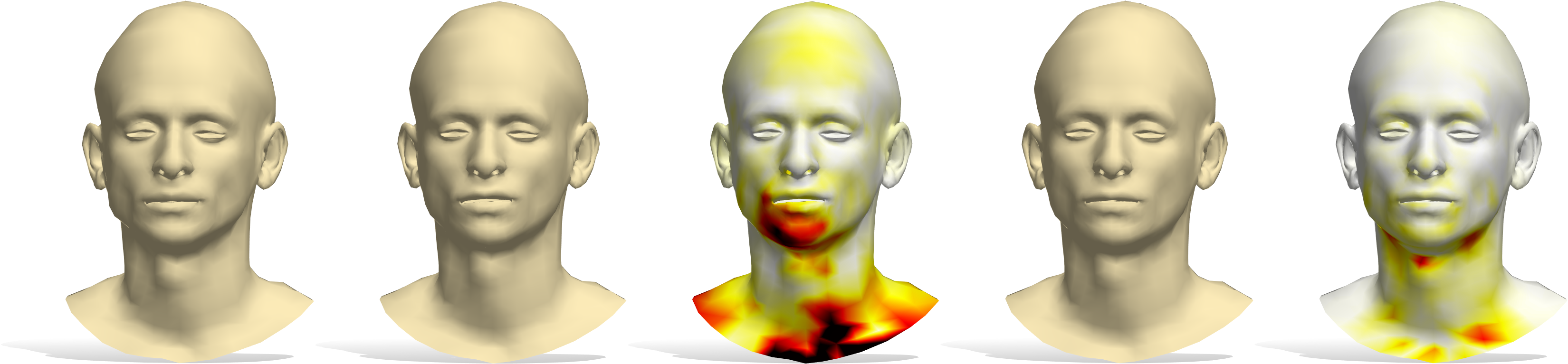}
  \end{overpic}
  
  \begin{overpic}
  [trim=0cm -4cm 0cm 0cm,clip,width=\linewidth]{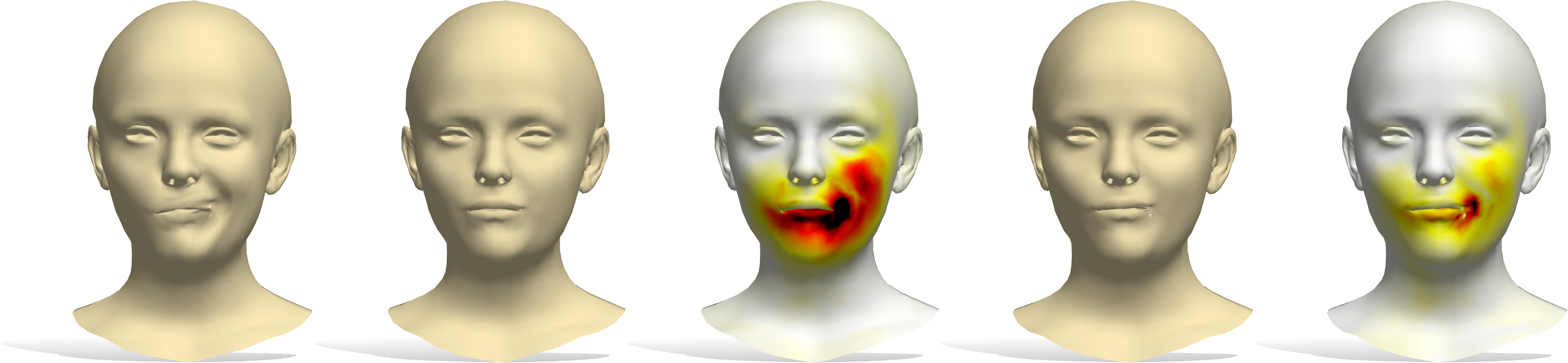}
  \end{overpic}
  
  \begin{overpic}
  [trim=0cm -4cm 0cm 0cm,clip,width=\linewidth]{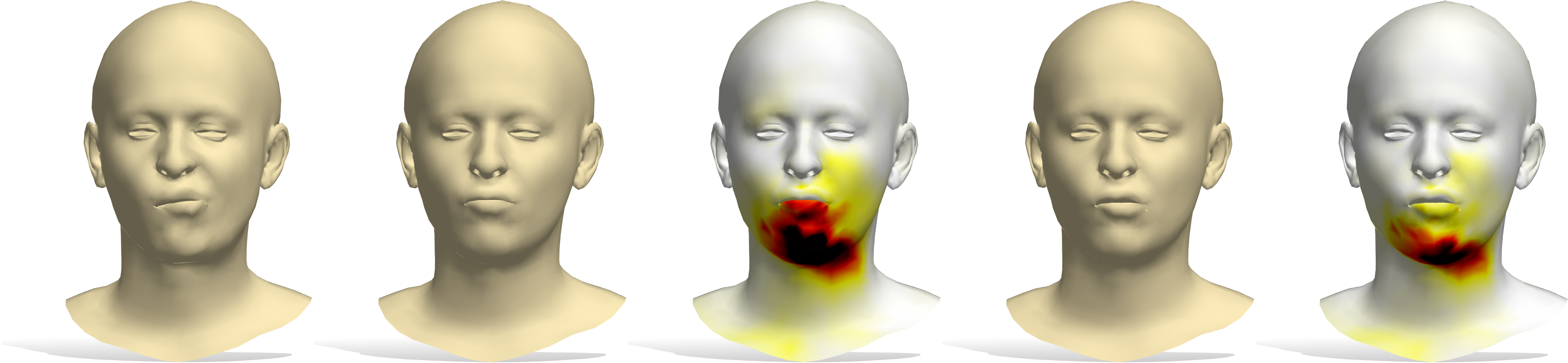}
  \put(5.2,-0.4){\footnotesize unknown}
  \put(7.8,-3.4){\footnotesize target}
  \put(34.3,-0.7){\footnotesize linear FEM}
  \put(74.3,-0.7){\footnotesize cubic FEM}
  \end{overpic}
  \vspace{0.05cm}
  
  \caption{\label{fig:FEM2}Additional comparisons (one per row) between the use of eigenvalues of the Laplacian discretized with linear FEM or with cubic FEM in our shape-from-spectrum pipeline. The heatmap encodes point-wise reconstruction error, growing from white to dark red. ``Unknown target'' in the left-most column refers to the source shape from which the eigenvalues are computed.}
\end{figure}

One of the advantages of our pipeline is that we can use the {\em cubic} FEM (\eg \cite[Sec. 4.1]{reuter2010hierarchical}) to discretize the Laplace-Beltrami operator, virtually without any additional cost if compared with the linear finite elements.
The cubic FEM yields a more accurate discretization and improves the quality of the results produced by our pipeline. In Fig.~\ref{fig:FEM2} we collect additional comparisons between the two different discretizations.

%

\begin{figure}[t!]
    \includegraphics[width=0.95\linewidth]{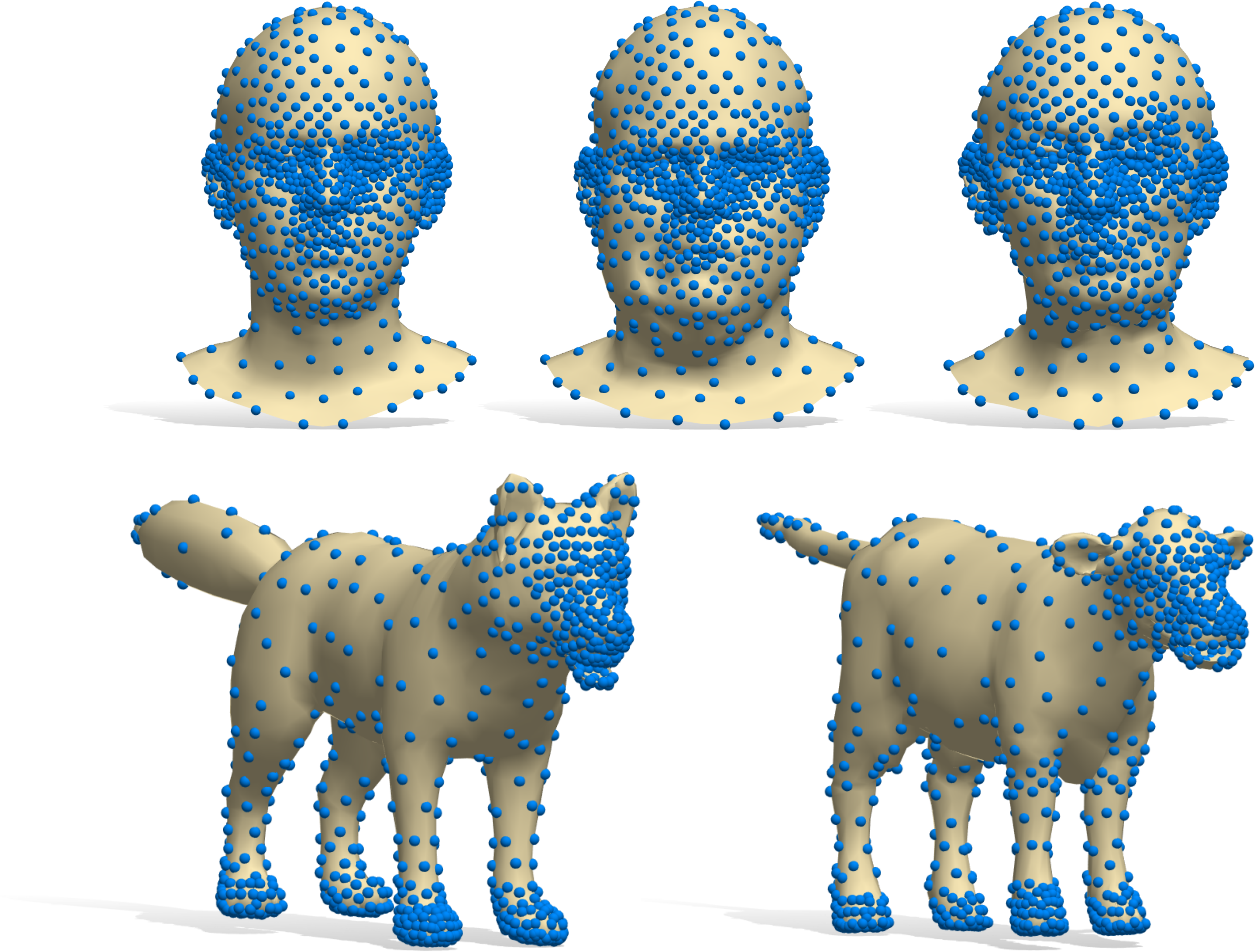}
    \caption{Examples of point clouds used to train the PointNet version of our network in case of CoMa (first row) and SMAL (second row).}
    \label{fig:training-sets}
\end{figure}

\section{Training and test details}
\label{sec:datadetails}

Here we summarize the data involved in the trainings and in the experiments. In the code that we will release upon acceptance, we will attach pre-processing scripts and when possible the data used to permit replication.

Overall, we perform five different trainings: for CoMa and SMAL datasets we train the dense and the PointNet version of our Network on the same set of data. We obtained the point clouds for PointNet trainings with a sparse sampling over the original meshes (the number of sampled point is around $20\%$ of the mesh vertices at full resolution): some examples of training point clouds are shown in Figure \ref{fig:training-sets}. For the ShapeNet dataset, only the PointNet version is trained using $500$ points for shape.  The applications use these five models, and we do not do any ad-hoc optimization for different tasks.

\begin{figure*}[t!]
    \centering
    \vspace{0.5cm}
    \begin{overpic}
        [trim=0cm 0cm 0cm 0cm,clip,width=0.8\linewidth]{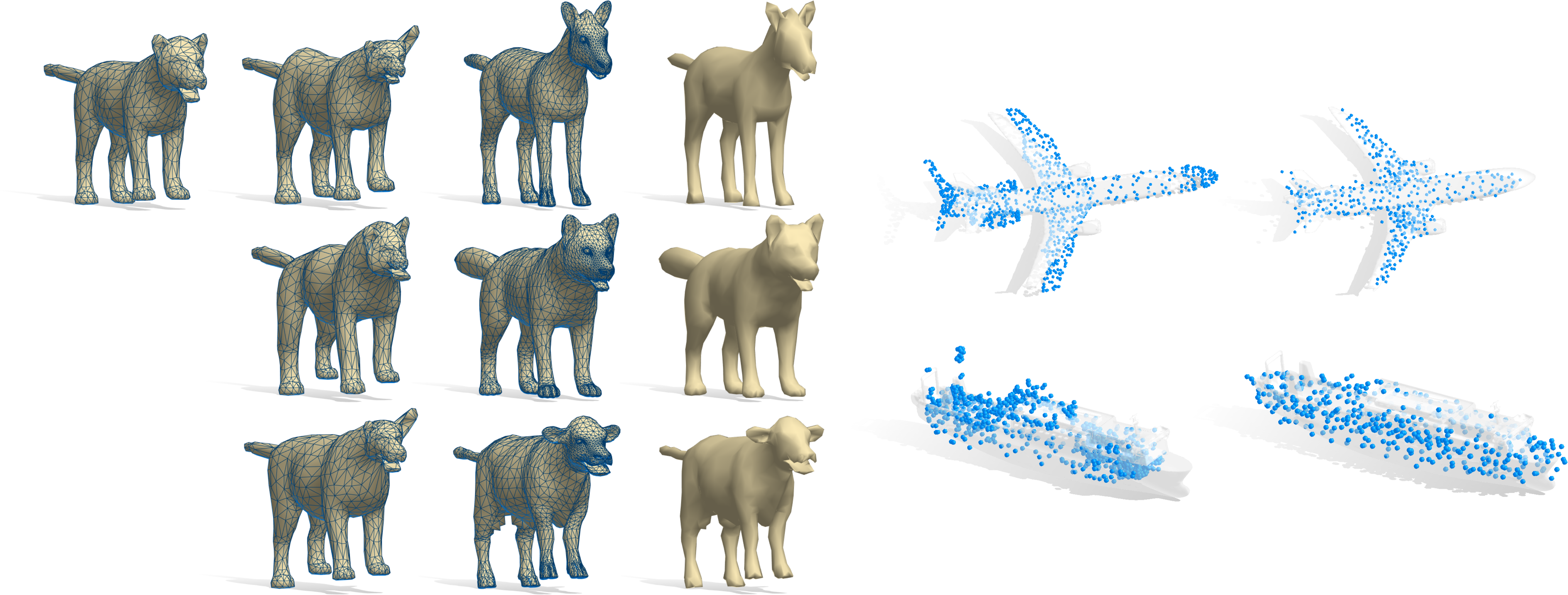}
        \put(7,38){\footnotesize Init.}
        \put(16,38){\footnotesize Cosmo \etal~\cite{isosp}}
        \put(32,38){\footnotesize \textbf{Ours}}
        \put(44.5,38){\footnotesize Target}
        \put(61.5,33){\footnotesize Cosmo \etal~\cite{isosp}}
        \put(85.5,33){\footnotesize \textbf{Ours}}
        \put(55,0){\line(0,1){41}}
    \end{overpic}
    \caption{Comparison in shape from spectrum estimation between isospectralization \cite{isosp} and our method in the case of meshes (left) and point clouds (right). The \textit{Init.} shape on the left is the neutral shape of the SMAL\cite{SMAL} generative model used as initialization of isospectralization algorithm for animals. The target shapes in the point cloud case are shown under the point clouds.}
    \label{fig:isospec_bad} 
\end{figure*}
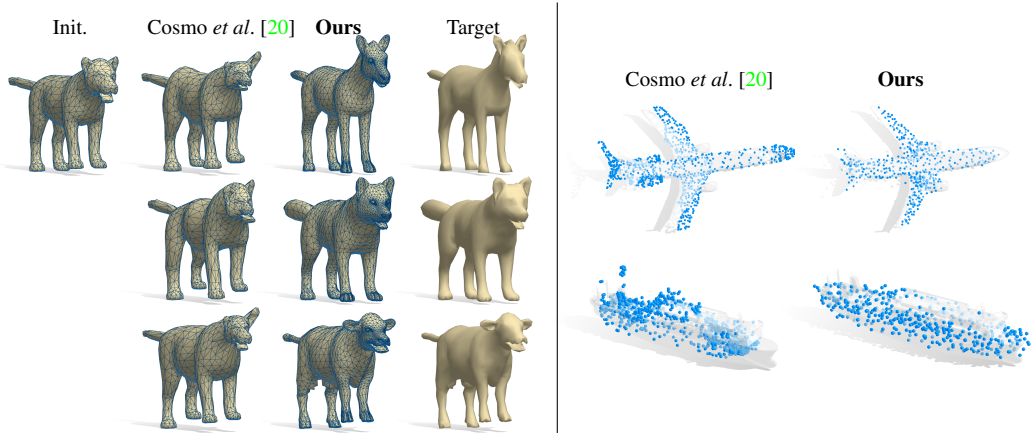
Concerning the involved training sets:
\vspace{1ex}

\noindent\textbf{{\em CoMa.}} We use 1,853 shapes randomly picked between 11 subjects from the CoMa dataset~\cite{COMA} (168 shapes for each subject), leaving out the 12th subject for testing purpose.
\vspace{1ex}

\noindent\textbf{{\em SMAL.}} We use 4,430 animal meshes generated with the SMAL~\cite{SMAL} generative model, among all the five possible classes. Thanks to the method explained in the original paper, we can cluster them in the correct races while we explore the inter-class variety.
\vspace{1ex}

\noindent\textbf{{\em ShapeNet.}} We would remark that ShapeNetCore dataset~\cite{shapenet2015} is not practical for standard spectral analysis since the models consist of non-manifold disconnected pieces. For this reason, we used a processed watertight version of it~\cite{huang2018robust}. From it, we pruned the meshes that still have artifacts (e.g. disconnected components). Finally, we remeshed the remaining ones to have at maximum 10K vertices. For training, we relies on four different classes: airplanes, boats, screens, and chairs. We pick 2,047 meshes from these classes, among the ones with a more reliable spectrum (i.e. removing outliers case). This last property was not considered for the test set, to challenge ourselves also against non-standard models (as you can see in Figure 11 of the main manuscript, where the target airplane has a tail's part missing).

Concerning the quantitative experiments:
\vspace{1ex}

\noindent\textbf{{\em Shape from spectrum.}} This experiment was performed on a total of $400$ shapes; $100$ shapes of the unseen subject from the CoMa dataset at four different resolutions.
\vspace{1ex}

\noindent\textbf{{\em Estimating point cloud spectra.}} Here we consider a total of $1,186$ models for the test: $393$ shapes of different subjects from FLAME~\cite{FLAME} dataset at two different samplings, and $400$ shapes from ShapeNet from the four classes.

\noindent\textbf{{\em Shape matching.}} Finally, for the shape matching, we consider $100$ couples of random animals of different species. These animals are generated with the same method of the above training set, but with higher variance in the generative space of SMAL. In this way, the testset presents more extreme characteristics of the ones seen at training time. For matching, we mainly consider extreme cases where high non-isometric deformation occurs (e.g.as shown in Figure 11 between hippos and horses).
As a final comment, we would emphasize that our results against Nearest-Neighbor suggest our generalization capability and the real challenge of our testsets.

\section{Shape from spectrum}
\label{sec:shape-from-spec}
In the task of recovering shapes from Laplacian spectra we compared with the \textit{isospectralization} approach introduced in \cite{isosp}. In Fig. \ref{fig:isospec_bad}, we report some qualitative examples from SMAL dataset \cite{SMAL} and ShapeNet dataset \cite{huang2018robust}. For SMAL, both the unknown shape to be recovered and our reconstruction are meshes (left panel of Fig. \ref{fig:isospec_bad}); for objects from ShapeNet, our network has been trained on unorganized sparse point clouds, so the reconstruction in this case is a point cloud as well (right panel of Fig. \ref{fig:isospec_bad}). Our method recovers shape from spectrum in a \textit{single forward pass}, while isospectralization \textit{deforms} an initial shape such that its spectrum align the one of a target shape using geometric regularizers. As initialization for isospectralization we used the template of SMAL \cite{SMAL} for animals, and the closest (in terms of distance between the eigenvalues) shape from our training set for objects of ShapeNet. Even so, isospectralization produces unrealistic instances or fails to capture the geometry of the target compared to our method.

%
\begin{figure*}[t!]
\begin{center}
  \vspace{0.3cm}

  \begin{overpic}
  [trim=0cm 0cm 0cm 0cm,clip,width=0.31\linewidth]{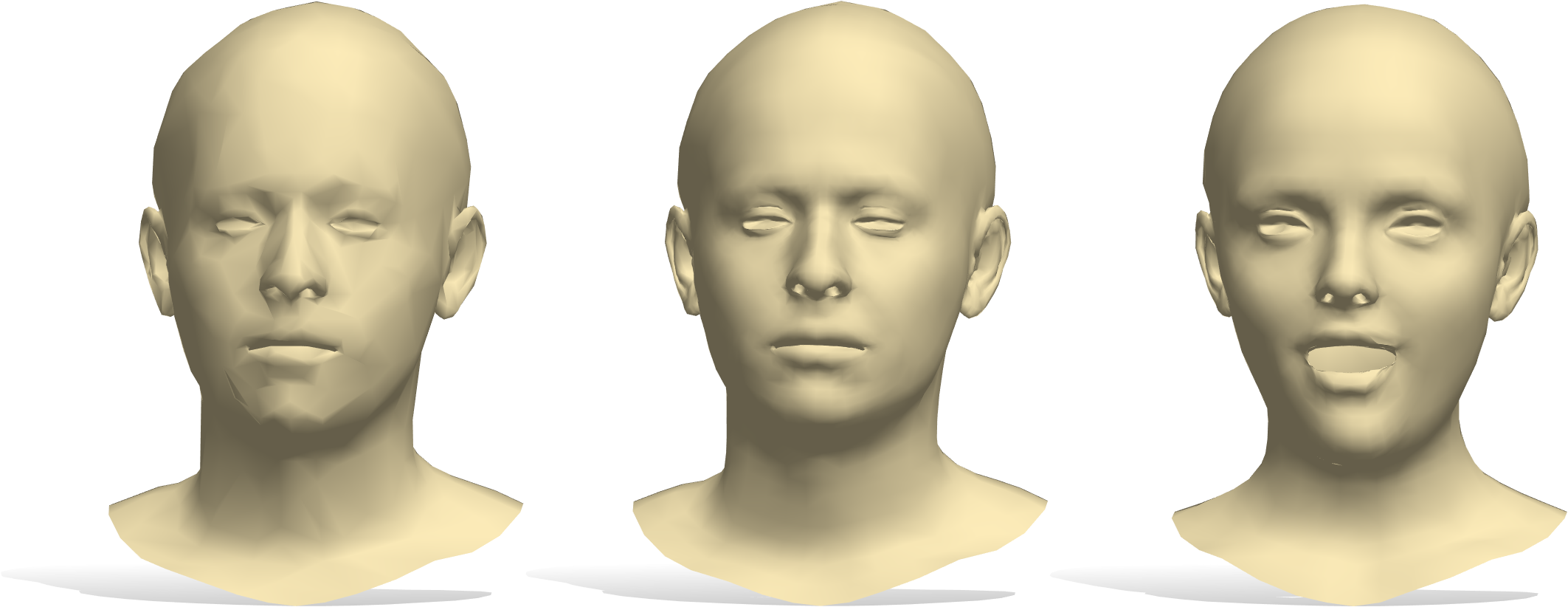}
    \put(11,41){\footnotesize \style{Target}}
    \put(46,41){\footnotesize \textbf{\our{Ours}}}
    \put(83,41){\footnotesize NN}
    \put(-3,15){\rotatebox{90}{\footnotesize 1000}}
  \end{overpic}
  \hspace{0.15cm}
  \begin{overpic}
  [trim=0cm 0cm 0cm 0cm,clip,width=0.31\linewidth]{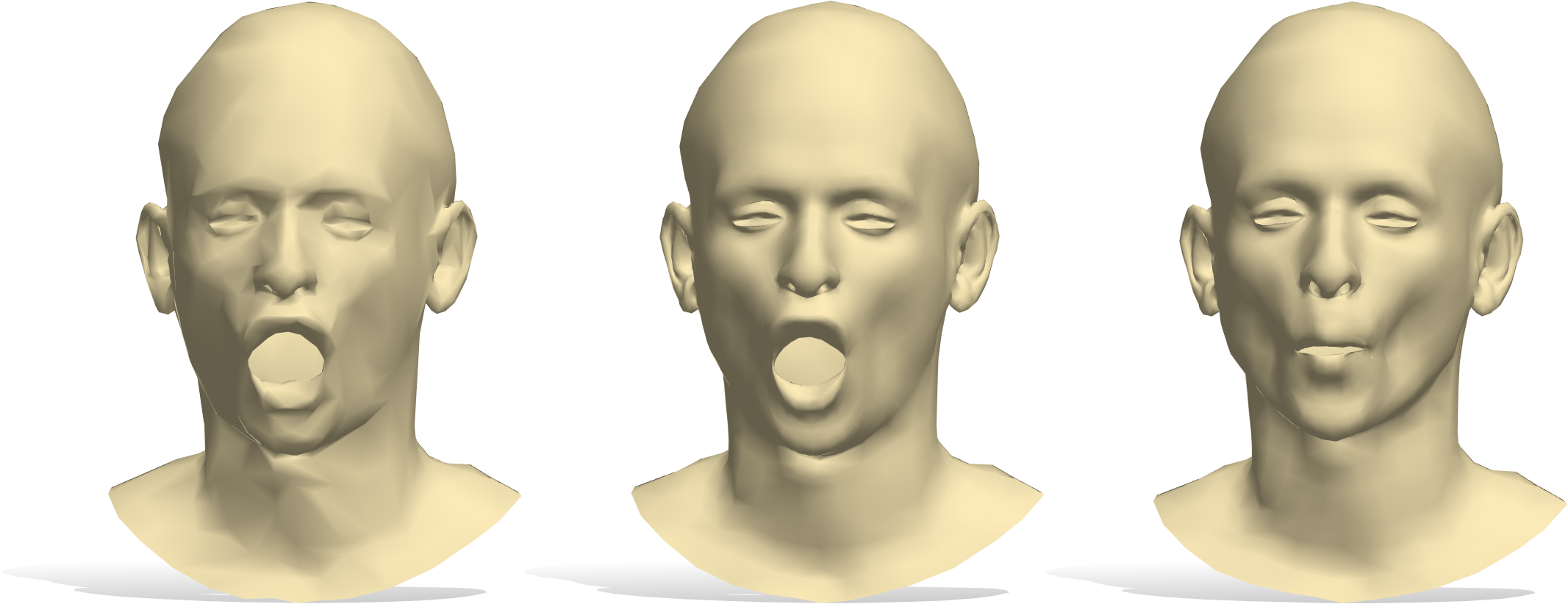}
    \put(11,41){\footnotesize \style{Target}}
    \put(46,41){\footnotesize \textbf{\our{Ours}}}
    \put(83,41){\footnotesize NN}
  \end{overpic}
  \hspace{0.15cm}
  \begin{overpic}
  [trim=0cm 0cm 0cm 0cm,clip,width=0.31\linewidth]{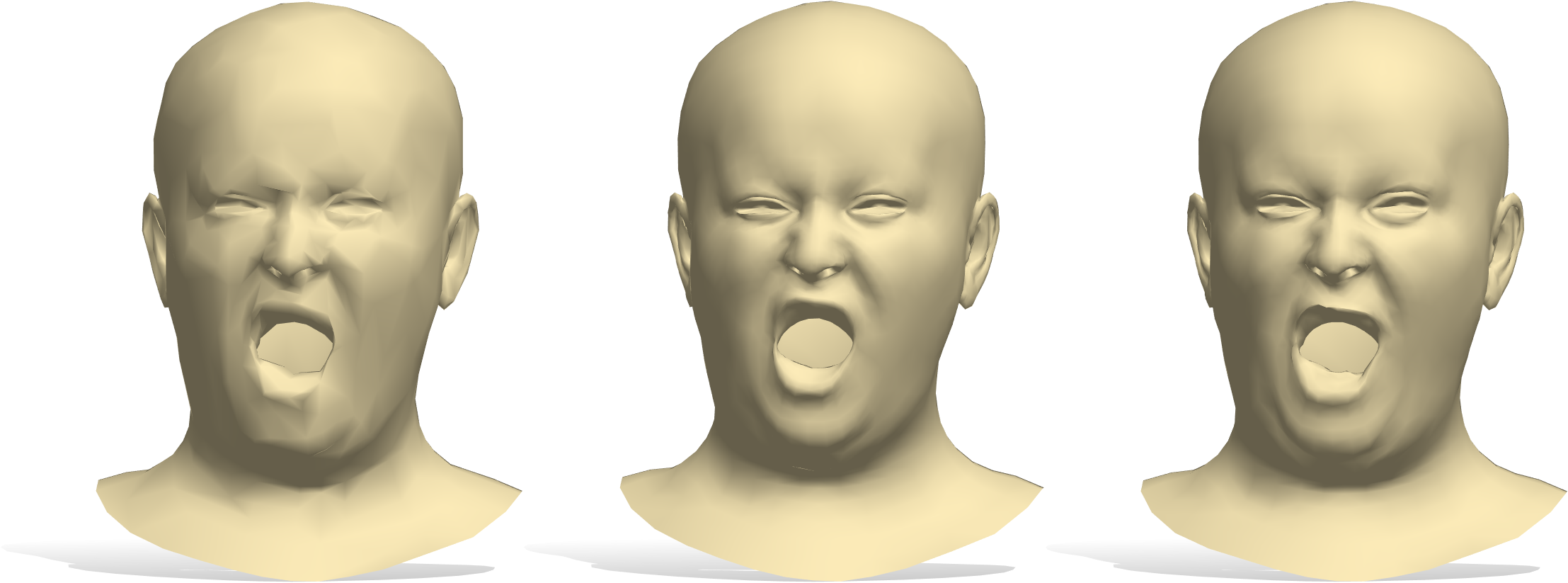}
    \put(11,41){\footnotesize \style{Target}}
    \put(46,41){\footnotesize \textbf{\our{Ours}}}
    \put(83,41){\footnotesize NN}
  \end{overpic}
  
  \begin{overpic}
  [trim=0cm 0cm 0cm 0cm,clip,width=0.31\linewidth]{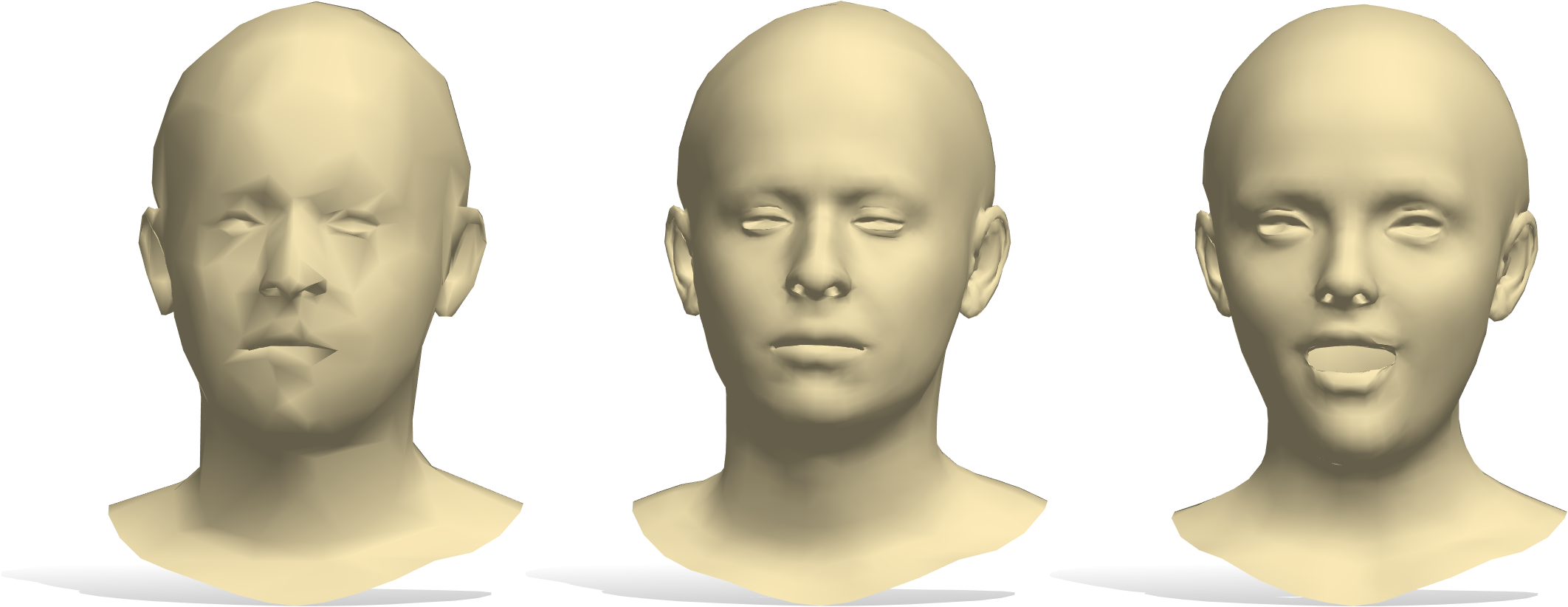}
      \put(-3,15){\rotatebox{90}{\footnotesize 500}}
  \end{overpic}
  \hspace{0.15cm}
  \begin{overpic}
  [trim=0cm 0cm 0cm 0cm,clip,width=0.31\linewidth]{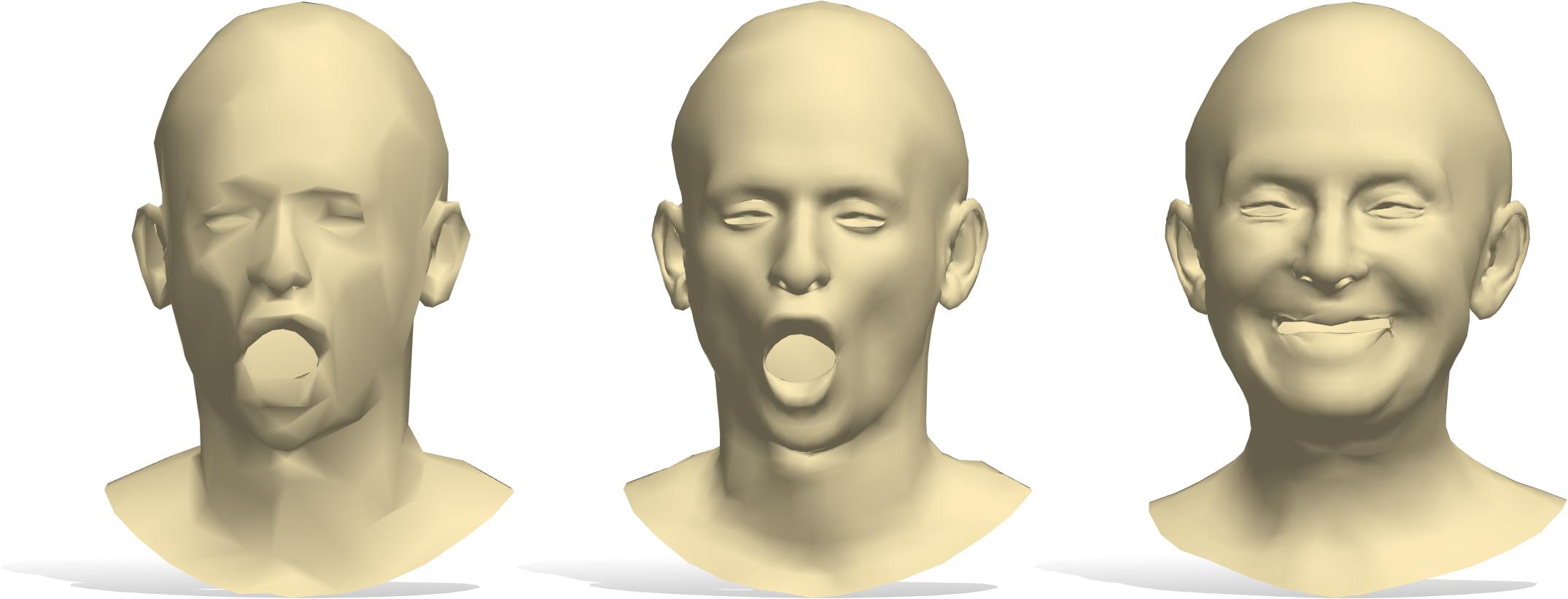}
  \end{overpic}
  \hspace{0.15cm}
  \begin{overpic}
  [trim=0cm 0cm 0cm 0cm,clip,width=0.31\linewidth]{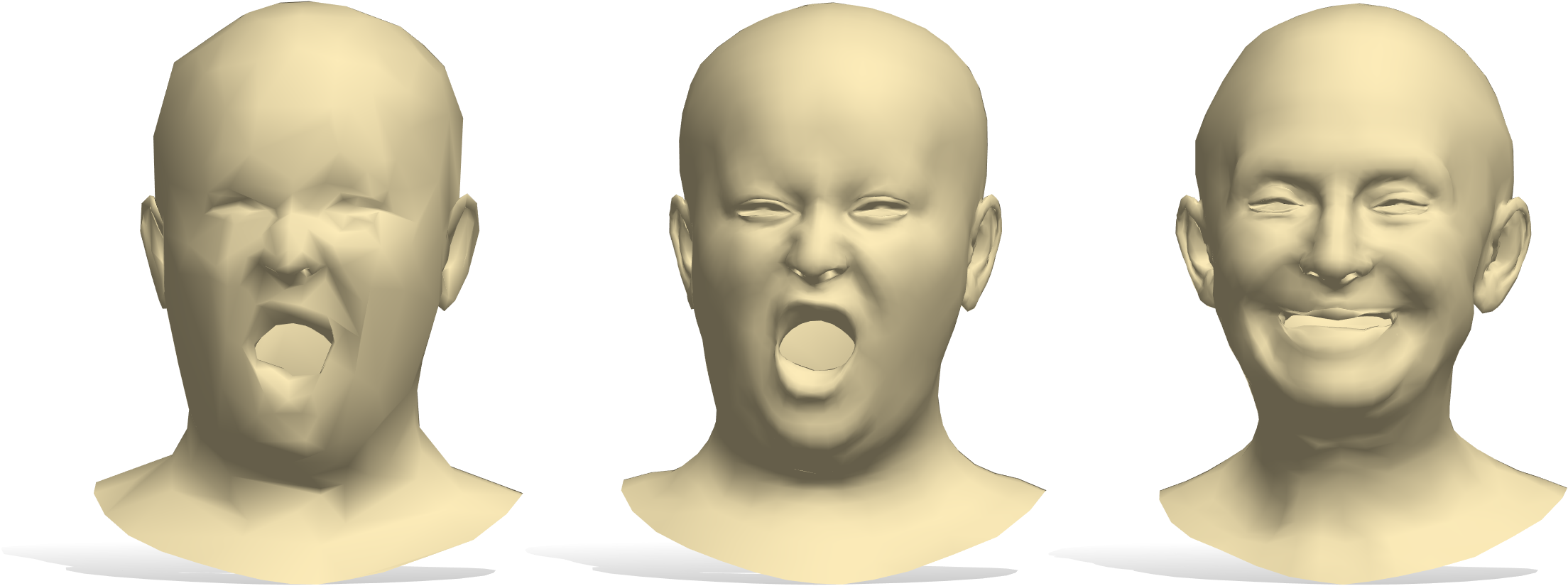}
  \end{overpic}
  
  \begin{overpic}
  [trim=0cm 0cm 0cm 0cm,clip,width=0.31\linewidth]{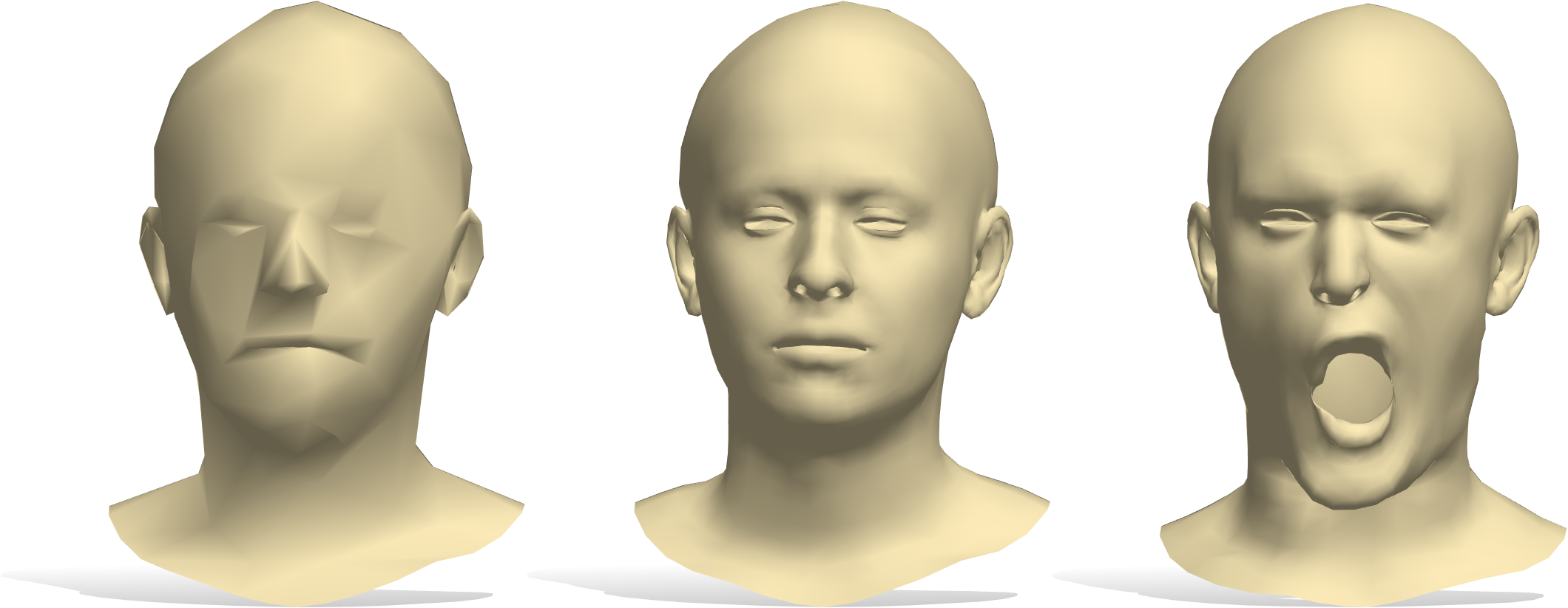}
    \put(-3,15){\rotatebox{90}{\footnotesize 200}}
    \put(107,0){\color{black}\line(0,1){125}}
  \end{overpic}
  \hspace{0.15cm}
  \begin{overpic}
  [trim=0cm 0cm 0cm 0cm,clip,width=0.31\linewidth]{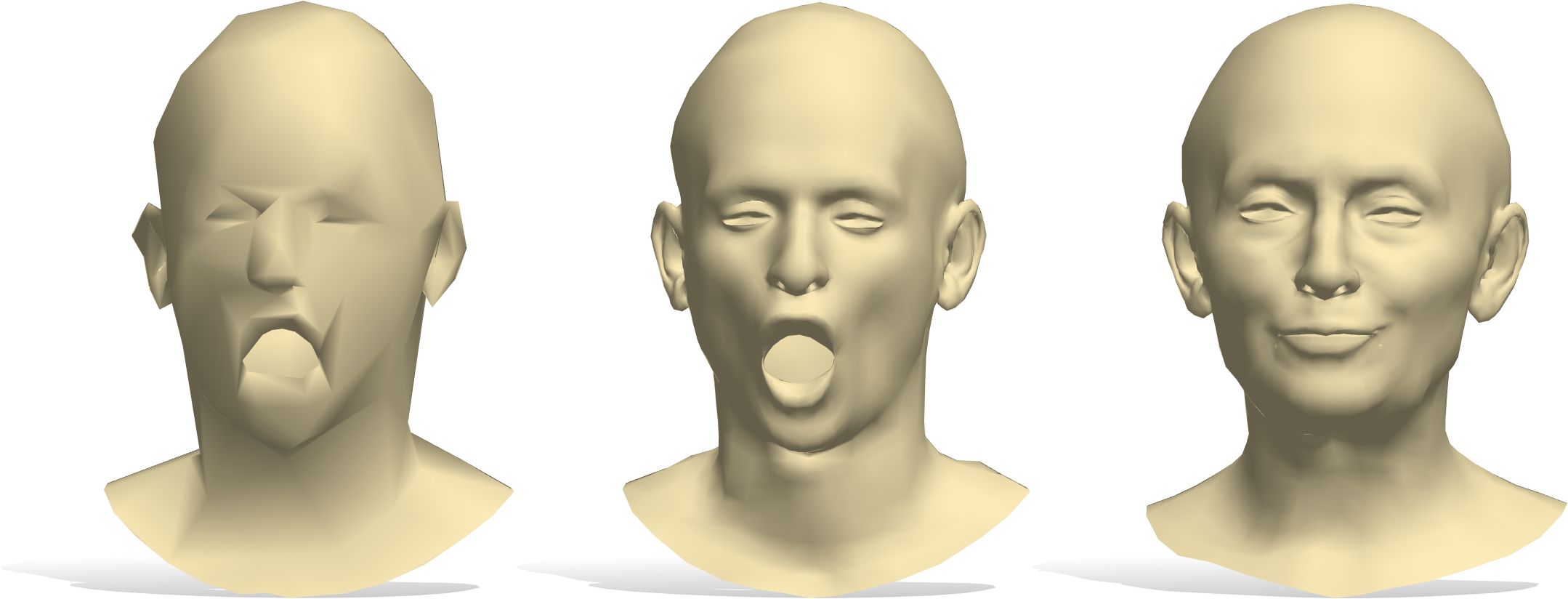}
    \put(107,0){\color{black}\line(0,1){125}}
  \end{overpic}
  \hspace{0.15cm}
  \begin{overpic}
  [trim=0cm 0cm 0cm 0cm,clip,width=0.31\linewidth]{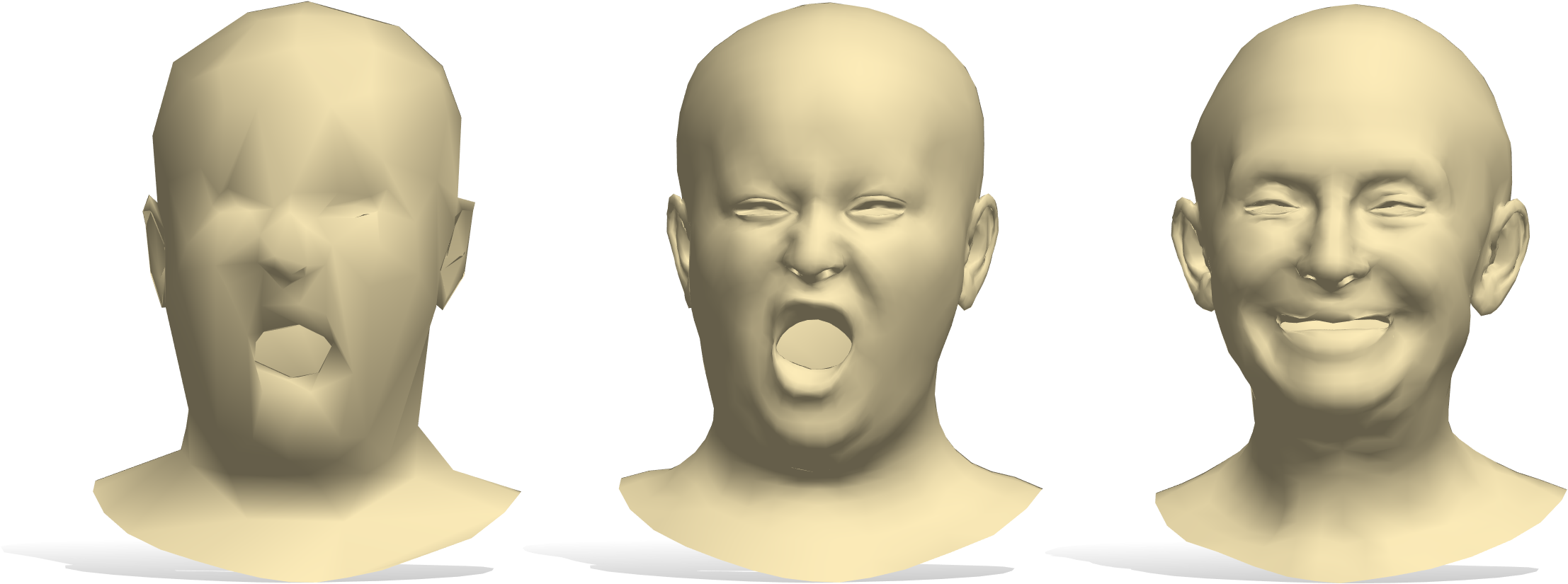}
  \end{overpic}

  \vspace{0.02cm}
  \caption{\label{fig:remesh2} Mesh super-resolution for input shapes at decreasing resolution (top to bottom, target shapes have respectively $1000$, $500$ and $200$ vertices). Our solutions match closely with the original high-resolution version of the input shapes, while the nearest neighbor baseline (NN) predicts the wrong identity or pose.}
  \end{center}
\end{figure*}

\begin{figure*}[h!]
\centering
\vspace{0.3cm}

  \begin{overpic}[trim=0cm 0cm 0cm 0cm,clip,width=0.48\linewidth]{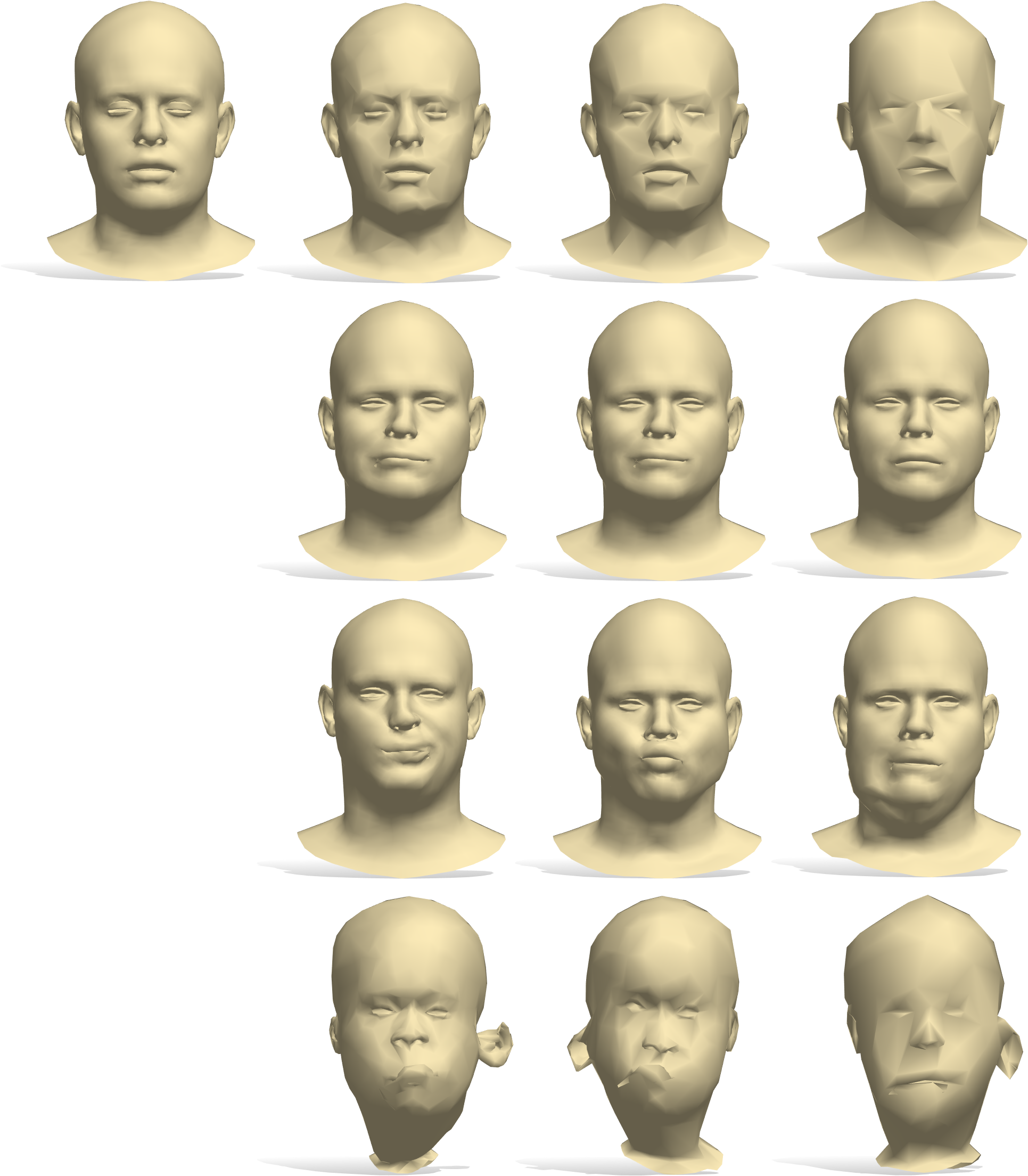}
    \put(2.5,85.5){\rotatebox{90}{\footnotesize target}}
    \put(7.5,100.5){\footnotesize $\sim$4000}
    \put(31.5,100.5){\footnotesize 1000}
    \put(54,100.5){\footnotesize 500}
    \put(76.2,100.5){\footnotesize 200}
    \put(24,86){\rotatebox{90}{\footnotesize input}}
    \put(24,4){\rotatebox{90}{\footnotesize Cosmo~\cite{isosp}}}
    \put(24,36){\rotatebox{90}{\footnotesize NN}}
    \put(24,61){\rotatebox{90}{\footnotesize \textbf{ours}}}
  \end{overpic}
 \hspace{0.2cm}
\vspace{1.2cm}
\begin{overpic}[trim=0cm 0cm 0cm 0cm,clip,width=0.48\linewidth]{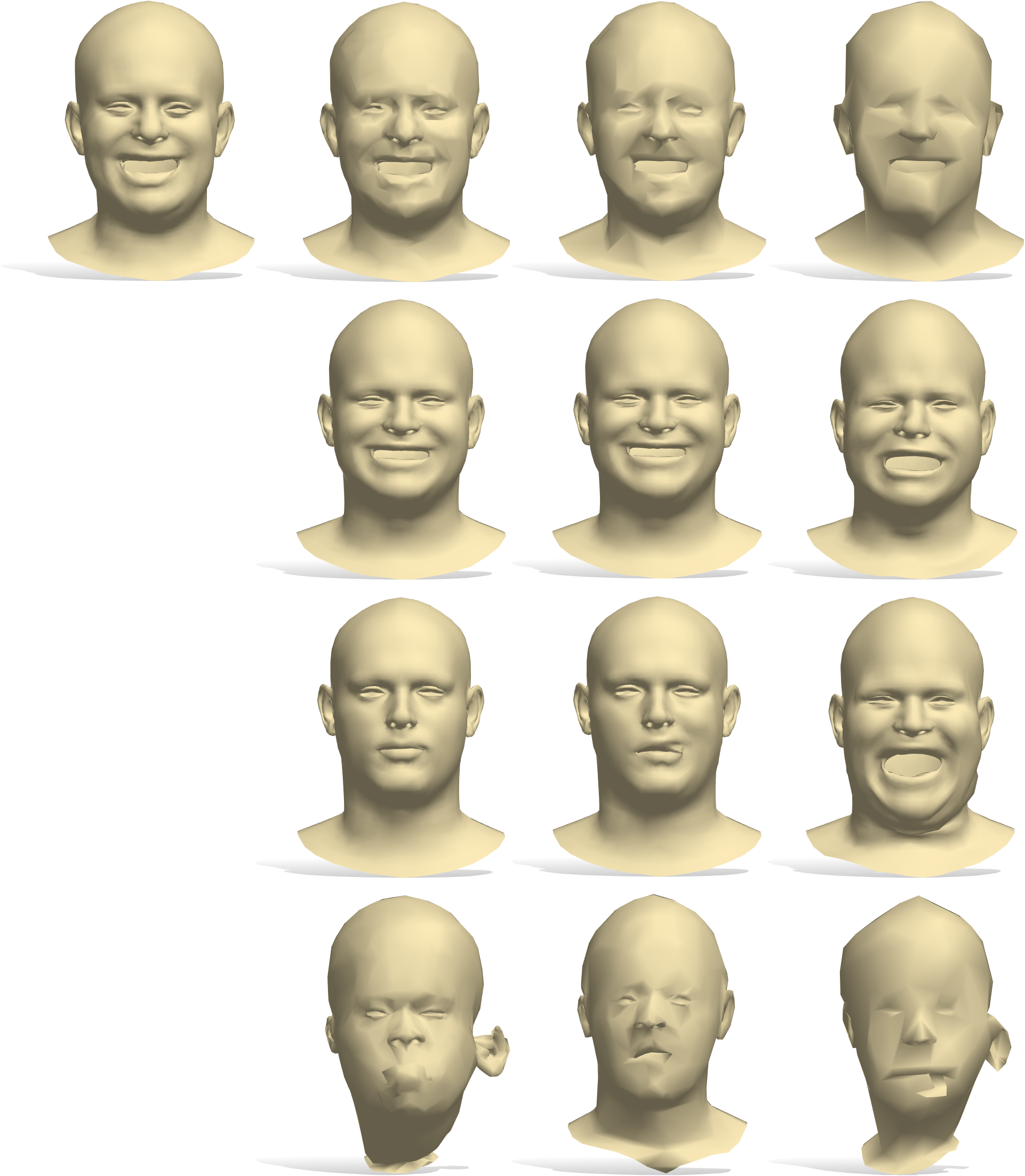}
    \put(2.5,85.5){\rotatebox{90}{\footnotesize target}}
    \put(7.5,100.5){\footnotesize $\sim$4000}
    \put(31.5,100.5){\footnotesize 1000}
    \put(54,100.5){\footnotesize 500}
    \put(76.2,100.5){\footnotesize 200}
    \put(24,86){\rotatebox{90}{\footnotesize input}}
    \put(24,4.5){\rotatebox{90}{\footnotesize Cosmo~\cite{isosp}}}
    \put(24,36){\rotatebox{90}{\footnotesize NN}}
    \put(24,61){\rotatebox{90}{\footnotesize \textbf{ours}}}
  \end{overpic}
\vspace{-1cm}
\caption{\label{fig:remesh3}Two mesh super-resolution examples for input shapes at decreasing resolution (top row of each of the two images, left to right). Our solutions match closely with the original high-resolution version of the input shapes (top left), while other approaches either predict the wrong pose (NN baseline) or generate an unrealistic shape (Cosmo~\etal).}
\end{figure*}
\begin{figure*}[t!]
\begin{center}
  \begin{overpic}
  [trim=0cm 0cm 0cm 0cm,clip,width=0.49\linewidth]{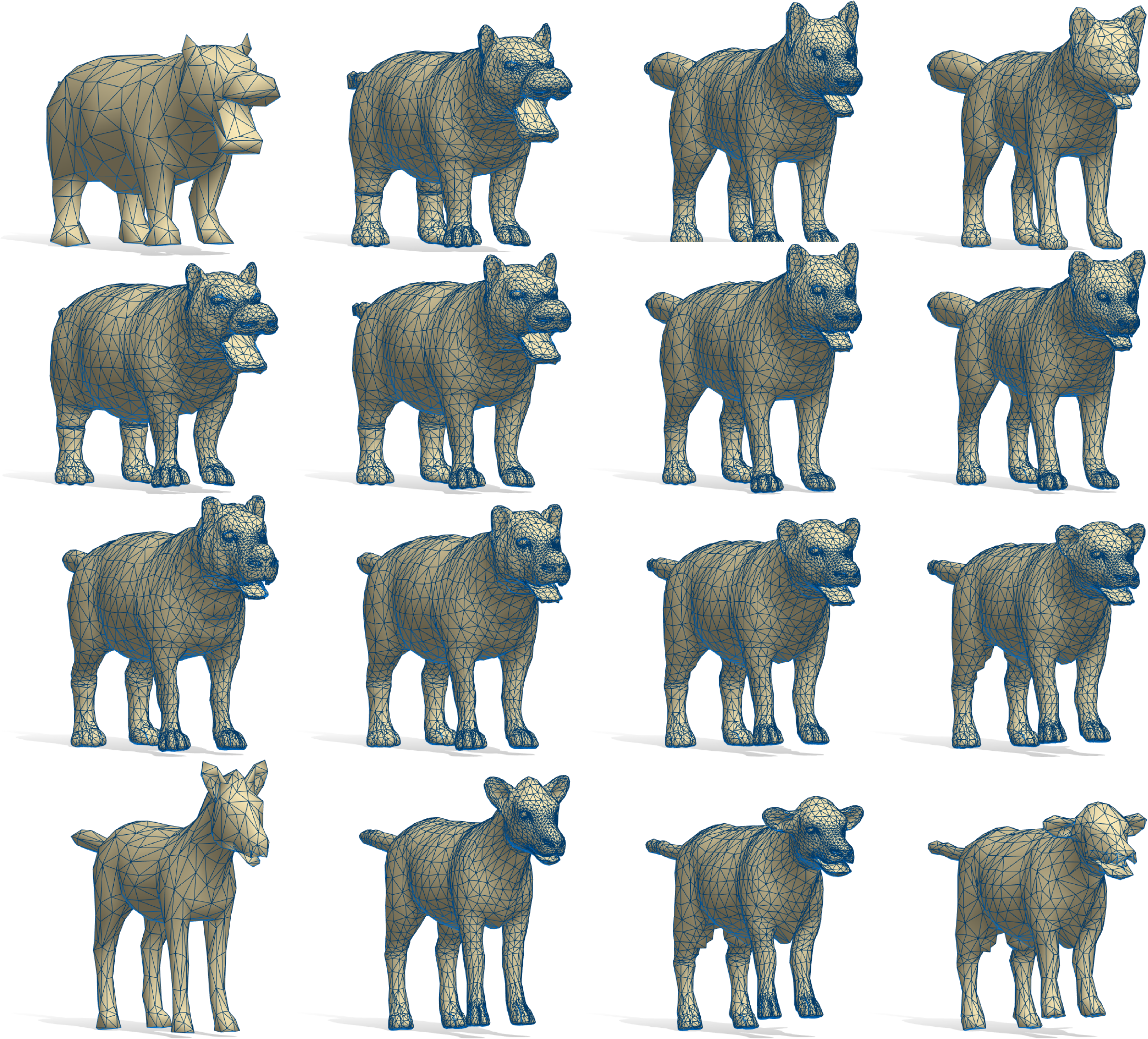}
  \put(35,92.5){\footnotesize \textbf{Interpolation of eigenvalues}}
  \end{overpic}
  \hspace{0.2cm}
  \begin{overpic}
  [trim=0cm 0cm 0cm 0cm,clip,width=0.46\linewidth]{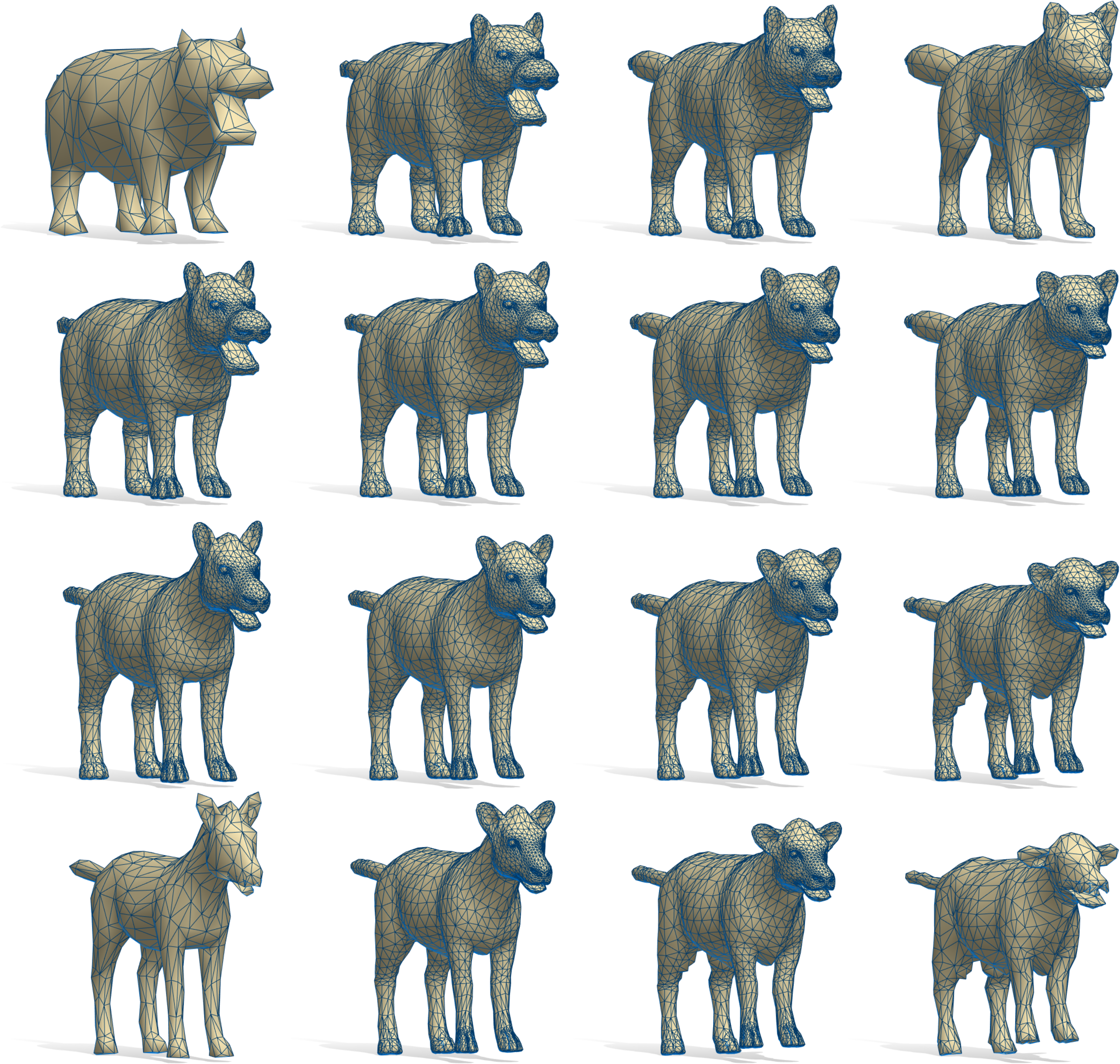}
  \put(-0.5,2){\line(0,1){87}}
  \put(30,98){\footnotesize \textbf{Interpolation of latent vectors}}
  \end{overpic}
  \caption{\label{fig:interp2} Comparison between interpolation in the space of eigenvalues (left) and the interpolation in the latent space (right). The input shapes are four low-resolution versions of as many shapes from the training set.
  }
  \end{center}
\end{figure*}
\section{Super-resolution}
\label{sec:super_res}
Our trained network model is largely insensitive to mesh resolution and sampling as we show in our experiments in the main manuscript.
This property makes our network appropriate for the task of mesh super-resolution. Given a low-resolution mesh as input, we can recover a higher resolution counterpart of the mesh that is also in dense point-to-point correspondence with models from the training set. We can do that in a single shot starting only from the spectrum of the input low-resolution mesh.
In Fig.~\ref{fig:remesh2}, we show a comparison with nearest-neighbors (NN) between eigenvalues (with shapes within the training set), and the \textit{isospectralization} method of Cosmo~\etal~\cite{isosp} for two different shapes from a subject that is not involved in the training of our network model.

In Fig.~\ref{fig:remesh3}, we perform the same experiment on three different shapes never seen during the training, and belonging to a subject whose other poses are in the training set. This comparison highlights that our method generalizes better than NN also for shapes that should be well described by the training set. This allows us to state that navigation driven by the eigenvalues is more accurate than NN navigation of the latent space.

\section{Spectral shape exploration}
\label{sec:interpolation}
In the main manuscript, we show two different sub-tasks of shape exploration: shape interpolation and interactive spectrum-driven exploration.

As we claim in the main manuscript, an alternative for shape interpolation is to regard the eigenvalues themselves as a space that we can explore, as is typically done with latent spaces.
In Fig.~\ref{fig:interp2}, on the same four low-resolution shapes, we compare interpolation in the latent space (on the right) with interpolation in the space of the eigenvalues (on the left). The former evaluation is also included in the main manuscript. For the latter, given the four shapes, we first compute their spectra, directly perform a bilinear interpolation among these spectra, and finally reconstruct the corresponding shapes.
Also in this case, the entire procedure can be performed for shapes with different connectivity. 
The linear interpolation among spectra gives rise to meaningful results. However, observe that the space of spectra (where each point is a different sequence of eigenvalues) is not a proper vector space, and taking linear steps is not completely appropriate in this domain.

\begin{figure*}[h!]
\begin{center}
  \begin{overpic}
  [trim=0cm 0cm 0cm 0cm,clip,width=0.8\linewidth]{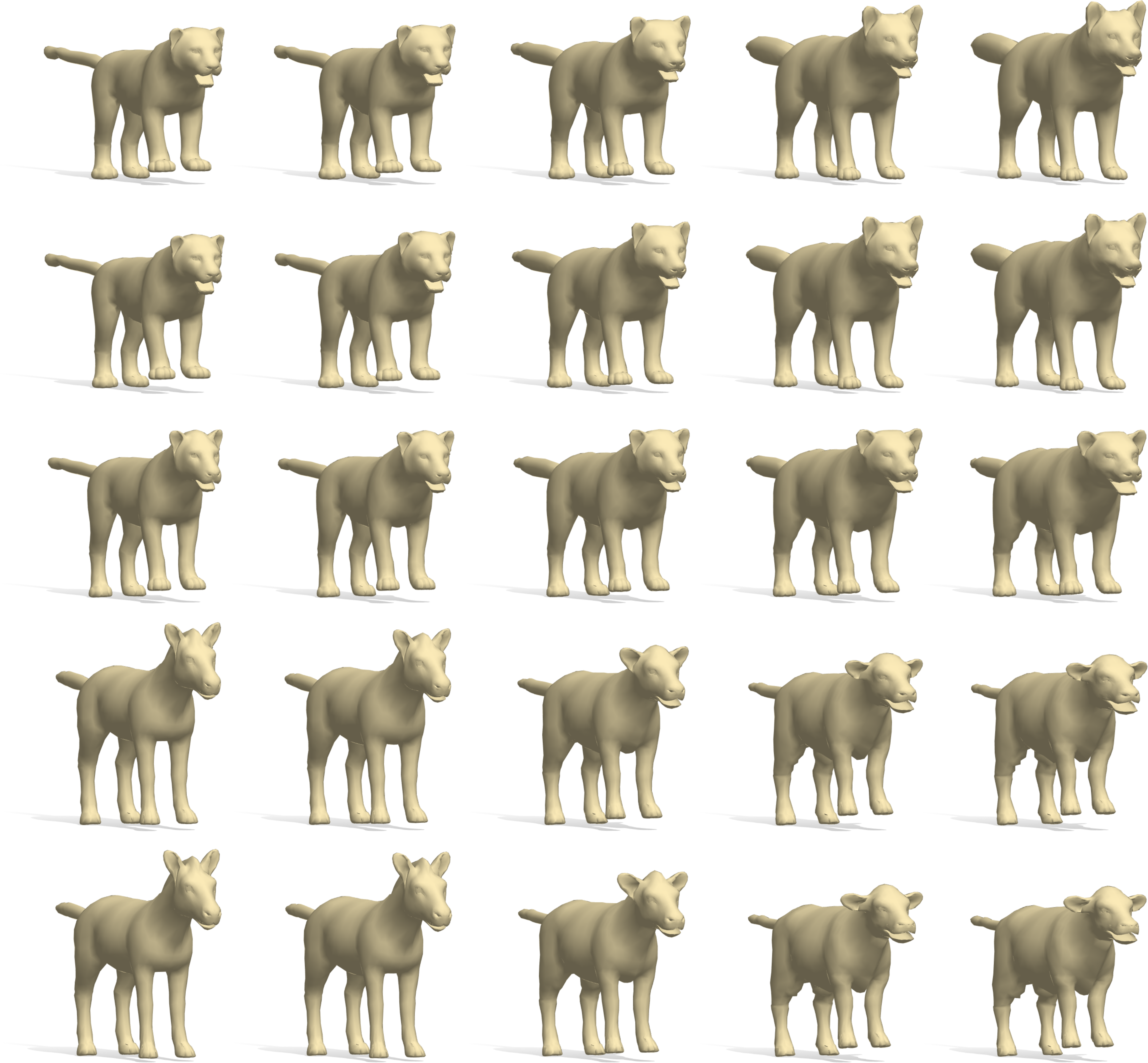}
  \end{overpic}
  \end{center}
  \vspace{0.05cm}
  \caption{Interpolation {\em between spectra} (as opposed to latent vectors) between four shapes in the training set.}
  \label{fig:interp3}
\end{figure*}

\begin{figure*}[h!]
\begin{center}
  \begin{overpic}
  [trim=0cm 0cm 0cm 0cm,clip,width=1\linewidth]{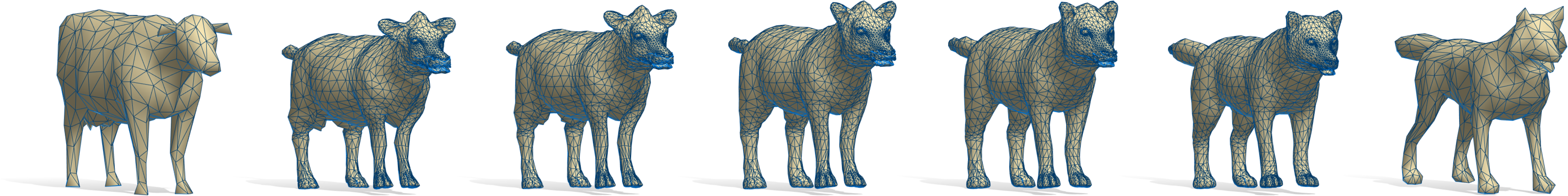}
  \put(8,-2){\footnotesize{Source}}
  \put(94,-2){\footnotesize{Target}}
  \end{overpic}
  \vspace{0.05cm}
  \caption{\label{fig:interp4}Interpolation between the spectra of two low-poly shapes that belong to two different datasets, never seen at training time.}
  \end{center}
\end{figure*}

We support this with two additional figures.
In Fig.~\ref{fig:interp3}, for four shapes in the training set, we provide a denser linear interpolation in the space of the eigenvalues.
In Fig.~\ref{fig:interp4}, we show the same linear interpolation between the spectra of a pair of shapes that come from two different datasets, and which were never seen in the training phase.
Both cases confirm that interpolation can indeed be performed in the space of eigenvalues.  

Finally, in a video attached to these supplementary materials, we show a demo of interactive spectrum-driven shape exploration. In the video, we modify the eigenvalues of a given shape and we show how these modifications allow us to interactively navigate the space of shapes. 
%
%
%

\begin{figure}[!t]
\centering

 \setlength{\tabcolsep}{0pt}
 \begin{tabular}{l r}
 \hspace{-3.85cm}
 

\begin{minipage}{0.69\linewidth}
    \vspace{-0.15cm}

     \begin{overpic}[trim=0cm 0cm 0cm 0cm,clip,width=0.94\linewidth]{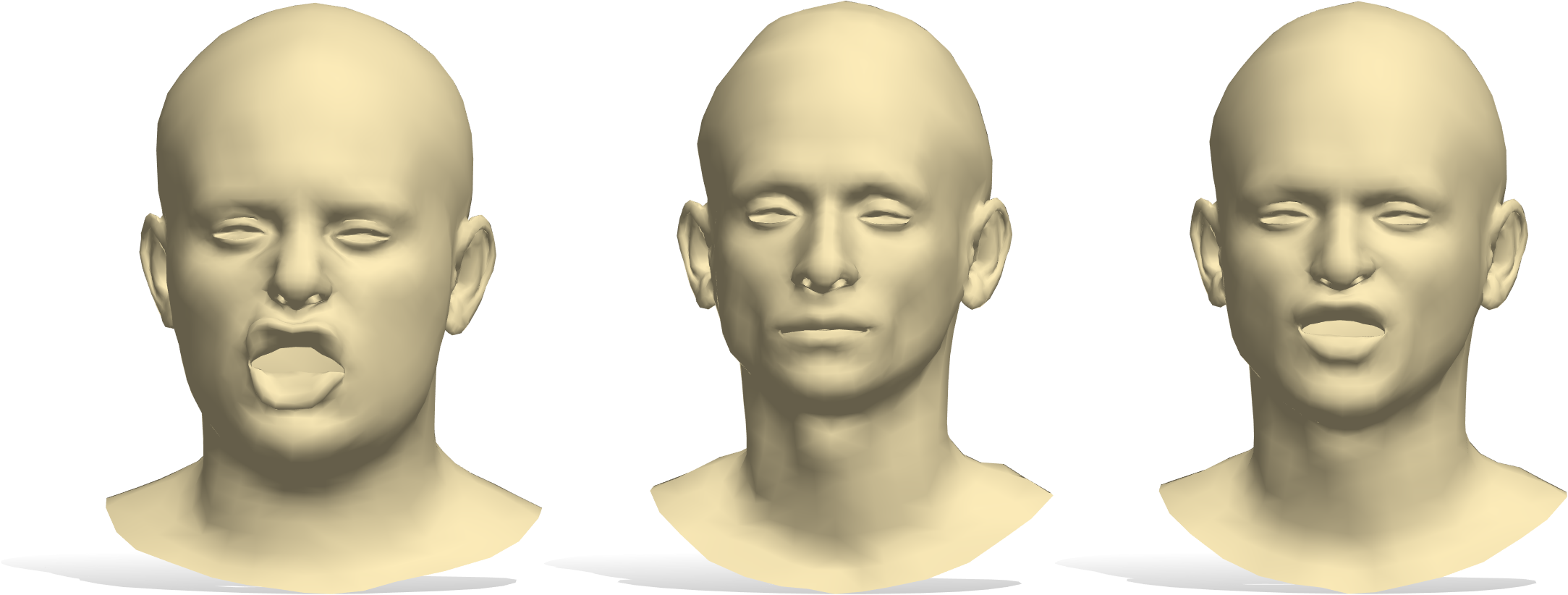}
         \put(8.5,41){\footnotesize \pose{pose target}}
         \put(42.5,41){\footnotesize \style{style target}}
         \put(75.5,41){\footnotesize \textbf{\our{our result}}}
    \end{overpic}
    \vspace{0.57cm}

    \begin{overpic}[trim=0cm 0cm 0cm 0cm,clip,width=0.94\linewidth]{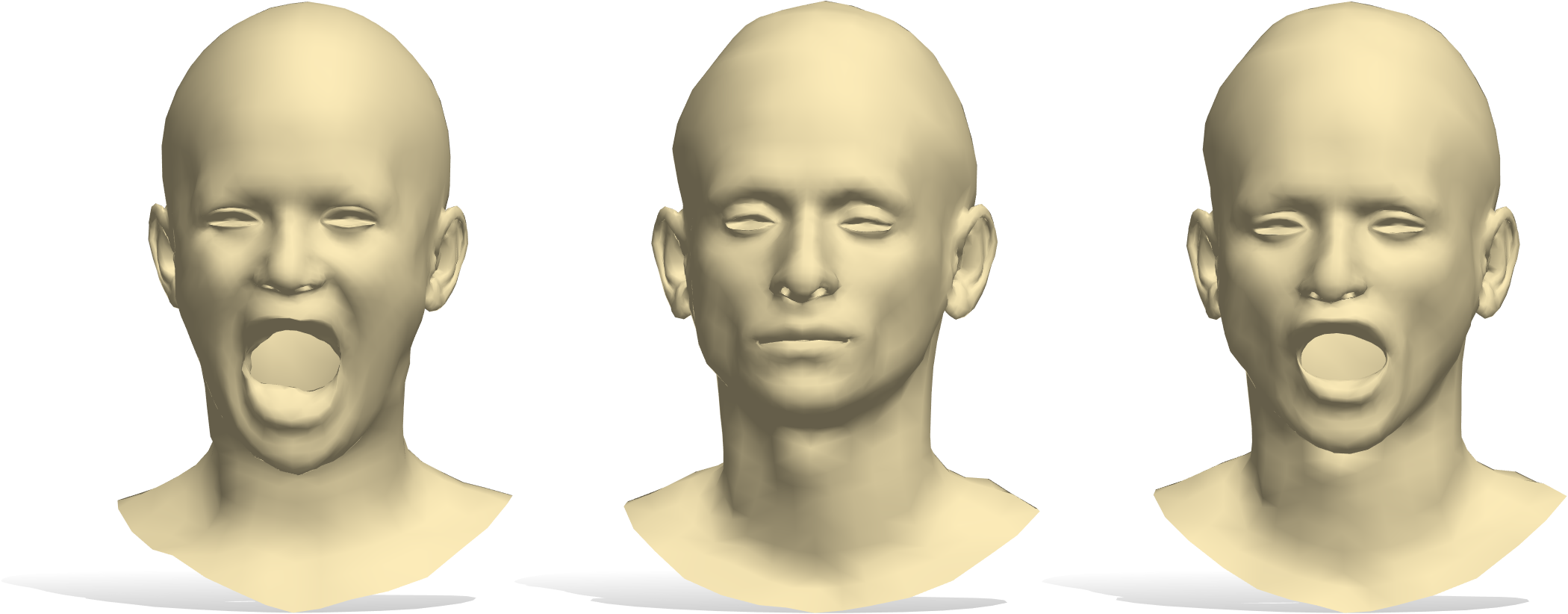}
    \end{overpic}
    \vspace{0.61cm}

    \begin{overpic}[trim=0cm 0cm 0cm 0cm,clip,width=0.94\linewidth]{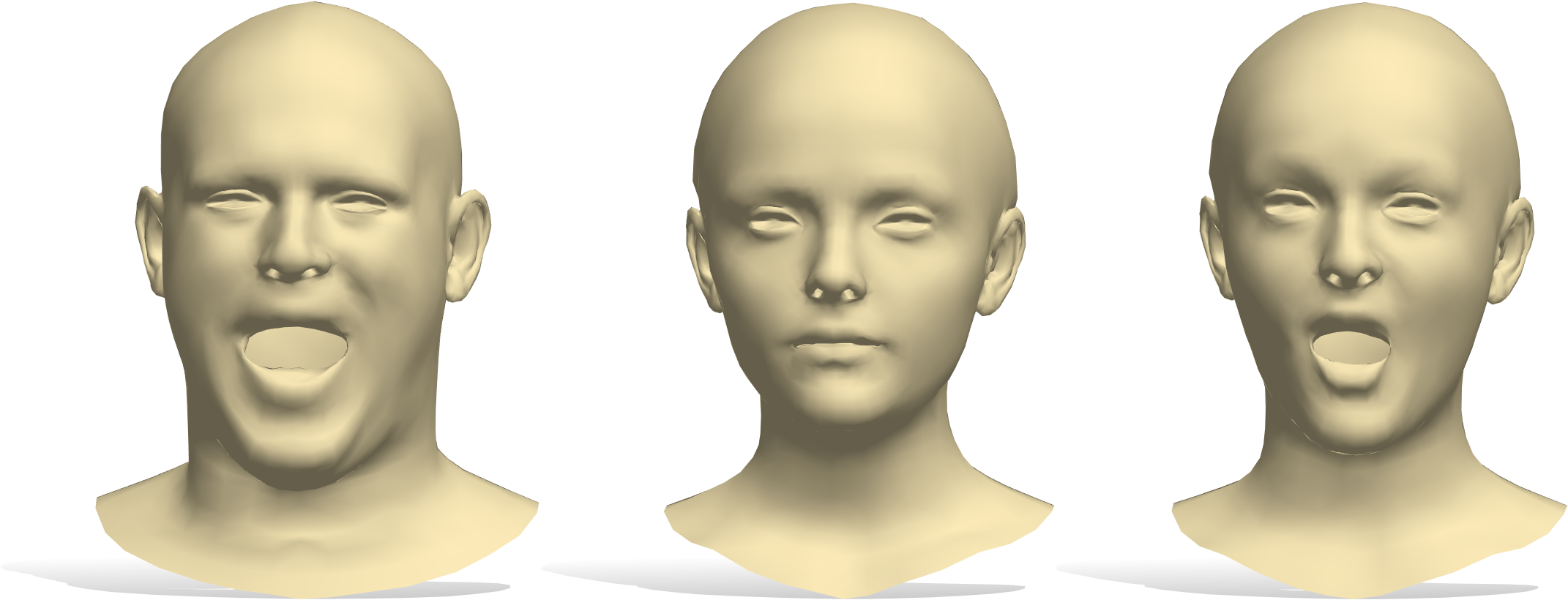}
    \end{overpic}
    \vspace{0.575cm}

    \begin{overpic}[trim=0cm 0cm 0cm 0cm,clip,width=0.93\linewidth]{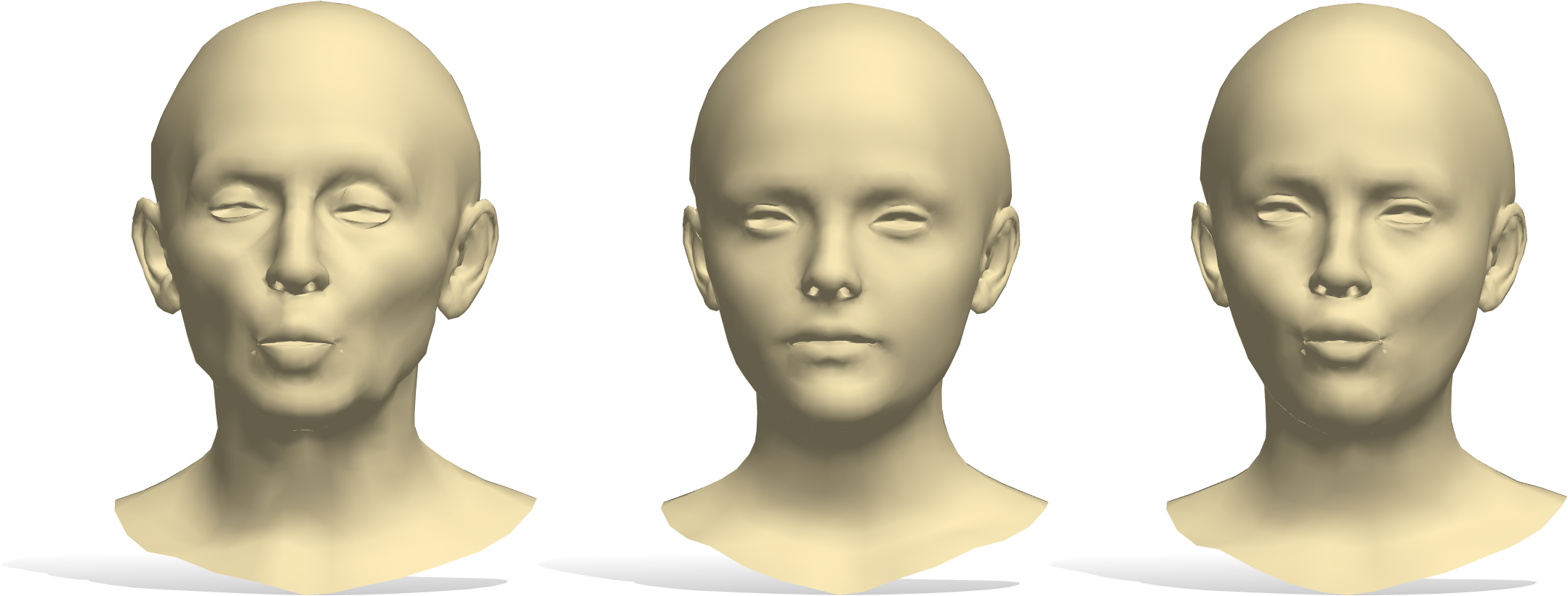}
    \end{overpic}
    \vspace{0.59cm}
    
    \begin{overpic}[trim=0cm 0cm 0cm 0cm,clip,width=0.93\linewidth]{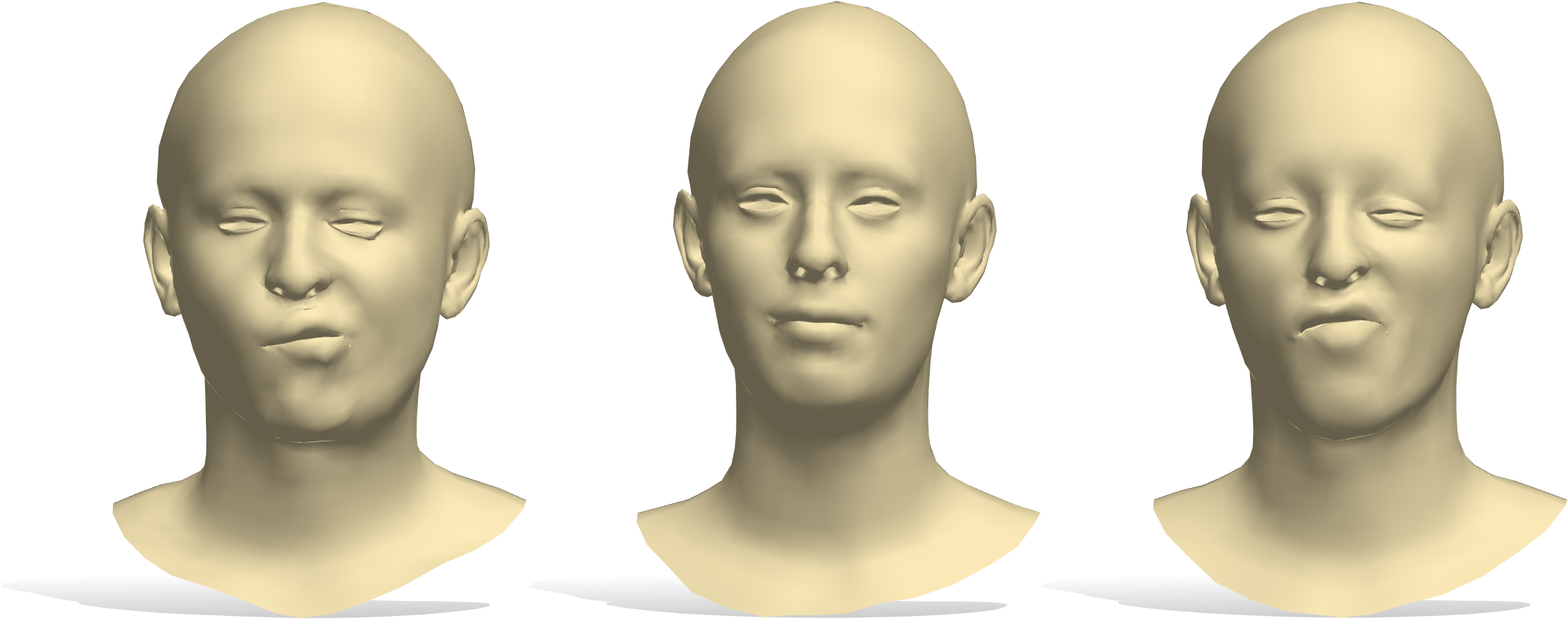}
    \end{overpic}
    \vspace{0.59cm}
    
    \begin{overpic}[trim=0cm 0cm 0cm 0cm,clip,width=0.93\linewidth]{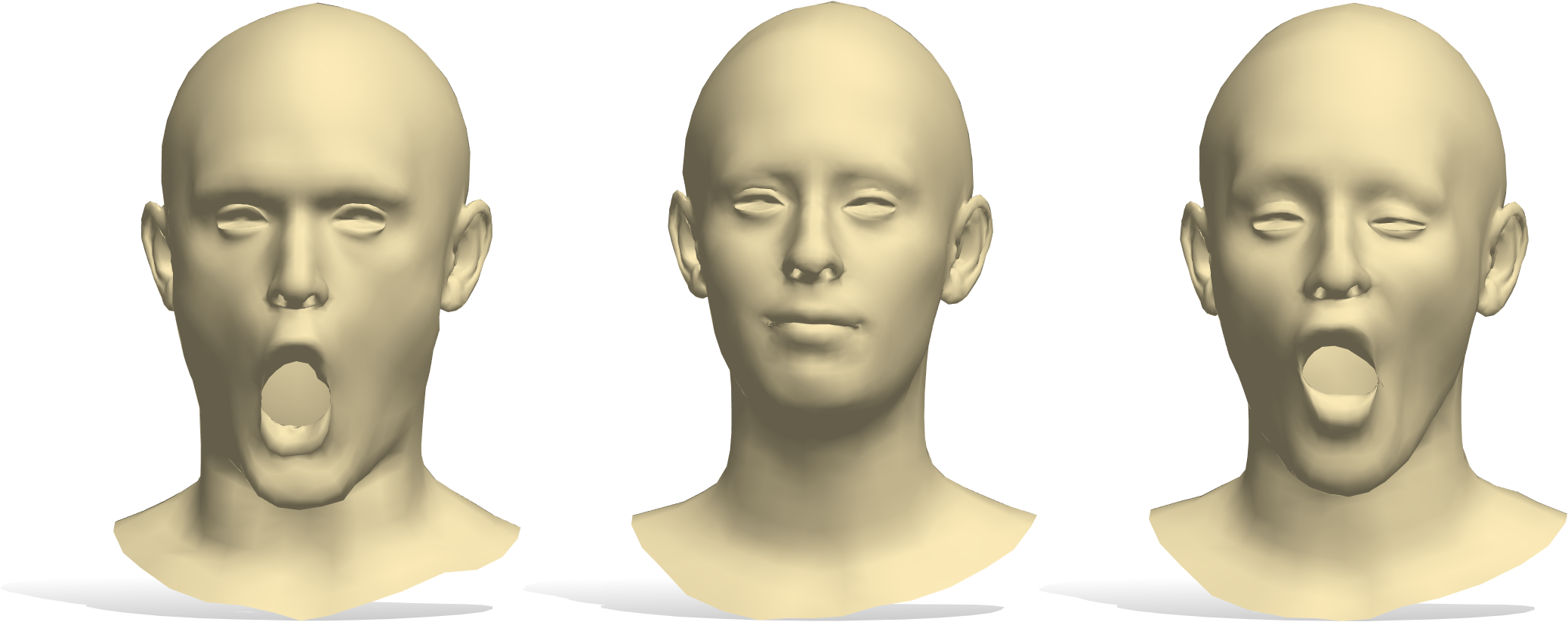}
    \end{overpic}
    \vspace{0.63cm}
    
    
    \begin{overpic}[trim=0cm 0cm 0cm 0cm,clip,width=0.94\linewidth]{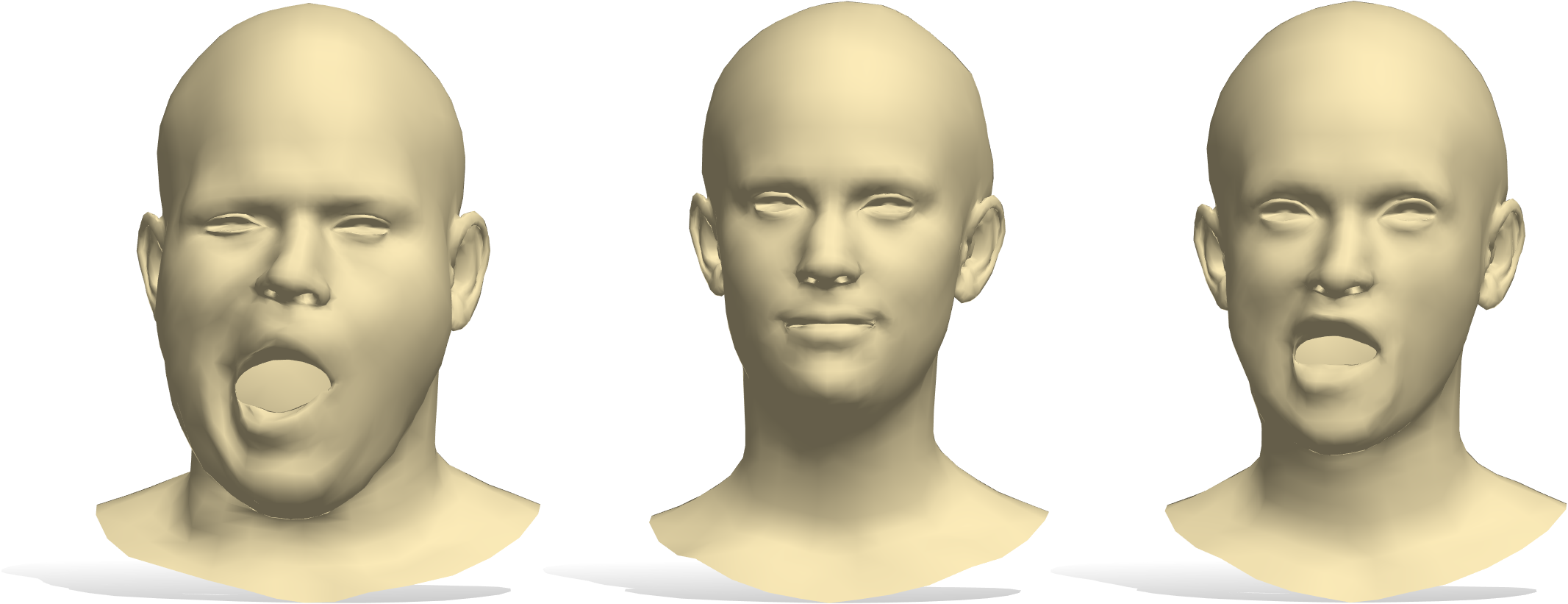}
    \end{overpic}
    \vspace{0.63cm}
    
    
    \begin{overpic}[trim=0cm 0cm 0cm 0cm,clip,width=0.94\linewidth]{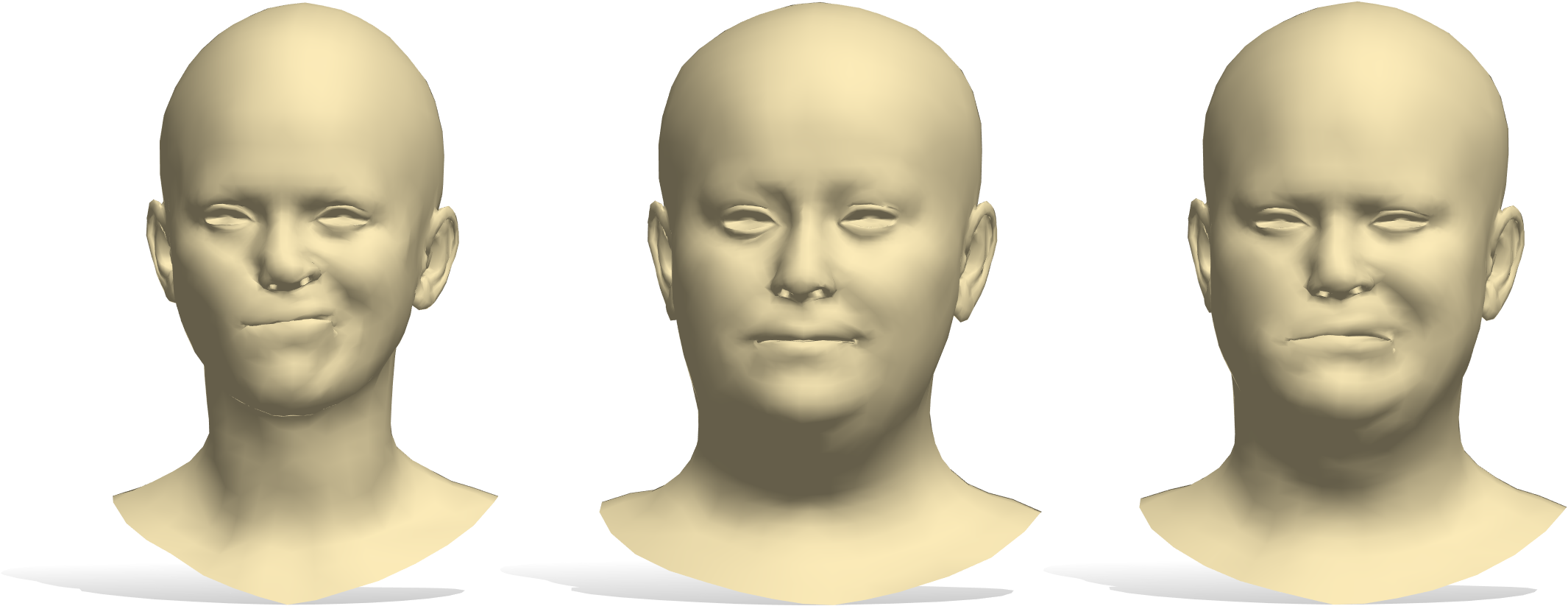}
    \end{overpic}
\end{minipage}%
&
\hspace{-4.25cm}
 
\begin{minipage}{0.22\linewidth}
\vspace{-0.02cm}

%
%
\definecolor{mycolor1}{rgb}{0.85000,0.32500,0.09800}%
\definecolor{mycolor2}{rgb}{0.92900,0.69400,0.12500}%
\definecolor{mycolor3}{rgb}{0.00000,0.44700,0.74100}%
\begin{tikzpicture}

\begin{axis}[%
width=\linewidth,
height=\linewidth,
at={(0.758in,0.481in)},
scale only axis,
xmin=1,
xmax=30,
ymin=0,
ymax=2000,
xtick={10, 20, 30},
xticklabels={ , , },
ytick={500, 1000, 1500, 2000},
yticklabels={,  ,  },
xmajorgrids,
ymajorgrids,
every x tick label/.append style={font=\color{black}, font=\tiny},
every y tick label/.append style={font=\color{black}, font=\tiny},
title={\footnotesize eigenvalues},
title style={yshift=-0.75em},
axis background/.style={fill=white},
legend columns=-1,
legend style={column sep=0.4ex,
at={(-2.5,-0.4)},anchor=south west,legend cell align=left,align=left,draw=white!15!black}
]
\addplot [color=mycolor1,solid,line width=1.5pt]
  table[row sep=crcr]{%
1	81.1384523321054\\
2	157.979118795331\\
3	186.161628940011\\
4	278.249882819152\\
5	285.008945486572\\
6	292.32867847876\\
7	433.518126552016\\
8	556.848632271077\\
9	565.483199357002\\
10	658.795002127404\\
11	667.81927437278\\
12	791.198139712387\\
13	828.222126609436\\
14	887.479659668246\\
15	946.146081665899\\
16	956.169607374905\\
17	1119.44658247935\\
18	1134.7477110466\\
19	1206.25510025473\\
20	1310.49678828915\\
21	1332.68841326791\\
22	1453.06759891881\\
23	1587.32104193396\\
24	1591.38451033849\\
25	1619.99432647082\\
26	1657.88517251683\\
27	1724.85689162826\\
28	1857.0477119277\\
29	1869.93260028013\\
30	1986.80929202785\\
};
\addlegendentry{\footnotesize style};

\addplot [color=mycolor2,solid,line width=1.5pt]
  table[row sep=crcr]{%
1	86.2514875605386\\
2	127.69574875459\\
3	151.605922324227\\
4	251.033244936644\\
5	264.216072166086\\
6	303.046736678832\\
7	390.597309516592\\
8	499.806694065856\\
9	521.253649958562\\
10	585.099116326026\\
11	622.285041444549\\
12	669.213788059632\\
13	720.930061556924\\
14	827.152346100035\\
15	868.216141750707\\
16	907.269065241297\\
17	1046.81462728699\\
18	1066.82987041791\\
19	1077.84436237327\\
20	1154.33520958081\\
21	1234.9170380596\\
22	1294.07993585757\\
23	1336.92378812357\\
24	1427.92712087261\\
25	1492.45432118932\\
26	1566.32831878521\\
27	1618.51216991067\\
28	1652.24388555112\\
29	1669.37417434768\\
30	1810.86289071361\\
};
\addlegendentry{\footnotesize pose 2.90};

\addplot [color=mycolor3,solid,line width=1.5pt]
  table[row sep=crcr]{%
1	82.3723121671407\\
2	149.201531070665\\
3	175.201422884151\\
4	262.921368571841\\
5	281.862506452454\\
6	290.773135105229\\
7	428.195167408999\\
8	544.614956657702\\
9	548.968011081605\\
10	635.143548560589\\
11	650.705668265617\\
12	739.517975931548\\
13	773.347604415687\\
14	873.962689215484\\
15	924.76524780211\\
16	951.095118754927\\
17	1107.08950611045\\
18	1115.75241259517\\
19	1152.15725980842\\
20	1263.94679441858\\
21	1303.96281964315\\
22	1417.5885423715\\
23	1491.07604333139\\
24	1514.5091640763\\
25	1600.5544566553\\
26	1612.5470102842\\
27	1688.49956414204\\
28	1778.96709477562\\
29	1816.67460243361\\
30	1930.5181956729\\
};
\addlegendentry{\footnotesize our 0.92};

\addplot [color=black,dotted,line width=1.5pt]
  table[row sep=crcr]{%
1	81.3501892089844\\
2	152.432281494141\\
3	180.35823059082\\
4	269.895111083984\\
5	281.878448486328\\
6	290.214141845703\\
7	424.569213867188\\
8	547.932556152344\\
9	560.196228027344\\
10	648.510620117188\\
11	660.615905761719\\
12	760.672302246094\\
13	792.485961914063\\
14	881.266174316406\\
15	921.570556640625\\
16	950.28515625\\
17	1108.21765136719\\
18	1121.16247558594\\
19	1186.01354980469\\
20	1288.98449707031\\
21	1322.98486328125\\
22	1422.95812988281\\
23	1529.7880859375\\
24	1547.75048828125\\
25	1595.98742675781\\
26	1638.12585449219\\
27	1699.68041992188\\
28	1804.81115722656\\
29	1836.11584472656\\
30	1958.68579101563\\
};

\end{axis}
\end{tikzpicture}%
%
%
\definecolor{mycolor1}{rgb}{0.85000,0.32500,0.09800}%
\definecolor{mycolor2}{rgb}{0.92900,0.69400,0.12500}%
\definecolor{mycolor3}{rgb}{0.00000,0.44700,0.74100}%
\begin{tikzpicture}

\begin{axis}[%
width=\linewidth,
height=\linewidth,
at={(0.758in,0.481in)},
scale only axis,
xmin=1,
xmax=30,
ymin=0,
ymax=2000,
xtick={10, 20, 30},
xticklabels={ , , },
ytick={500, 1000, 1500, 2000},
yticklabels={,  ,  },
xmajorgrids,
ymajorgrids,
every x tick label/.append style={font=\color{black}, font=\tiny},
every y tick label/.append style={font=\color{black}, font=\tiny},
axis background/.style={fill=white},
legend columns=-1,
legend style={column sep=0.4ex,
at={(-2.5,-0.4)},anchor=south west,legend cell align=left,align=left,draw=white!15!black}
]
\addplot [color=mycolor1,solid,line width=1.5pt]
  table[row sep=crcr]{%
1	81.1384523321054\\
2	157.979118795331\\
3	186.161628940011\\
4	278.249882819152\\
5	285.008945486572\\
6	292.32867847876\\
7	433.518126552016\\
8	556.848632271077\\
9	565.483199357002\\
10	658.795002127404\\
11	667.81927437278\\
12	791.198139712387\\
13	828.222126609436\\
14	887.479659668246\\
15	946.146081665899\\
16	956.169607374905\\
17	1119.44658247935\\
18	1134.7477110466\\
19	1206.25510025473\\
20	1310.49678828915\\
21	1332.68841326791\\
22	1453.06759891881\\
23	1587.32104193396\\
24	1591.38451033849\\
25	1619.99432647082\\
26	1657.88517251683\\
27	1724.85689162826\\
28	1857.0477119277\\
29	1869.93260028013\\
30	1986.80929202785\\
};
\addlegendentry{\footnotesize style};

\addplot [color=mycolor2,solid,line width=1.5pt]
  table[row sep=crcr]{%
1	70.482430779491\\
2	152.788732756713\\
3	179.957684252256\\
4	232.282014911946\\
5	273.178164908867\\
6	286.005851503333\\
7	447.003422372159\\
8	502.74596742307\\
9	525.20102229177\\
10	597.630620840037\\
11	614.97783086971\\
12	730.816924990524\\
13	744.767639937386\\
14	795.284621082092\\
15	912.400396765394\\
16	961.799819181239\\
17	1031.67638995556\\
18	1051.97814231313\\
19	1143.89199948559\\
20	1195.75639403036\\
21	1291.44619318207\\
22	1395.43217920892\\
23	1463.03753350548\\
24	1489.78436744023\\
25	1549.94515411838\\
26	1590.74788041432\\
27	1601.88179533275\\
28	1693.19604643235\\
29	1708.28367833028\\
30	1874.29135339683\\
};
\addlegendentry{\footnotesize pose 2.00};

\addplot [color=mycolor3,solid,line width=1.5pt]
  table[row sep=crcr]{%
1	76.4396677774136\\
2	154.194121566951\\
3	181.536978022345\\
4	250.259576002539\\
5	277.709642305002\\
6	284.817674766432\\
7	436.305600635191\\
8	530.683673885022\\
9	538.395313513996\\
10	633.054844273981\\
11	648.321203495865\\
12	735.080888054502\\
13	756.684003688229\\
14	821.786388893005\\
15	939.5490057502\\
16	957.781325552503\\
17	1082.57924233475\\
18	1103.62117641989\\
19	1180.22590798469\\
20	1281.54553067397\\
21	1336.59567364121\\
22	1393.13570254809\\
23	1466.28260483123\\
24	1479.7670577497\\
25	1572.35265841656\\
26	1599.39856834949\\
27	1681.4721443907\\
28	1771.19913813959\\
29	1782.86608171397\\
30	1916.21168083228\\
};
\addlegendentry{\footnotesize  our 1.20};

\addplot [color=black,dotted,line width=1.5pt]
  table[row sep=crcr]{%
1	78.7154846191406\\
2	154.642333984375\\
3	183.709777832031\\
4	263.367095947266\\
5	285.067565917969\\
6	288.89111328125\\
7	431.490325927734\\
8	538.289367675781\\
9	558.387390136719\\
10	646.719116210938\\
11	654.961608886719\\
12	763.399719238281\\
13	790.051330566406\\
14	864.82080078125\\
15	933.470336914063\\
16	961.916625976563\\
17	1106.1044921875\\
18	1117.92163085938\\
19	1188.54614257813\\
20	1290.35217285156\\
21	1328.17578125\\
22	1420.71301269531\\
23	1530.69689941406\\
24	1546.24060058594\\
25	1591.21813964844\\
26	1641.4296875\\
27	1698.5263671875\\
28	1803.97412109375\\
29	1830.55187988281\\
30	1967.26611328125\\
};

\end{axis}
\end{tikzpicture}%
%
%
\definecolor{mycolor1}{rgb}{0.85000,0.32500,0.09800}%
\definecolor{mycolor2}{rgb}{0.92900,0.69400,0.12500}%
\definecolor{mycolor3}{rgb}{0.00000,0.44700,0.74100}%
\begin{tikzpicture}

\begin{axis}[%
width=\linewidth,
height=\linewidth,
at={(0.758in,0.481in)},
scale only axis,
xmin=1,
xmax=30,
ymin=0,
ymax=2000,
xtick={10, 20, 30},
xticklabels={ , , },
ytick={500, 1000, 1500, 2000},
yticklabels={,  ,  },
xmajorgrids,
ymajorgrids,
every x tick label/.append style={font=\color{black}, font=\tiny},
every y tick label/.append style={font=\color{black}, font=\tiny},
axis background/.style={fill=white},
legend columns=-1,
legend style={column sep=0.4ex,
at={(-2.5,-0.4)},anchor=south west,legend cell align=left,align=left,draw=white!15!black}
]
\addplot [color=mycolor1,solid,line width=1.5pt]
  table[row sep=crcr]{%
1	78.2510565576848\\
2	146.508297086675\\
3	177.574355547056\\
4	254.605198591946\\
5	291.971053417035\\
6	310.707156199576\\
7	419.041403762707\\
8	501.614595639535\\
9	525.960177223618\\
10	581.008829552778\\
11	617.693455038977\\
12	821.686753012428\\
13	860.688121360393\\
14	884.855558038208\\
15	911.600859443231\\
16	932.750323497078\\
17	1043.02055705434\\
18	1054.99276797046\\
19	1101.99797125091\\
20	1147.2241484082\\
21	1276.05085942945\\
22	1344.28462425999\\
23	1529.09507610706\\
24	1544.62082411615\\
25	1598.83377386618\\
26	1677.10168849299\\
27	1719.79990521025\\
28	1753.90469893573\\
29	1762.35641250669\\
30	1843.30176959409\\
};
\addlegendentry{\footnotesize style};

\addplot [color=mycolor2,solid,line width=1.5pt]
  table[row sep=crcr]{%
1	82.6492687544362\\
2	133.579013349477\\
3	151.604342859165\\
4	233.796412254886\\
5	259.660362044112\\
6	297.953549406802\\
7	428.776312998129\\
8	495.719910294317\\
9	517.145781903565\\
10	585.624597503925\\
11	589.688070709374\\
12	654.376283706083\\
13	686.470296790777\\
14	806.946106465201\\
15	857.140910215396\\
16	964.371384211749\\
17	1030.61962876152\\
18	1048.25451512194\\
19	1112.29329572646\\
20	1123.04609724129\\
21	1224.47034842448\\
22	1293.18633760695\\
23	1314.44618927495\\
24	1428.51156762868\\
25	1489.84847032913\\
26	1499.27458977992\\
27	1646.5062240243\\
28	1708.75539150798\\
29	1718.13417372683\\
30	1767.95453182715\\
};
\addlegendentry{\footnotesize  pose 1.90};

\addplot [color=mycolor3,solid,line width=1.5pt]
  table[row sep=crcr]{%
1	75.6835361195083\\
2	146.435647953604\\
3	174.362654195266\\
4	250.166257834861\\
5	281.540404907306\\
6	291.214588497542\\
7	422.438115301877\\
8	491.078587081872\\
9	523.551203872499\\
10	580.644127181931\\
11	590.773609986515\\
12	781.70148229414\\
13	811.182893998293\\
14	836.787678289235\\
15	901.692251310843\\
16	931.621037533406\\
17	1033.66186750848\\
18	1051.93003494899\\
19	1071.16080909272\\
20	1140.59268687911\\
21	1272.0620392532\\
22	1339.51256201143\\
23	1474.63555045946\\
24	1517.98214069382\\
25	1587.63635278914\\
26	1623.47503798667\\
27	1647.01614883213\\
28	1672.18187653708\\
29	1721.92945595116\\
30	1843.65334767731\\
};
\addlegendentry{\footnotesize our 0.67};

\addplot [color=black,dotted,line width=1.5pt]
  table[row sep=crcr]{%
1	76.6855239868164\\
2	143.88151550293\\
3	174.535171508789\\
4	248.640029907227\\
5	282.037048339844\\
6	301.429260253906\\
7	417.534332275391\\
8	490.220306396484\\
9	518.923767089844\\
10	576.982604980469\\
11	600.044860839844\\
12	790.132751464844\\
13	820.484497070313\\
14	853.761169433594\\
15	903.197387695313\\
16	918.786804199219\\
17	1028.64855957031\\
18	1044.86877441406\\
19	1088.79699707031\\
20	1135.43908691406\\
21	1263.701171875\\
22	1333.60302734375\\
23	1489.171875\\
24	1515.32153320313\\
25	1572.2578125\\
26	1633.49206542969\\
27	1670.19189453125\\
28	1694.86413574219\\
29	1729.57043457031\\
30	1811.4228515625\\
};

\end{axis}
\end{tikzpicture}%
%
%
\definecolor{mycolor1}{rgb}{0.85000,0.32500,0.09800}%
\definecolor{mycolor2}{rgb}{0.92900,0.69400,0.12500}%
\definecolor{mycolor3}{rgb}{0.00000,0.44700,0.74100}%
\begin{tikzpicture}

\begin{axis}[%
width=\linewidth,
height=\linewidth,
at={(0.758in,0.481in)},
scale only axis,
xmin=1,
xmax=30,
ymin=0,
ymax=2500,
xtick={10, 20, 30},
xticklabels={, , },
ytick={500, 1000, 1500, 2000},
yticklabels={,  ,  },
xmajorgrids,
ymajorgrids,
every x tick label/.append style={font=\color{black}, font=\tiny},
every y tick label/.append style={font=\color{black}, font=\tiny},
axis background/.style={fill=white},
legend columns=-1,
legend style={column sep=0.4ex,
at={(-2.5,-0.4)},anchor=south west,legend cell align=left,align=left,draw=white!15!black}
]
\addplot [color=mycolor1,solid,line width=1.5pt]
  table[row sep=crcr]{%
1	78.2510565576848\\
2	146.508297086675\\
3	177.574355547056\\
4	254.605198591946\\
5	291.971053417035\\
6	310.707156199576\\
7	419.041403762707\\
8	501.614595639535\\
9	525.960177223618\\
10	581.008829552778\\
11	617.693455038977\\
12	821.686753012428\\
13	860.688121360393\\
14	884.855558038208\\
15	911.600859443231\\
16	932.750323497078\\
17	1043.02055705434\\
18	1054.99276797046\\
19	1101.99797125091\\
20	1147.2241484082\\
21	1276.05085942945\\
22	1344.28462425999\\
23	1529.09507610706\\
24	1544.62082411615\\
25	1598.83377386618\\
26	1677.10168849299\\
27	1719.79990521025\\
28	1753.90469893573\\
29	1762.35641250669\\
30	1843.30176959409\\
};
\addlegendentry{\footnotesize style};

\addplot [color=mycolor2,solid,line width=1.5pt]
  table[row sep=crcr]{%
1	89.389439077576\\
2	158.918672675349\\
3	184.499398580706\\
4	276.627143210527\\
5	302.588686437827\\
6	330.00261111266\\
7	460.523036678392\\
8	585.474890212198\\
9	594.054120741261\\
10	657.581905304933\\
11	726.183632497664\\
12	800.011248275266\\
13	829.679227648984\\
14	986.607902686637\\
15	997.508038711131\\
16	1030.43954275823\\
17	1192.61939620007\\
18	1201.67949022786\\
19	1229.20175789964\\
20	1318.83863078016\\
21	1451.92125639281\\
22	1539.47073441341\\
23	1631.76329056178\\
24	1642.94930822369\\
25	1737.04794715391\\
26	1778.54529235023\\
27	1821.18139661284\\
28	1907.80718927081\\
29	1934.37384572106\\
30	2066.34081958045\\
};
\addlegendentry{\footnotesize pose 3.00};

\addplot [color=mycolor3,solid,line width=1.5pt]
  table[row sep=crcr]{%
1	78.2891791089268\\
2	145.954195019706\\
3	177.564330298362\\
4	252.314250922214\\
5	290.406799547486\\
6	295.056581385928\\
7	421.45472936923\\
8	513.305489257316\\
9	534.333885733169\\
10	589.454568715474\\
11	625.886920489215\\
12	812.795428447651\\
13	827.868127220178\\
14	866.703283986106\\
15	911.003803643469\\
16	923.561294777646\\
17	1058.09414684116\\
18	1073.04949168531\\
19	1096.02373182038\\
20	1160.41160650008\\
21	1301.67293111484\\
22	1373.88140644066\\
23	1544.76580708196\\
24	1566.26446822756\\
25	1606.5570634204\\
26	1631.45355593278\\
27	1655.73958870233\\
28	1719.57302878128\\
29	1729.88591869972\\
30	1877.68829700352\\
};
\addlegendentry{\footnotesize our 0.46};

\addplot [color=black,dotted,line width=1.5pt]
  table[row sep=crcr]{%
1	76.9518737792969\\
2	143.218032836914\\
3	175.083724975586\\
4	251.862533569336\\
5	281.147827148438\\
6	300.089874267578\\
7	407.627075195313\\
8	497.740814208984\\
9	520.383605957031\\
10	574.858947753906\\
11	611.555480957031\\
12	800.216003417969\\
13	828.227172851563\\
14	865.310791015625\\
15	892.857116699219\\
16	905.732055664063\\
17	1030.6884765625\\
18	1045.01525878906\\
19	1086.94714355469\\
20	1137.568359375\\
21	1259.18383789063\\
22	1330.19592285156\\
23	1498.53271484375\\
24	1519.22778320313\\
25	1568.94274902344\\
26	1634.00415039063\\
27	1658.89978027344\\
28	1695.99255371094\\
29	1730.876953125\\
30	1820.04467773438\\
};

\end{axis}
\end{tikzpicture}%
%
%
\definecolor{mycolor1}{rgb}{0.85000,0.32500,0.09800}%
\definecolor{mycolor2}{rgb}{0.92900,0.69400,0.12500}%
\definecolor{mycolor3}{rgb}{0.00000,0.44700,0.74100}%
\begin{tikzpicture}

\begin{axis}[%
width=\linewidth,
height=\linewidth,
at={(0.758in,0.481in)},
scale only axis,
xmin=1,
xmax=30,
ymin=0,
ymax=2500,
xtick={10, 20, 30},
xticklabels={ , , },
ytick={625, 1250, 1875, 2500},
yticklabels={,  ,  },
xmajorgrids,
ymajorgrids,
every x tick label/.append style={font=\color{black}, font=\tiny},
every y tick label/.append style={font=\color{black}, font=\tiny},
axis background/.style={fill=white},
legend columns=-1,
legend style={column sep=0.4ex,
at={(-2.5,-0.4)},anchor=south west,legend cell align=left,align=left,draw=white!15!black}
]
\addplot [color=mycolor1,solid,line width=1.5pt]
  table[row sep=crcr]{%
1	80.7694808616918\\
2	172.069872947381\\
3	210.018024323256\\
4	291.060749413091\\
5	303.998223998724\\
6	320.669901922451\\
7	475.723202452604\\
8	584.245875285047\\
9	621.224892472125\\
10	686.810889287323\\
11	725.065594224612\\
12	865.548976252085\\
13	907.352542041044\\
14	942.605829287358\\
15	996.253016690333\\
16	1031.0771568435\\
17	1218.11086824929\\
18	1239.30952843981\\
19	1311.76804322346\\
20	1347.37238359595\\
21	1428.01333279007\\
22	1576.58961612362\\
23	1685.54304506291\\
24	1748.70977203439\\
25	1764.84312841011\\
26	1801.42170105587\\
27	1834.4974542213\\
28	1956.75027802414\\
29	2051.19330187336\\
30	2077.57095973741\\
};
\addlegendentry{\footnotesize style};

\addplot [color=mycolor2,solid,line width=1.5pt]
  table[row sep=crcr]{%
1	80.5611965182435\\
2	139.109229116719\\
3	168.712691414525\\
4	255.539400281394\\
5	270.921098434852\\
6	286.640851129247\\
7	408.17749389898\\
8	503.029094455344\\
9	518.829728050383\\
10	585.981046445677\\
11	615.252172323623\\
12	764.618913506262\\
13	782.6476173407\\
14	835.665587470946\\
15	877.267292452829\\
16	900.497039522496\\
17	1045.05681827139\\
18	1057.29189545173\\
19	1064.95738685104\\
20	1147.47445276789\\
21	1261.91110193311\\
22	1341.41072217828\\
23	1481.63783321246\\
24	1504.01860464546\\
25	1538.3974202324\\
26	1562.86487132769\\
27	1593.42424046843\\
28	1659.24223261945\\
29	1721.37858540813\\
30	1821.94681352959\\
};
\addlegendentry{\footnotesize pose 4.10};

\addplot [color=mycolor3,solid,line width=1.5pt]
  table[row sep=crcr]{%
1	82.3458426819957\\
2	159.035284748623\\
3	190.948148209418\\
4	273.752998767997\\
5	295.289616916698\\
6	301.78718553808\\
7	450.095478308296\\
8	564.457704611687\\
9	581.3790513435\\
10	668.570958603857\\
11	671.593252908882\\
12	800.427570330354\\
13	831.892806894318\\
14	911.801814717908\\
15	962.886255606648\\
16	993.593119585814\\
17	1162.1134614642\\
18	1171.60063729187\\
19	1212.79996485679\\
20	1317.64628766842\\
21	1361.76251870472\\
22	1490.37233335239\\
23	1608.03736811176\\
24	1612.67785291151\\
25	1641.96550036366\\
26	1700.58921483981\\
27	1758.10178601115\\
28	1870.15056458144\\
29	1896.55260853235\\
30	2039.67454093184\\
};
\addlegendentry{\footnotesize  our 1.60};

\addplot [color=black,dotted,line width=1.5pt]
  table[row sep=crcr]{%
1	83.3001480102539\\
2	165.761322021484\\
3	200.359878540039\\
4	287.475067138672\\
5	300.166900634766\\
6	315.249633789063\\
7	461.858673095703\\
8	579.622131347656\\
9	606.046447753906\\
10	683.193481445313\\
11	709.520263671875\\
12	840.909118652344\\
13	872.936767578125\\
14	929.796020507813\\
15	982.044372558594\\
16	1021.56414794922\\
17	1192.66552734375\\
18	1216.69409179688\\
19	1278.77087402344\\
20	1345.34851074219\\
21	1411.08825683594\\
22	1547.12646484375\\
23	1652.20971679688\\
24	1691.12121582031\\
25	1718.21203613281\\
26	1760.86193847656\\
27	1817.67053222656\\
28	1920.39672851563\\
29	1992.25378417969\\
30	2062.51049804688\\
};

\end{axis}
\end{tikzpicture}%
%
%
\definecolor{mycolor1}{rgb}{0.85000,0.32500,0.09800}%
\definecolor{mycolor2}{rgb}{0.92900,0.69400,0.12500}%
\definecolor{mycolor3}{rgb}{0.00000,0.44700,0.74100}%
\begin{tikzpicture}

\begin{axis}[%
width=\linewidth,
height=\linewidth,
at={(0.758in,0.481in)},
scale only axis,
xmin=1,
xmax=30,
ymin=0,
ymax=2500,
xtick={10, 20, 30},
xticklabels={ , , },
ytick={625, 1250, 1875, 2500},
yticklabels={,  ,  },
xmajorgrids,
ymajorgrids,
every x tick label/.append style={font=\color{black}, font=\tiny},
every y tick label/.append style={font=\color{black}, font=\tiny},
axis background/.style={fill=white},
legend columns=-1,
legend style={column sep=0.4ex,
at={(-2.5,-0.4)},anchor=south west,legend cell align=left,align=left,draw=white!15!black}
]
\addplot [color=mycolor1,solid,line width=1.5pt]
  table[row sep=crcr]{%
1	80.7694808616918\\
2	172.069872947381\\
3	210.018024323256\\
4	291.060749413091\\
5	303.998223998724\\
6	320.669901922451\\
7	475.723202452604\\
8	584.245875285047\\
9	621.224892472125\\
10	686.810889287323\\
11	725.065594224612\\
12	865.548976252085\\
13	907.352542041044\\
14	942.605829287358\\
15	996.253016690333\\
16	1031.0771568435\\
17	1218.11086824929\\
18	1239.30952843981\\
19	1311.76804322346\\
20	1347.37238359595\\
21	1428.01333279007\\
22	1576.58961612362\\
23	1685.54304506291\\
24	1748.70977203439\\
25	1764.84312841011\\
26	1801.42170105587\\
27	1834.4974542213\\
28	1956.75027802414\\
29	2051.19330187336\\
30	2077.57095973741\\
};
\addlegendentry{\footnotesize style};

\addplot [color=mycolor2,solid,line width=1.5pt]
  table[row sep=crcr]{%
1	77.4839161148015\\
2	144.20568436831\\
3	168.520022475561\\
4	241.44716724712\\
5	276.932704134468\\
6	287.292526735284\\
7	423.934622159717\\
8	518.865627937657\\
9	526.549657414184\\
10	609.081904369055\\
11	610.553316672326\\
12	722.404163074678\\
13	740.200358417489\\
14	807.42660907339\\
15	874.633527841727\\
16	932.717382892465\\
17	1051.72891457651\\
18	1062.2395180876\\
19	1103.95605140828\\
20	1171.20496295037\\
21	1280.68636061779\\
22	1335.76025429927\\
23	1452.93861988105\\
24	1482.34942974924\\
25	1525.48233850962\\
26	1562.43416103452\\
27	1636.42575739452\\
28	1687.96157414305\\
29	1727.14314026344\\
30	1850.27456899982\\
};
\addlegendentry{\footnotesize pose 4.00};

\addplot [color=mycolor3,solid,line width=1.5pt]
  table[row sep=crcr]{%
1	84.7570528449817\\
2	163.18314940664\\
3	191.246853398106\\
4	279.5610969166\\
5	308.523260294179\\
6	310.187828695825\\
7	462.51103037182\\
8	567.20396839585\\
9	581.769057421432\\
10	671.340417560553\\
11	684.328476255972\\
12	824.11688031427\\
13	847.112658208575\\
14	906.438080720506\\
15	958.821869566918\\
16	1044.80030169447\\
17	1177.10268282737\\
18	1197.19128617468\\
19	1239.68772258841\\
20	1324.89028718489\\
21	1431.32063945973\\
22	1497.03746878325\\
23	1642.40593702077\\
24	1651.86790520349\\
25	1678.62440684598\\
26	1721.76618657855\\
27	1800.15887646243\\
28	1910.82322540101\\
29	1934.05782309407\\
30	2078.01635680164\\
};
\addlegendentry{\footnotesize our 1.10};

\addplot [color=black,dotted,line width=1.5pt]
  table[row sep=crcr]{%
1	83.3538970947266\\
2	165.394790649414\\
3	198.199996948242\\
4	283.546234130859\\
5	303.096801757813\\
6	318.056091308594\\
7	463.764709472656\\
8	571.64794921875\\
9	604.762878417969\\
10	674.962036132813\\
11	702.307250976563\\
12	840.249938964844\\
13	868.213806152344\\
14	922.391357421875\\
15	974.482971191406\\
16	1036.15625\\
17	1192.75720214844\\
18	1217.15087890625\\
19	1270.21459960938\\
20	1339.50109863281\\
21	1416.13989257813\\
22	1538.13439941406\\
23	1655.17321777344\\
24	1693.48657226563\\
25	1722.14807128906\\
26	1769.4775390625\\
27	1813.16577148438\\
28	1918.10705566406\\
29	1989.18615722656\\
30	2077.8251953125\\
};

\end{axis}
\end{tikzpicture}%
%
%
\definecolor{mycolor1}{rgb}{0.85000,0.32500,0.09800}%
\definecolor{mycolor2}{rgb}{0.92900,0.69400,0.12500}%
\definecolor{mycolor3}{rgb}{0.00000,0.44700,0.74100}%
\begin{tikzpicture}

\begin{axis}[%
width=\linewidth,
height=\linewidth,
at={(0.758in,0.481in)},
scale only axis,
xmin=1,
xmax=30,
ymin=0,
ymax=2000,
xtick={10, 20, 30},
xticklabels={ , , },
ytick={500, 1000, 1500, 2000},
yticklabels={,  ,  },
xmajorgrids,
ymajorgrids,
every x tick label/.append style={font=\color{black}, font=\tiny},
every y tick label/.append style={font=\color{black}, font=\tiny},
axis background/.style={fill=white},
legend columns=-1,
legend style={column sep=0.4ex,
at={(-2.5,-0.4)},anchor=south west,legend cell align=left,align=left,draw=white!15!black}
]
\addplot [color=mycolor1,solid,line width=1.5pt]
  table[row sep=crcr]{%
1	76.9010714327261\\
2	160.216320903881\\
3	197.321572519246\\
4	267.458440351925\\
5	290.474824634916\\
6	302.321847949289\\
7	453.488205339374\\
8	533.015973175395\\
9	580.020065477077\\
10	647.673300409127\\
11	666.209297965717\\
12	834.455090948341\\
13	846.887647568087\\
14	877.775954754195\\
15	957.115592207953\\
16	975.339604858764\\
17	1125.71121394052\\
18	1135.73715796364\\
19	1217.83087155009\\
20	1291.00032754264\\
21	1310.89448767166\\
22	1443.40610158362\\
23	1595.4266087241\\
24	1664.89829521915\\
25	1669.98018998061\\
26	1699.50792374435\\
27	1729.8598920633\\
28	1842.46236167822\\
29	1885.50899545658\\
30	1971.34984772159\\
};
\addlegendentry{\footnotesize style};

\addplot [color=mycolor2,solid,line width=1.5pt]
  table[row sep=crcr]{%
1	88.3629707737712\\
2	134.454141268387\\
3	153.075216787356\\
4	250.173790476382\\
5	275.962727517534\\
6	304.366942634902\\
7	429.966197130343\\
8	503.460227287679\\
9	537.55584903689\\
10	603.119322569931\\
11	609.308283937582\\
12	681.745642960553\\
13	740.305858073451\\
14	847.94618365977\\
15	872.605105490315\\
16	954.447693139654\\
17	1070.75139111217\\
18	1091.84616972481\\
19	1116.871998529\\
20	1134.69912045252\\
21	1257.17918411179\\
22	1349.11421396093\\
23	1390.23619466782\\
24	1467.18544579465\\
25	1506.71428594609\\
26	1605.58097796319\\
27	1637.57892699824\\
28	1703.08521672123\\
29	1779.47205189078\\
30	1819.6004988811\\
};
\addlegendentry{\footnotesize pose 2.50};

\addplot [color=mycolor3,solid,line width=1.5pt]
  table[row sep=crcr]{%
1	79.2989723504283\\
2	156.459998891922\\
3	183.702088267314\\
4	262.692223101826\\
5	284.930102759357\\
6	303.773212546819\\
7	445.424677174353\\
8	532.682589446355\\
9	550.003597846589\\
10	637.278968497589\\
11	637.824277588884\\
12	803.483601275348\\
13	809.368252195014\\
14	861.937167897623\\
15	934.224089855366\\
16	981.379268205223\\
17	1098.33127892653\\
18	1103.17512060531\\
19	1173.00675126468\\
20	1252.05099562721\\
21	1348.91582759957\\
22	1415.31479007149\\
23	1567.59923182577\\
24	1595.59777120562\\
25	1619.28523484596\\
26	1663.66867319533\\
27	1689.19490186562\\
28	1789.64618483622\\
29	1815.87032337913\\
30	1953.10950443495\\
};
\addlegendentry{\footnotesize our 0.80};

\addplot [color=black,dotted,line width=1.5pt]
  table[row sep=crcr]{%
1	77.252326965332\\
2	156.935363769531\\
3	188.590194702148\\
4	262.574096679688\\
5	287.258697509766\\
6	297.611358642578\\
7	437.062744140625\\
8	526.889526367188\\
9	564.394714355469\\
10	637.803527832031\\
11	650.991943359375\\
12	795.350158691406\\
13	817.857971191406\\
14	860.328247070313\\
15	934.784118652344\\
16	971.039916992188\\
17	1105.78527832031\\
18	1120.43676757813\\
19	1193.53076171875\\
20	1268.56372070313\\
21	1318.49890136719\\
22	1416.74755859375\\
23	1559.7021484375\\
24	1592.25952148438\\
25	1615.61645507813\\
26	1674.66979980469\\
27	1703.19384765625\\
28	1813.29968261719\\
29	1845.39770507813\\
30	1948.556640625\\
};

\end{axis}
\end{tikzpicture}%
%
%
\definecolor{mycolor1}{rgb}{0.85000,0.32500,0.09800}%
\definecolor{mycolor2}{rgb}{0.92900,0.69400,0.12500}%
\definecolor{mycolor3}{rgb}{0.00000,0.44700,0.74100}%
\begin{tikzpicture}

\begin{axis}[%
width=\linewidth,
height=\linewidth,
at={(0.758in,0.481in)},
scale only axis,
xmin=1,
xmax=30,
ymin=0,
ymax=2000,
xtick={10, 20, 30},
xticklabels={, , },
ytick={500, 1000, 1500, 2000},
yticklabels={,  ,  },
xmajorgrids,
ymajorgrids,
every x tick label/.append style={font=\color{black}, font=\tiny},
every y tick label/.append style={font=\color{black}, font=\tiny},
axis background/.style={fill=white},
legend columns=-1,
legend style={column sep=0.4ex,
at={(-2.5,-0.4)},anchor=south west,legend cell align=left,align=left,draw=white!15!black}
]
\addplot [color=mycolor1,solid,line width=1.5pt]
  table[row sep=crcr]{%
1	86.5769171616991\\
2	147.580872430346\\
3	177.33762092185\\
4	253.073702363761\\
5	294.275272105751\\
6	306.419315798608\\
7	444.822503363459\\
8	551.26619601695\\
9	563.431277556403\\
10	650.416619502916\\
11	681.848752586112\\
12	732.098162620744\\
13	738.272874077429\\
14	890.074607134684\\
15	966.298126553559\\
16	977.062268107\\
17	1148.2454636891\\
18	1157.20538355355\\
19	1183.74215584459\\
20	1266.54832277462\\
21	1341.09417878632\\
22	1440.92226286031\\
23	1451.15279875867\\
24	1490.78812814107\\
25	1606.91035374424\\
26	1656.5084594141\\
27	1771.04105828289\\
28	1796.18767571959\\
29	1840.02333720174\\
30	1974.43050463968\\
};
\addlegendentry{\footnotesize  style};

\addplot [color=mycolor2,solid,line width=1.5pt]
  table[row sep=crcr]{%
1	70.5958581431916\\
2	146.043542324981\\
3	184.770491195557\\
4	239.62145412957\\
5	267.806042419954\\
6	286.026679432119\\
7	412.219440128059\\
8	515.42879949267\\
9	534.107279603331\\
10	611.628722098408\\
11	619.033797009472\\
12	748.607123116742\\
13	776.11530128949\\
14	835.207408373799\\
15	889.129991559542\\
16	929.073301480069\\
17	1042.61101330734\\
18	1053.47261757463\\
19	1146.32519132369\\
20	1204.85828502895\\
21	1230.61743093987\\
22	1338.23762951288\\
23	1489.74141019347\\
24	1521.44575846944\\
25	1539.06801946462\\
26	1588.73966975127\\
27	1609.13022726167\\
28	1669.13688068606\\
29	1726.12727043797\\
30	1865.81543184612\\
};
\addlegendentry{\footnotesize pose 1.90};

\addplot [color=mycolor3,solid,line width=1.5pt]
  table[row sep=crcr]{%
1	82.2708859988278\\
2	142.120629755038\\
3	172.732452227866\\
4	243.038951276527\\
5	283.692038304759\\
6	293.946128411987\\
7	425.93392524188\\
8	540.18856657102\\
9	544.487780838987\\
10	634.958923101332\\
11	660.612474367961\\
12	698.163191929998\\
13	707.997493087727\\
14	866.155120895888\\
15	931.721286055146\\
16	941.167629201206\\
17	1103.84800546097\\
18	1114.9444105644\\
19	1157.03447759194\\
20	1229.0236026771\\
21	1295.87593875045\\
22	1375.15956049011\\
23	1386.07036322712\\
24	1439.77374596849\\
25	1574.218806827\\
26	1605.93560851308\\
27	1716.34624755265\\
28	1742.50645169469\\
29	1760.13994150573\\
30	1898.75375020775\\
};
\addlegendentry{\footnotesize our 1.00};

\addplot [color=black,dotted,line width=1.5pt]
  table[row sep=crcr]{%
1	85.6326141357422\\
2	143.234252929688\\
3	171.815246582031\\
4	249.364639282227\\
5	285.7685546875\\
6	301.901641845703\\
7	434.410125732422\\
8	546.136352539063\\
9	556.309814453125\\
10	640.426818847656\\
11	674.407043457031\\
12	704.757873535156\\
13	718.499694824219\\
14	884.199157714844\\
15	944.133117675781\\
16	960.303527832031\\
17	1124.75354003906\\
18	1139.67150878906\\
19	1174.16906738281\\
20	1254.18737792969\\
21	1320.47314453125\\
22	1390.75451660156\\
23	1410.94396972656\\
24	1472.27697753906\\
25	1585.12658691406\\
26	1629.00842285156\\
27	1732.10400390625\\
28	1764.58276367188\\
29	1803.1640625\\
30	1946.61157226563\\
};

\end{axis}
\end{tikzpicture}%
\end{minipage}%
\end{tabular}
\vspace{0.01cm}

\caption{\label{fig:style2}Examples of style transfer. The target style (middle) is applied to the target pose (left), obtaining our result (right).}
\end{figure}

\section{Style transfer}
\label{sec:style_tran}

As described in the main manuscript, we apply our trained network for the style transfer task.
In Fig.~\ref{fig:style2} we show some additional examples. 
We emphasize here that our method is correspondence-free, since the input eigenvalues completely encode the target style. Our method also does not rely on the presence of the undeformed source shape, used in previous analogies applications, e.g. \cite{BosEynKouBro15} that phrased the problem of finding $X$ such that ($A \rightarrow B$, $C\rightarrow X$) for some known $A,B,C$.
These examples confirm the robustness and accuracy of the proposed method in this application.

\end{document}